%% file: main.tex
\documentclass[conference]{IEEEtran}
\IEEEoverridecommandlockouts
\usepackage{cite}
\usepackage{amsmath,amssymb,amsfonts}
\usepackage{algorithmic}
\usepackage{graphicx}
\usepackage{textcomp}
\def\BibTeX{{\rm B\kern-.05em{\sc i\kern-.025em b}\kern-.08em
    T\kern-.1667em\lower.7ex\hbox{E}\kern-.125emX}}

\input{setup/packages}
\input{setup/acronyms}

\usepackage{hyperref}  
\usepackage[absolute,showboxes]{textpos}

\TPMargin*{3pt}
% ...
\newcommand\copyrighttext{
	\footnotesize
	\noindent
	\textcopyright\,2023 IEEE.
	Personal use of this material is permitted.
	Permission from IEEE must be obtained for all other uses, in any current or future media, including reprinting/republishing this material for advertising or promotional purposes, creating new collective works, for resale or redistribution to servers or lists, or reuse of any copyrighted component of this work in other works.}%
\newcommand\copyrightnotice{%
	\begin{textblock*}{7in}(0.75in,0.25in)
		\copyrighttext
	\end{textblock*}
}

\begin{document}

\title{Bridging the Gap Between Multi-Step and One-Shot Trajectory Prediction via Self-Supervision\\
{}
\thanks{This work was financially supported by the Federal Ministry of Economic Affairs and Energy of Germany, grant number 19A20026H, based on a decision of the German Bundestag.}
}

\author{\IEEEauthorblockN{Faris Janjo\v{s}}
	\IEEEauthorblockA{\textit{Corporate Research} \\
		\textit{Robert Bosch GmbH}\\
		71272 Renningen, Germany \\
		faris.janjos@de.bosch.com}
	\and
	\IEEEauthorblockN{Max Keller}
	\IEEEauthorblockA{\textit{Corporate Research} \\
		\textit{Robert Bosch GmbH}\\
		71272 Renningen, Germany \\
		max.keller@de.bosch.com}
	\and
	\IEEEauthorblockN{Maxim Dolgov}
	\IEEEauthorblockA{\textit{Corporate Research} \\
		\textit{Robert Bosch GmbH}\\
		71272 Renningen, Germany \\
		maxim.dolgov@de.bosch.com}
	\and
	\IEEEauthorblockN{J. Marius Z\"ollner}
	\IEEEauthorblockA{\textit{Research Center for} \\
		\textit{Information Technology (FZI)}\\
		76131 Karlsruhe, Germany \\
		zoellner@fzi.de}
}

\copyrightnotice

\maketitle

\input{chapters/abstract}
\input{chapters/introduction}
\input{chapters/related_work}
\input{chapters/methods}
\input{chapters/results}
\input{chapters/conclusion}
\bibliographystyle{IEEEtran}
\bibliography{references/bibliography}

\end{document}

%% file: setup/packages.tex
\usepackage{amsmath} 
\usepackage{amssymb}  
\usepackage{mathtools}
\usepackage{siunitx}

\usepackage[table,xcdraw]{xcolor}
\usepackage{graphicx}
\usepackage{tabularx,booktabs}
\newcolumntype{Y}{>{\centering\arraybackslash}X}

\usepackage{todonotes}
\usepackage[nolist]{acronym}
\usepackage{flushend}
\usepackage[font=small]{caption}
\usepackage[font=small]{subcaption}

\makeatletter
\renewcommand*\env@matrix[1][*\c@MaxMatrixCols c]{%
	\hskip -\arraycolsep
	\let\@ifnextchar\new@ifnextchar
	\array{#1}}
\makeatother

\makeatletter
\let\NAT@parse\undefined
\makeatother

\usepackage{tikz}
\usetikzlibrary{arrows}
\usetikzlibrary{fadings}
\usetikzlibrary{patterns}
\usetikzlibrary{shadows.blur}
\usetikzlibrary{shapes}
\tikzset{
	vertex/.style={circle,draw,minimum size=2.5em},
	edge/.style={->,> = latex'}
}

\usepackage{pifont}
\newcommand{\cmark}{\ding{51}}%
\newcommand{\xmark}{\ding{55}}%

%% file: setup/acronyms.tex
\begin{acronym}
\acro{GPU}{Graphics Processing Units}
\acro{DNN}{Deep Neural Networks}
\acro{CNN}{Convolutional Neural Network}
\acro{RNN}{Recurrent Neural Networks}
\acro{GNN}{Graph Neural Networks}
\acro{GAT}{Graph Attention Networks}
\acro{GCN}{Graph Convolutional Networks}
\acro{GRU}{Gated Recurrent Unit}
\acro{LSTM}{Long Short-Term Memory}
\acro{KF}{Kalman Filters}
\acro{BEV}{Bird's Eye View}
\acro{GAN}{Generative Adversarial Networks}
\acro{MTP}{Multi-Modal Trajectory Prediction}
\acro{MLP}{Multilayer Perceptron}
\acro{MAE}{Mean Absolute Error}
\acro{MSE}{Mean Squared Error}
\acro{ADE}{Average Displacement Error}
\acro{FDE}{Final Displacement Error}
\acro{minADE}{Minimum Average Displacement Error}
\acro{minFDE}{Minimum Final Displacement Error}
\acro{DKM}{Deep Kinematic Models}
\acro{FF-ASP}{Feed-Forward Action-Space Predictor}
\acro{SS-ASP}{Self-Supervised Action-Space Predictor}
\acro{VAE}{Variational Autoencoders}
\acro{CVAE}{Conditional Variational Autoencoder}
\acro{AV}{autonomous vehicle}
\acro{MHA}{Multi-Head Attention}
\acro{SAN}{Scene Anchor Networks}
\acro{MB-SS-ASP}{Multi-Branch Self-Supervised Action-Space Predictor}
\acro{AD}{Autonomous Driving}
\acro{DL}{Deep Learning}
\acro{RL}{Reinforcement Learning}
\acro{MR}{Miss Rate}
\acro{IL}{Imitation Learning}
\acro{MDP}{Markov Decision Process}
\acro{POMDP}{Partially Observable Markov Decision Process}
\acro{RSSM}{Recurrent State Space Models}
\end{acronym}

%% file: chapters/abstract.tex
\begin{abstract}
Accurate vehicle trajectory prediction is an unsolved problem in autonomous driving with various open research questions. State-of-the-art approaches regress trajectories either in a one-shot or step-wise manner. Although one-shot approaches are usually preferred for their simplicity, they relinquish powerful self-supervision schemes that can be constructed by chaining multiple time-steps. We address this issue by proposing a middle-ground where multiple trajectory segments are chained together. Our proposed Multi-Branch Self-Supervised Predictor receives additional training on new predictions starting at intermediate future segments. In addition, the model 'imagines' the latent context and 'predicts the past' while combining multi-modal trajectories in a tree-like manner. We deliberately keep aspects such as interaction and environment modeling simplistic and nevertheless achieve competitive results on the INTERACTION dataset. Furthermore, we investigate the sparsely explored uncertainty estimation of deterministic predictors. We find positive correlations between the prediction error and two proposed metrics, which might pave way for determining prediction confidence.

\end{abstract}

%% file: chapters/introduction.tex
\section{INTRODUCTION}\label{sec:intro}
The common separation of the \ac{AD} stack into perception, prediction, and planning components drives a need for accurate forecasts of the future as a planner input. In prediction, different
challenges exist such as but not limited to, representing the environment~\cite{gao2020vectornet}, modeling multi-agent interactions~\cite{AgentFormer}, capturing the multi-modality of the future motion distribution~\cite{varadarajan2021multipath++}, adhering to kinematic constraints~\cite{janjos2021action}, as well as modeling a long prediction horizon~\cite{KEMP}. In solving these tasks, various \ac{DL} models usually generate predicted trajectories of the agents sharing a road situation with an \ac{AV}.

\begin{figure}[h]
	\centering
	\hspace{-6pt}\scalebox{0.5}{\input{images/multi-tree-illustration.tex}}\vspace{-10pt}
	\caption{\small Left: the Multi-Branch Self-Supervised Predictor splits the trajectory prediction problem into three equal time-segments, where the ground-truth history and future are shown in green. It builds a tree of multi-modal trajectories over two future segments (yellow and orange), with a multi-modal trajectory (two modes) chained at the output of each previous segment mode. Additionally, it 'predicts' a uni-modal trajectory in the past (blue). Right: knowing the ground-truth in training, the model builds another tree shifted by one segment into the future. By constructing multiple trees, reconstructing the past, as well as 'imagining' how the context will evolve in each segment (not depicted), the model receives additional (self-)supervision.}
	\vspace{-13pt}
	\label{fig:illustration}
\end{figure}
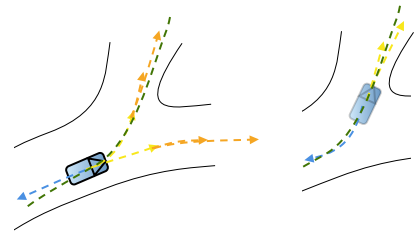

In terms of the approach to construct trajectories, prediction models can be categorized into one-shot approaches \cite{janjos2021action, janjovs2021starnet, gilles2021gohome, casas2020implicit}, where a full trajectory is directly regressed, and step-wise, autoregressive approaches that generate a trajectory sequentially for each time step based on the previously predicted time steps~\cite{huang2022hyper, rhinehart2019precog, tang2019multiple, OccupancyFlowFields, scibior2021imagining, KEMP}. The main drawback of one-shot approaches is that long prediction horizons make it difficult to reason about comprehensive changes in scene dynamics. Since the models do not condition on future observations, the potential for errors and high uncertainty is much larger toward the end of a long prediction horizon. As opposed to one-shot models, step-wise, autoregressive approaches build each prediction based on the previous inference step. Similarly to one-shot approaches, they tend to accumulate errors for longer horizons, especially in interactive settings under distribution shift~\cite{ross2011reduction}. Furthermore, the autoregressive context makes it non-trivial to combine multi-modal predictions between successive time steps. In our work, we aim to bridge the two approaches and offer potential answers to the drawbacks at hand. 

Our starting point in addressing the question of combining the different paradigms is the \ac{SS-ASP} model in~\cite{janjos2021action}. Its main characteristic is the ability to predict a latent representation of the driving context (e.g. map and motion of non-predicted agents) prior to predicting future trajectories for the vehicle of interest. In achieving this, it is trained with future context data in addition to future trajectories of the predicted agents, which separates it from the existing state-of-the-art models. Furthermore, it can 'go back in time' and check whether its reconstructions of the past are consistent with the observed ground-truth history through an auxiliary, inverse model task. These two aspects can be regarded as a form of self-supervision and they enable the model to interpret the trajectory prediction problem in a segment-wise manner, where each segment consists of multiple time steps (for example, one second length) and multiple segments are chained together. With each new segment, future trajectories and context are predicted, and past trajectories of the previous segment are reconstructed. This is in contrast to pure one-shot or step-wise approaches, which do not employ such strategies. However, these abilities are sparsely explored in~\cite{janjos2021action} and the proposed chaining of segments actually regresses the results compared to a simple one-shot-like setup. 

In this work, we aim to incorporate novel autoregressive schemes from \cite{Venkatraman2015ImprovingMP, Overshooting, LatentDynamicsModelPredictiveAgents} to enable a multi-segment formulation of a one-shot trajectory prediction model. With this, we aim to answer the \textit{research question (i): can the prediction performance of a one-shot model be improved by an autoregressive formulation that creates room for additional self-supervision?} Furthermore, in light of the multi-segment structure, we aim to investigate how the model uncertainty evolves with each segment, given a deterministic problem formulation. In this sense, we aim to answer the \textit{research question (ii): can the confidence of the deterministic multi-segment model be accurately rated, in order to define the limitations of the model?} This might pave the way toward the prediction of a variable number of segments and thus a variable time-horizon.\footnote{A use-case is stopping the prediction after a certain segment in case the uncertainty is too high. Providing only the 'confident' predictions to a planner might improve its performance since the output is not based on highly uncertain predictions (and thus potentially overconfident).} The main contributions of our work can be summarized along: 
\begin{itemize}
	\item Segment-wise prediction: a novel paradigm where the resulting trajectory is obtained by predicting a number of $1$-second time segments and chaining them together.
	\item Multi-Branch \ac{SS-ASP}: a multi-branch and multi-segment extension of the approach from~\cite{janjos2021action}, trained by generating additional prediction paths (termed branches). The proposed model 'imagines' future context, reconstructs past trajectories, and combines segment-wise multi-modal predictions with various tree-search strategies. The model is deliberately simplistic in terms of interaction and multi-modality modeling but still competitive on the INTERACTION dataset~\cite{zhan2019interaction}. 
	\item Uncertainty of deterministic predictors: we assess the given model's confidence along evolving time-segments with two novel approaches that exhibit a significant positive correlation with the measured prediction error.
\end{itemize}

%% file: images/multi-tree-illustration.tex

\tikzset {_186ryotoc/.code = {\pgfsetadditionalshadetransform{ \pgftransformshift{\pgfpoint{0 bp } { 0 bp }  }  \pgftransformrotate{0 }  \pgftransformscale{2 }  }}}
\pgfdeclarehorizontalshading{_4u5cckn9j}{150bp}{rgb(0bp)=(0.81,0.91,0.98);
	rgb(37.5bp)=(0.81,0.91,0.98);
	rgb(62.5bp)=(0.39,0.58,0.76);
	rgb(100bp)=(0.39,0.58,0.76)}


\tikzset {_sr6hzgb0t/.code = {\pgfsetadditionalshadetransform{ \pgftransformshift{\pgfpoint{0 bp } { 0 bp }  }  \pgftransformrotate{0 }  \pgftransformscale{2 }  }}}
\pgfdeclarehorizontalshading{_j9ugjwnul}{150bp}{rgb(0bp)=(0.81,0.91,0.98);
	rgb(37.5bp)=(0.81,0.91,0.98);
	rgb(62.5bp)=(0.39,0.58,0.76);
	rgb(100bp)=(0.39,0.58,0.76)}
\tikzset{every picture/.style={line width=0.75pt}} 

\begin{tikzpicture}[x=0.75pt,y=0.75pt,yscale=-1,xscale=1]
	
	\path  [shading=_4u5cckn9j,_186ryotoc] (345.68,94.42) .. controls (343.79,93.53) and (342.99,91.29) .. (343.9,89.4) -- (357.57,60.98) .. controls (358.47,59.1) and (360.74,58.29) .. (362.63,59.18) -- (372.91,64.01) .. controls (374.8,64.9) and (375.6,67.15) .. (374.69,69.03) -- (361.02,97.45) .. controls (360.11,99.33) and (357.84,100.14) .. (355.95,99.25) -- cycle ; 
	\draw  [color={rgb, 255:red, 0; green, 0; blue, 0 }  ,draw opacity=0.2 ][line width=1.5]  (345.68,94.42) .. controls (343.79,93.53) and (342.99,91.29) .. (343.9,89.4) -- (357.57,60.98) .. controls (358.47,59.1) and (360.74,58.29) .. (362.63,59.18) -- (372.91,64.01) .. controls (374.8,64.9) and (375.6,67.15) .. (374.69,69.03) -- (361.02,97.45) .. controls (360.11,99.33) and (357.84,100.14) .. (355.95,99.25) -- cycle ; 
	
	\draw  [color={rgb, 255:red, 0; green, 0; blue, 0 }  ,draw opacity=0.2 ][line width=1.5]  (368.23,62.13) -- (369.31,78.89) -- (353.19,71.41) -- cycle ;
	
	\path  [shading=_j9ugjwnul,_sr6hzgb0t] (57.85,147.3) .. controls (57.09,145.29) and (58.08,142.98) .. (60.04,142.13) -- (90.39,129.07) .. controls (92.35,128.22) and (94.57,129.16) .. (95.33,131.17) -- (99.45,142.07) .. controls (100.21,144.08) and (99.23,146.39) .. (97.26,147.24) -- (66.92,160.3) .. controls (64.95,161.15) and (62.74,160.21) .. (61.98,158.2) -- cycle ; 
	\draw  [line width=1.5]  (57.85,147.3) .. controls (57.09,145.29) and (58.08,142.98) .. (60.04,142.13) -- (90.39,129.07) .. controls (92.35,128.22) and (94.57,129.16) .. (95.33,131.17) -- (99.45,142.07) .. controls (100.21,144.08) and (99.23,146.39) .. (97.26,147.24) -- (66.92,160.3) .. controls (64.95,161.15) and (62.74,160.21) .. (61.98,158.2) -- cycle ; 
	
	\draw  [line width=1.5]  (97.35,137.36) -- (86.25,151.34) -- (79.7,134.31) -- cycle ;
	
	\draw [color={rgb, 255:red, 74; green, 144; blue, 226 }  ,draw opacity=1 ][line width=1.5]  [dash pattern={on 5.63pt off 4.5pt}]  (77.15,146) .. controls (49.89,156.57) and (34.09,164.31) .. (11.43,174.21) ;
	\draw [shift={(7.79,175.79)}, rotate = 336.61] [fill={rgb, 255:red, 74; green, 144; blue, 226 }  ,fill opacity=1 ][line width=0.08]  [draw opacity=0] (9.29,-4.46) -- (0,0) -- (9.29,4.46) -- cycle    ;
	\draw    (-6.55,139.85) .. controls (-1.22,131.45) and (-16.81,129.72) .. (41.19,107.72) .. controls (99.19,85.72) and (96.81,49.5) .. (101.86,16.92) ;
	\draw [color={rgb, 255:red, 248; green, 231; blue, 28 }  ,draw opacity=1 ][line width=1.5]  [dash pattern={on 5.63pt off 4.5pt}]  (77.95,145.2) .. controls (106.98,136.8) and (117.04,114.53) .. (124.86,89.55) ;
	\draw [shift={(125.95,86)}, rotate = 106.86] [fill={rgb, 255:red, 248; green, 231; blue, 28 }  ,fill opacity=1 ][line width=0.08]  [draw opacity=0] (9.29,-4.46) -- (0,0) -- (9.29,4.46) -- cycle    ;
	\draw [color={rgb, 255:red, 245; green, 166; blue, 35 }  ,draw opacity=1 ][line width=1.5]  [dash pattern={on 5.63pt off 4.5pt}]  (122.75,93.2) .. controls (130.51,75.35) and (139.02,49.98) .. (150.48,9.04) ;
	\draw [shift={(151.55,5.2)}, rotate = 105.52] [fill={rgb, 255:red, 245; green, 166; blue, 35 }  ,fill opacity=1 ][line width=0.08]  [draw opacity=0] (9.29,-4.46) -- (0,0) -- (9.29,4.46) -- cycle    ;
	\draw [color={rgb, 255:red, 245; green, 166; blue, 35 }  ,draw opacity=1 ][line width=1.5]  [dash pattern={on 5.63pt off 4.5pt}]  (124.35,93.2) .. controls (128.42,74.83) and (126.1,89.84) .. (134.72,50.6) ;
	\draw [shift={(135.55,46.8)}, rotate = 102.31] [fill={rgb, 255:red, 245; green, 166; blue, 35 }  ,fill opacity=1 ][line width=0.08]  [draw opacity=0] (9.29,-4.46) -- (0,0) -- (9.29,4.46) -- cycle    ;
	\draw [color={rgb, 255:red, 248; green, 231; blue, 28 }  ,draw opacity=1 ][line width=1.5]  [dash pattern={on 5.63pt off 4.5pt}]  (85.15,142.8) .. controls (107.92,134.42) and (120.88,131.81) .. (147.36,122.41) ;
	\draw [shift={(150.75,121.2)}, rotate = 160.14] [fill={rgb, 255:red, 248; green, 231; blue, 28 }  ,fill opacity=1 ][line width=0.08]  [draw opacity=0] (9.29,-4.46) -- (0,0) -- (9.29,4.46) -- cycle    ;
	\draw [color={rgb, 255:red, 245; green, 166; blue, 35 }  ,draw opacity=1 ][line width=1.5]  [dash pattern={on 5.63pt off 4.5pt}]  (144.05,124.07) .. controls (172.7,115.47) and (171.71,116.91) .. (196.82,115.75) ;
	\draw [shift={(200.55,115.57)}, rotate = 177.04] [fill={rgb, 255:red, 245; green, 166; blue, 35 }  ,fill opacity=1 ][line width=0.08]  [draw opacity=0] (9.29,-4.46) -- (0,0) -- (9.29,4.46) -- cycle    ;
	\draw    (174.52,18.06) .. controls (121.86,95.39) and (164.52,82.92) .. (203.86,79.59) ;
	\draw    (-3.55,211.85) .. controls (30.45,189.85) and (47.86,183.39) .. (98.52,164.06) .. controls (149.19,144.72) and (186.52,143.59) .. (207.86,141.59) ;
	\draw  [draw opacity=0][fill={rgb, 255:red, 255; green, 255; blue, 255 }  ,fill opacity=1 ] (-25.55,78.95) -- (5.28,78.95) -- (5.28,232.8) -- (-25.55,232.8) -- cycle ;
	\draw    (226.95,128.8) .. controls (232.28,120.4) and (216.69,118.67) .. (274.69,96.67) .. controls (332.69,74.67) and (330.31,38.45) .. (335.36,5.87) ;
	\draw [color={rgb, 255:red, 248; green, 231; blue, 28 }  ,draw opacity=1 ][line width=1.5]  [dash pattern={on 5.63pt off 4.5pt}]  (361.25,75.65) .. controls (363.32,65.77) and (363.37,70.85) .. (377.01,19.29) ;
	\draw [shift={(377.87,16.07)}, rotate = 104.77] [fill={rgb, 255:red, 248; green, 231; blue, 28 }  ,fill opacity=1 ][line width=0.08]  [draw opacity=0] (9.29,-4.46) -- (0,0) -- (9.29,4.46) -- cycle    ;
	\draw [color={rgb, 255:red, 248; green, 231; blue, 28 }  ,draw opacity=1 ][line width=1.5]  [dash pattern={on 5.63pt off 4.5pt}]  (360.85,75.15) .. controls (367.17,57.12) and (380.53,32.12) .. (396.84,-1.31) ;
	\draw [shift={(398.37,-4.43)}, rotate = 115.91] [fill={rgb, 255:red, 248; green, 231; blue, 28 }  ,fill opacity=1 ][line width=0.08]  [draw opacity=0] (9.29,-4.46) -- (0,0) -- (9.29,4.46) -- cycle    ;
	\draw    (408.02,7.01) .. controls (355.36,84.34) and (398.02,71.87) .. (437.36,68.54) ;
	\draw    (229.95,200.8) .. controls (263.95,178.8) and (281.36,172.34) .. (332.02,153.01) .. controls (382.69,133.67) and (420.02,132.54) .. (441.36,130.54) ;
	\draw [color={rgb, 255:red, 65; green, 117; blue, 5 }  ,draw opacity=1 ][line width=1.5]  [dash pattern={on 5.63pt off 4.5pt}]  (18.87,182.57) .. controls (31.87,172.57) and (60.87,156.07) .. (85.87,142.07) .. controls (110.87,128.07) and (128.37,94.57) .. (144.87,45.07) .. controls (161.37,-4.43) and (159.28,6.58) .. (162.37,-8.43) ;
	\draw  [draw opacity=0][fill={rgb, 255:red, 255; green, 255; blue, 255 }  ,fill opacity=1 ] (246.06,167.44) -- (299.42,130.56) -- (308.44,143.6) -- (255.08,180.48) -- cycle ;
	\draw [color={rgb, 255:red, 74; green, 144; blue, 226 }  ,draw opacity=1 ][line width=1.5]  [dash pattern={on 5.63pt off 4.5pt}]  (356.45,81.35) .. controls (346.9,121.46) and (327.7,129.39) .. (301.6,135.25) ;
	\draw [shift={(297.87,136.07)}, rotate = 347.91] [fill={rgb, 255:red, 74; green, 144; blue, 226 }  ,fill opacity=1 ][line width=0.08]  [draw opacity=0] (9.29,-4.46) -- (0,0) -- (9.29,4.46) -- cycle    ;
	\draw  [draw opacity=0][fill={rgb, 255:red, 255; green, 255; blue, 255 }  ,fill opacity=1 ] (406,37.67) -- (469.87,37.67) -- (469.87,190.23) -- (406,190.23) -- cycle ;
	\draw [color={rgb, 255:red, 65; green, 117; blue, 5 }  ,draw opacity=1 ][line width=1.5]  [dash pattern={on 5.63pt off 4.5pt}]  (250.87,171.23) .. controls (263.87,161.23) and (292.87,144.73) .. (317.87,130.73) .. controls (342.87,116.73) and (360.37,83.23) .. (376.87,33.73) .. controls (393.37,-15.77) and (391.28,-4.76) .. (394.37,-19.77) ;
	\draw  [draw opacity=0][fill={rgb, 255:red, 255; green, 255; blue, 255 }  ,fill opacity=1 ] (222.87,54.5) -- (294.87,54.5) -- (294.87,207.07) -- (222.87,207.07) -- cycle ;
	\draw [color={rgb, 255:red, 245; green, 166; blue, 35 }  ,draw opacity=1 ][line width=1.5]  [dash pattern={on 5.63pt off 4.5pt}]  (146.05,124.07) .. controls (174.85,115.43) and (219.32,116.46) .. (248.49,115.23) ;
	\draw [shift={(252.05,115.07)}, rotate = 177.04] [fill={rgb, 255:red, 245; green, 166; blue, 35 }  ,fill opacity=1 ][line width=0.08]  [draw opacity=0] (9.29,-4.46) -- (0,0) -- (9.29,4.46) -- cycle    ;

\end{tikzpicture}

%% file: chapters/related_work.tex
\section{RELATED WORK}\label{sec:rel_work}
The focus of our work is self-supervised autoregressive trajectory prediction and uncertainty estimation in the context of our proposed model. Therefore, we are interested in \textbf{autoregressive models} (both in trajectory prediction and other contexts), \textbf{self-supervision}, and \textbf{uncertainty estimation} in trajectory prediction; we outline the section accordingly.

\subsection{Autoregressive prediction}

\label{sec:autoregressive-trajectory-prediction-rw}
Autoregressive trajectory prediction models~\cite{huang2022hyper, KEMP, rhinehart2019precog, tang2019multiple, OccupancyFlowFields, scibior2021imagining} incrementally model changes in scene dynamics compared to one-shot models~\cite{janjos2021action, janjovs2021starnet, gilles2021gohome, casas2020implicit}, which have larger requirements on model expressiveness within a single prediction. Among such models,~\cite{tang2019multiple} predicts the waypoints of the next step based on an agent's own state and the waypoints of the surrounding agents at the previous step. It handles multi-modality by predicting a fixed amount of latent intents, which condition the step-wise rolled-out trajectories. Similarly, \cite{huang2022hyper} extends the problem by inferring time-varying discrete intents of the surrounding agents, which are incorporated into the step-wise discrete-continuous hybrid model. It uses a learned proposal function to 'traverse' the system and obtain multiple modes. In~\cite{hu2022model}, step-wise environment prediction is done, similar to the latent context prediction in~\cite{janjos2021action}. The environment prediction benefits from the autoregressive formulation; it is simpler to predict observations in a single time step than a long horizon.

Outside of vehicle trajectory prediction, autoregressive models are present in robotics and \ac{IL}~\cite{Overshooting,li2021replay,LatentDynamicsModelPredictiveAgents,Venkatraman2015ImprovingMP}. In~\cite{Overshooting}, the so-called latent overshooting method is introduced that rolls-out new autoregressive predictions at each intermediate future step, in parallel to the first prediction sequence. This allows to increase the learning signal without additional data. An autoregressive formulation is useful in Model-Based \ac{RL} as well;~\cite{LatentDynamicsModelPredictiveAgents} showed that a multi-step loss, based on the autoregressively predicted steps, increases the reward for deterministic planning modules compared to a single-step objective. Similarly, the Dreamer model in~\cite{hafner2019dream} learns behavior directly from autoregressive latent predictions ('imagination') instead of exploration. This significantly reduces the training time compared to an explorative \ac{RL} agent. In summary, the potential for performance improvement as well as the larger design space compared to one-shot models motivate the usage of an autoregressive formulation in this work.

\subsection{Self-supervision in trajectory prediction} \label{sec:self-supervised-trajectory-prediction-rw}
Existing trajectory prediction models incorporate self-supervision either through (i) a separate training stage or (ii) through additional tasks. There are fewer approaches performing (ii) in the literature; ~\cite{ma2021multi} proposes a contrastive pre-training in which rasterized representations of intersecting trajectories are rotated or semantics are exchanged in order to learn an internal interaction representation. In contrast,~\cite{geisslinger2021watch} fine-tunes a pre-trained predictor in an online-setting to adapt to behaviors observed in inference. Among (ii),~\cite{ye2022dcms} enforces additional temporal and spatial consistency tasks for trajectory refinement that robustify the outputs in terms of pertubations. Additionally,~\cite{VectorNet} takes a graph-based approach where certain map and agent node features are masked out and presented as a completion task for the model.

Learning environment models through self-supervision and using the internal representations of scene dynamics for planning has received strong attention in \ac{RL}~\cite{hu2022model,wu2022daydreamer,koh2021pathdreamer}. It has shown to be a promising direction in \ac{IL}-based \ac{AD} as well; in~\cite{hafner2019dream} the model predicts an evolution of the scene in the latent space and reconstructs camera images, semantic maps, and actions taken by the \ac{AV}. Similarly, the \ac{SS-ASP} model~\cite{janjos2021action} performs prediction in the latent space but without full observation reconstruction. Instead, latent context predictions are compared to encoded future observations and an inverse model is learned as well~\cite{agrawal2016learning, pathak2017curiosity}, which introduces another transition prior on the environment. Compared to~\cite{hafner2019dream}, this is more efficient (due to lower dimensionality), however, reconstructing rich observations from the latent space induces stronger requirements on the expressiveness of the latent space. 

\subsection{Prediction uncertainty estimation} \label{sec:prediction-uncertainty-evaluation-rw}
Despite the large number of deterministic trajectory predictors in literature and the importance of communicating the model uncertainty to a planner, the task has received limited attention in literature. In general, estimating the epistemic uncertainty of a prediction model is a challenging problem in \ac{DL}. Several approaches in trajectory prediction use the computationally cumbersome deep ensembles~\cite{filos2020can,varadarajan2021multipath++}. A more general approach is Bayesian inference, where uncertainty is directly estimated in conjunction with the prediction. However, significantly more effort is needed to design and train Bayesian networks compared to standard neural networks. In~\cite{DropoutAsBayesianApproximation}, a theoretical framework is described that ``casts dropout training in deep neural networks as approximate Bayesian inference in deep Gaussian processes``. In practice, the parameters of a Gaussian distribution are approximated by the mean and variance of multiple inference runs, each with different deactivated (dropped-out) neurons. Along these results,~\cite{nayak2022uncertainty} provide a study of dropout-based Bayesian approximation in pedestrian trajectory prediction and find improved accuracy through in inference. In our work, we apply dropout-based techniques to estimate the prediction uncertainty of the developed model, due to the theoretical grounding and ease-of-use.  


%% file: chapters/methods.tex
\section{METHOD}
\vspace{-1pt}
In this section, we describe our method. In Sec.~\ref{sec:trajectory-prediction} we define the addressed problem of trajectory prediction and introduce notation. Sec. \ref{sec:ssp-asp} describes the \ac{SS-ASP} model from ~\cite{janjos2021action}. Sec.~\ref{sec:multi-segment-ss-asp} extends the \ac{SS-ASP} into an autoregressive formulation and offers strategies how to combine segment-wise multi-modal predictions. The proposed Multi-Branch \ac{SS-ASP} model is given in Sec.~\ref{sec:multi-branch-ss-asp}. Strategies how to determine the prediction uncertainty of the Multi-Branch \ac{SS-ASP} are introduced in Sec.~\ref{sec:uncertainty-estimation-methods}. 
\subsection{Problem definition and notation}\label{sec:trajectory-prediction}
Vehicle trajectory prediction can be framed as non-interactive imitation learning~\cite{rhinehart2019precog}, where given the observed information $\mathcal{D}$ and ground-truth future trajectories $Y^*$, we learn the conditional distribution $P(\hat{Y}|\mathcal{D})$ of future trajectories $\hat{Y}$. In practice, deterministic models represent the distribution by predicting $K$ likely samples (modes) $\{Y_j \}^{K}_{j=1}$ as well as their associated pseudo-probabilities $\{p_j \}^{K}_{j=1}$. Furthermore, the prediction can be performed for a single vehicle or jointly for multiple vehicles in a scene. Even though joint prediction is a more sound approach to the problem~\cite{janjovs2021starnet}, we limit the analysis to a single-agent setting for simplicity. Nevertheless, the proposed model has no methodological restrictions preventing an extension to joint prediction.   

The proposed model uses a segment-wise prediction formulation where a time segment is a sequence of time steps and a full $T$ time step prediction consists of $N$ equal length segments\footnote{In this sense, a one-shot prediction is a single-segment prediction.}. Thus, we introduce supporting notation:
\begin{itemize}
	\item $i$ is the segment index, $i\in[1,...,N]$
	\item $t$ is the number of time steps in a segment, $T=N\cdot t$ 
	\item $\tau_{i}$ describes a future time segment $i$, $\left((i-1)\cdot t: i t\right]$
	\item $\tau_{0}$ describes a single past time segment, $\left(-t: 0\right]$
\end{itemize}

\subsection{Self-Supervised Action-Space Predictor (SS-ASP)} \label{sec:ssp-asp}
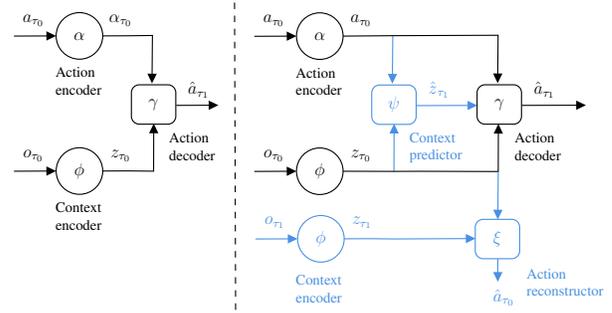
\begin{figure}
	\centering
	\scalebox{0.55}{\input{images/ssp_components.tex}}\vspace{-3pt}
	\caption{Left: the (non-self-supervised) \ac{FF-ASP}~\cite{janjos2021action}, conceptually very common in the literature (standard encoder-decoder structure) when action-spaces are excluded. Past actions~$a_{\tau_0}$ and observations~$o_{\tau_0}$ are encoded via an action encoder $\alpha$ and a context encoder $\phi$ into features~$\alpha_{\tau_0}$ (with a slight abuse of notation) and~$z_{\tau_0}$. Future actions~$\hat{a}_{\tau_1}$ are predicted by the decoder~$\gamma$. Right: self-supervised model with additional components in blue. The model additionally predicts future latent context~$\hat{z}_{\tau_1}$ with the context predictor~$\psi$ and trains it against the encoding~$z_{\tau_1}$ as pseudo-ground-truth. Furthermore, it reconstructs past actions~$\hat{a}_{\tau_0}$ with an inverse model~$\xi$ and trains them against the past ground-truth~$a_{\tau_0}$. Multi-modal predicted actions and kinematic models converting actions to positions are omitted for clarity. }
	\label{fig:ssp_components}\vspace{-13pt}
\end{figure}

The \ac{SS-ASP} model ~\cite{janjos2021action} is the basis for developing the multi-segment model proposed in this work. It is an action-space prediction model, i.e. it predicts actions (accelerations and steering angles) and obtains positions via a kinematic model. At a high-level, it uses encoders for capturing past environment context information (e.g. a CNN encoding birds-eye-view grids or a GNN operating on graphs) into a latent (context) feature vector. Furthermore, it uses an action-based encoder for encoding past actions (e.g. RNN) and a multi-modal action-based decoder for regressing future actions (e.g. RNN). This does not separate it conceptually from a multitude of state-of-the-art approaches (irrespective of the action-space), since a vast majority uses a similar setup of encoding past information (context, trajectories) and predicting future trajectories. The described architecture is depicted in the left part of Fig.~\ref{fig:ssp_components} (so-called \ac{FF-ASP}~\cite{janjos2021action}), where the context and action encoders, and action decoder are parameterized by $\phi$, $\alpha$, and $\gamma$, respectively.

The \ac{SS-ASP} model stands out in the sense that, additionally to the aforementioned encoder and decoder components, it predicts a latent future context prior to predicting future actions. It trains this predicted future context against its own encoding of the future context. Furthermore, it reconstructs past actions via an inverse model taking in future actions. These two self-supervised tasks serve as additional regularization for the model. The SS-ASP model is depicted in the right part of Fig.~\ref{fig:ssp_components}, where the new context predictor component is parameterized by $\psi$, and the action reconstructor with $\xi$. For encoding the future context, the same past context encoder $\phi$ is reused, in this case receiving future information during training. The loss function of the model is \vspace{-3pt}
\begin{align}
\mathcal{L}_{SS-ASP} = \mathcal{L}_{traj} + \mathcal{L}_{class} + \mathcal{L}_{context} + \mathcal{L}_{recon} \label{eq:ssp-asp-loss} \ , 
\end{align}
with weights omitted for clarity. The trajectory regression loss is $\mathcal{L}_{traj} = \Vert\hat{Y}_{\tau_1}(\hat{z}_{\tau_1}) - Y^*_{\tau_1}|| + ||\hat{Y}_{\tau_1}(z_{\tau_1}) - Y^*_{\tau_1}||$; its two terms reflect the fact that the action decoder $\gamma$ in Fig.~\ref{fig:ssp_components} is called with both predicted and encoded future context (Fig.~\ref{fig:ssp_components} only shows the former) in training to promote consistency between components. The loss function~\eqref{eq:ssp-asp-loss} considers multi-modal outputs via the winner-takes-all~\cite{khandelwal2020if} approach. The classification loss function~$\mathcal{L}_{class}$ considers mode probabilities via cross-entropy. The context loss~$\mathcal{L}_{context} = \Vert\hat{z}_{\tau_1} - z_{\tau_1}\Vert$ penalizes the mismatch between encoded and predicted future context, while the reconstruction $\mathcal{L}_{recon} = \Vert\hat{Y}_{\tau_0}(\hat{z}_{\tau_1}) - Y_{\tau_0}\Vert + \Vert\hat{Y}_{\tau_0}(z_{\tau_1}) - Y_{\tau_0}\Vert$ considers past 'predictions' (two terms promoting consistency similar to $\mathcal{L}_{traj}$). For more details, see~\cite{janjos2021action}. 

\subsection{Multi-Segment SS-ASP} \label{sec:multi-segment-ss-asp}
\begin{figure}
	\begin{subfigure}[t]{0.5\columnwidth}
	\scalebox{0.55}{\input{mutli_segment_ss_asp_vs_ss_asp.tex}}
\centering
\caption{High-level model comparison in terms of context and (uni-modal) action prediction. The \ac{FF-ASP} uses only past features and actions to predict future actions. The \ac{SS-ASP} predicts future features (blue) prior to future actions (action reconstruction not depicted). The Multi-Segment \ac{SS-ASP} splits the future into segments and chains successive feature/action predictions. This induces a distribution shift however, since predictions are chained on top of predictions. \label{fig:multi-segment-SS-ASP-abstract-example}}\vspace{-3pt}
	\end{subfigure}\hfill
\begin{subfigure}[]{0.45\columnwidth}
	\scalebox{0.55}{\input{multi_branch_ss_asp_abstract_latent_context_view.tex}}
	\centering
	\vspace{6pt}
	\caption{High-level visualization of branched overshooting~\cite{li2021replay} over $N=3$ future segments for predicted actions. At each intermediate future segment, a new prediction branch is started (denoted with superscript $b$, $\hat{a}^b_{\tau_i}$), with a shifted history and a shorter future prediction. In addition to predicting future actions in a branched manner, we branch context features as well as reconstructed actions (not visualized).\label{fig:prediction-branches}} 
\end{subfigure}
\caption{Multi-segment (a) and multi-branch (b) SS-ASP depictions.}
\vspace{-5pt}
\end{figure}
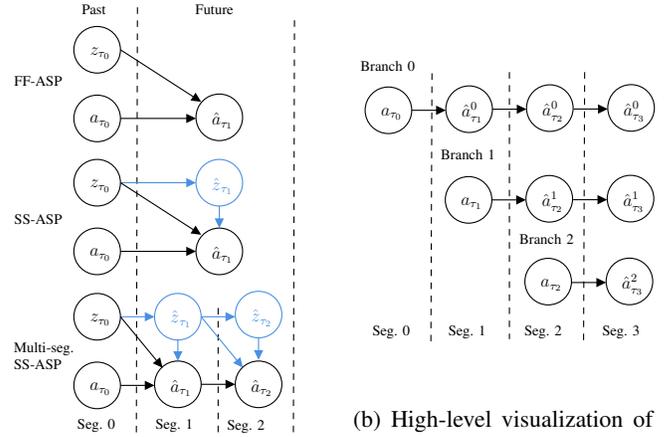

The \ac{SS-ASP} model can be extended into an autoregressive formulation with repeated calls of its components over successive time-segments. The components model the interplay between context and actions over a certain time-segment and chaining multiple calls is expected to perform reasonably well in inference. This extension is depicted in Fig.~\ref{fig:multi-segment-SS-ASP-abstract-example}. However, this naive formulation (partly presented in~\cite{janjos2021action}) actually regresses the performance due to the induced distribution drift of chaining predictions on top of predictions~\cite{janjos2021action}, see Sec.~\ref{sec:prediction-performance}. 

In addition to the distribution drift, chaining multi-modal predictions along trajectory segments is non-trivial. If the action decoder $\gamma$ generates $k$ modes per segment, a decision has to be made on which modes to expand in the next segment. If the prediction is continued for a single mode in a segment, diversity is suppressed, while considering all permutations results in $k^N$ trajectories. Therefore, different strategies for combining multi-modal predictions can be employed in order to traverse the tree and select $K$ out of possible $k^N$ modes.

In the following, six combination strategies are investigated, visualized in Fig.~\ref{fig:All-Modes-Best-Mode}. In implementing different strategies, we make use of mode probabilities. These probabilities can be either generated by the action decoder in addition to each predicted mode, or by a separate learned classification component.
Conceptually, all strategies can be placed between the All-Modes strategy, considering $k^N$ modes, and the Single-Mode strategy that selects the highest probability mode and discards others\footnote{Single-Mode can be viewed as an application of Best-first-search.}. Start-$k$ and End-$k$ strategies take $k$ modes at the start or the end, and a single mode otherwise. In Best-$m$-of-all, the product of the probabilities of previous and subsequent modes is calculated and the prediction is only continued for $m$ most likely modes\footnote{Best-$m$-of-all corresponds to the Beam search heuristic.} of all modes within a segment. The last strategy is Best-$m$-of-Prediction, where the prediction is continued for $m\leq k$ most likely modes in a single multi-modal prediction, disregarding probabilities of earlier segments. 
\newcommand{\lwd}{0.125}
\begin{figure}
	\subcaptionbox{All-Modes}
	{
		\resizebox{\lwd\linewidth}{!}{\input{images/multimodality_strategies/all_modes.tex}}
	} 
	\subcaptionbox{Single-Mode}
	{
		\resizebox{\lwd\linewidth}{!}{\input{images/multimodality_strategies/best_mode.tex}}
	} 
	\subcaptionbox{Start-$k$}
	{
		\resizebox{\lwd\linewidth}{!}{\input{images/multimodality_strategies/start_multimodal.tex}}
	} 
	\subcaptionbox{End-$k$}
	{
		\resizebox{\lwd\linewidth}{!}{\input{images/multimodality_strategies/end_multimodal.tex}}
	} 
	\subcaptionbox{Best-$m$-of-All}
	{
		\resizebox{\lwd\linewidth}{!}{\input{images/multimodality_strategies/maximum_modes.tex}}
	} 
	\subcaptionbox{Best-$m$-of-Pred.}
	{
		\resizebox{\lwd\linewidth}{!}{\input{images/multimodality_strategies/best_x_modes.tex}}
	} 
	\caption{Six investigated combination strategies for $N=3$ segments. The selected modes are depicted in blue and non-selected in gray. The number of predicted modes in a segment is $k=3$. Best-$m$-of-All uses $m=3$ and Best-$m$-of-Prediction uses $m=2$\label{fig:All-Modes-Best-Mode}.}\vspace{-14pt}
\end{figure}
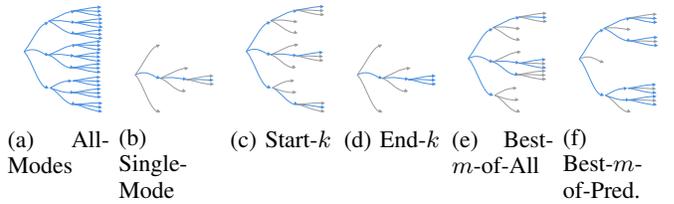

Each strategy has unique advantages and disadvantages. To quantify them, three properties are identified, see
Tab.~\ref{tab:multimodality-strategy-theoretical-properties}. The property (i) is the maximum number of multi-modal prediction calls $n$ for a sample, which serves as a proxy for the required computation time\footnote{$n$ can be larger than the number of segments $N$ in the prediction horizon, e.g. $n=13$ multi-modal prediction calls are required in the All-Modes visualization of Fig.~\ref{fig:All-Modes-Best-Mode}.}. The property (ii) is the total number of obtained modes $K$ over the entire prediction horizon. Property (iii) is qualitative and describes mode diversity via $\{$\textit{diverse}, \textit{partialy-diverse}, \textit{unclear}$\}$ qualifiers. In \textit{diverse} strategies the $K$ resulting modes share no segment trajectory subsets, in \textit{partialy-diverse} strategies the modes share a least one segment trajectory subset, and for the \textit{unclear} strategies the number of shared subsets varies per sample. The presented strategies of combining multi-modal prediction are general and can be integrated with different multi-modal decoders, i.e. do not depend on the specific action decoder used in this work.

\begin{table}[]
	\caption{Different strategies categorized by the maximum number of multi-modal prediction calls $n$ for a sample, the total number of predicted modes $K$, and the mode diversity. \label{tab:multimodality-strategy-theoretical-properties}}
	\begin{tabular}{|c|c|c|c|c|}
		\hline \multicolumn{1}{|c|}{Strategy } & \multicolumn{1}{c|}{\textbf{$K$}} & \multicolumn{1}{c|}{\textbf{$n$}} & \multicolumn{1}{c|}{Diversity} \\ \hline
		All-Modes  & $k^N$ & $\sum_{i=0}^{N-1} k^i$ & \textit{partially-diverse} \\		
		Single-Mode &  1 & $N$ & \textit{diverse}  \\
		Start-$k$ &  $k$ & $1+ k\cdot(N-1)$ & \textit{diverse} \\
		End-$k$ & $k$ & $N$ & \textit{partially-diverse} \\
		Best-$m$-of-All  & $m$ & $1 + \sum_{i=1}^{N-1} m$ & \textit{unclear} \\
		Best-$m$-of-Prediction & $m^N$ & $\sum_{i=0}^{N-1} m^i$ & \textit{partially-diverse} \\
		\hline
	\end{tabular}\vspace{-15pt}
\end{table}

\subsection{Multi-Branch SS-ASP}\label{sec:multi-branch-ss-asp}
The autoregressive formulation of the Multi-Segment \ac{SS-ASP} opens room for advanced training methods able to reduce the distribution drift of chaining multiple predictions. In the following, branched overshooting, termed in~\cite{li2021replay}, is used as well as a novel combination of context aggregation and prediction, designed to cope with the partial observability of the state through the autoregressive formulation. The resulting model aims to answer \textit{research question (i)} from Sec.~\ref{sec:intro}; it is termed as the Multi-Branch SS-ASP.

\subsubsection{Branched overshooting}\label{subsubsec:branched_overshoot}
It refers to a training method in which additional prediction branches starting from intermediate future time steps (segments) are trained in conjunction with the main branch~\cite{Overshooting, li2021replay}, visualized in Fig.~\ref{fig:prediction-branches}. A prediction branch is simply a prediction from a start segment to an end segment. For example, branch 0 refers to the full $N$ segment main branch, as used in the Multi-Segment SS-ASP, while subsequent prediction branches start from shifted time segments. This allows the model to perform $N-1$ additional, shorter predictions (of lengths $1$ to $N-1$) in addition to the $N$-segment prediction covering the entire prediction horizon. As a result, additional training is performed without adding training data. To the best of the authors' knowledge, this is the first work that applies such overshooting methods in trajectory prediction. We apply it for action prediction and action reconstruction, as well as latent context prediction.

Incorporating multiple segments as well as multiple branches into the self-supervised loss of Eq.~\eqref{eq:ssp-asp-loss} extends the loss function over time segments and branches. For example, the trajectory loss $\mathcal{L}_{traj}$ can be extended to a sum of losses per branch $b$, $\mathcal{L}_{traj} = \sum_{b=0}^{N-1} \mathcal{L}^b_{traj}$. A similar extension can be performed for the classification, context, and reconstruction components of Eq.~\eqref{eq:ssp-asp-loss}. Thus, the overall loss in Eq.~\eqref{eq:ssp-asp-loss} can be extended over branches and $\lambda_i$-weighted segments
\begin{equation}
\mathcal{L}_{MB-SS} = \textstyle  \sum_{b=0}^{N-1} \sum_{i=1}^{N-b} \lambda_i \mathcal{L}^{i, b}_{SS-ASP}\ . \label{eq:multi-branch-loss}
\end{equation}
Compared to Eq.~\eqref{eq:ssp-asp-loss}, the loss function above provides significant additional training of the model on the same data. 
\subsubsection{Combining context aggregation and prediction}\label{subsubsec:context_agg}
In autoregressive models, information from a previous recurrence step is used to predict the next step. An agent's observable state (e.g. its dynamics) does not constitute sufficient statistics for its behavior -- to address this limitation, autoregressive predictions can learn additional latent features. A natural framework for modeling such problems where partial observability occurs is the \ac{POMDP}. Furthermore, since historical behavior beyond the previous recurrence step is relevant to determining the prediction, an accumulation of `intent` over multiple recurrence steps should be possible. In this way, the future development of the scene can be modeled following Markovian assumptions.

The partial observability of intent over the entire prediction horizon can be handled through the use of recurrence over the latent context, which motivates the usage of a context aggregator component. For a segment $i$, it accumulates the encoded as well as predicted contexts of previous segments into an aggregated latent context $\bar{z}_i$, $\{z_{\tau_0},...,\hat{z}_{\tau_{i-1}}, \hat{z}_{\tau_{i}}\} \overset{\chi}{\rightarrow} \bar{z}_i$ (parameterized by $\chi$). The aggregated context~$\bar{z}_i$ can be considered as a latent context state that merges scene information of multiple consecutive segments. Thus, we use it as a stand-in for wherever predicted latent context is used, either as input to an action predictor or reconstructor component (e.g. in Fig.~\ref{fig:ssp_components} and Fig.~\ref{fig:multi-segment-SS-ASP-abstract-example}). A concept similar to incorporating recurrence in latent context prediction are the \ac{RSSM} in~\cite{Overshooting}, which include stochastic components as opposed to the fully deterministic Multi-Branch SS-ASP. 

\subsection{Prediction uncertainty estimation} \label{sec:uncertainty-estimation-methods}
In this section, we present novel prediction uncertainty estimation techniques for the deterministic Multi-Branch SS-ASP model, as well as a corresponding evaluation procedure. Since we perform prediction over time segments, we model the change in prediction error between successive segments $i$ and $i-1$ to capture the possibility of early predictions being accurate and later ones deviating significantly from the ground-truth. Specifically, we model the relative change in the \ac{minADE} for $k$ modes
\begin{equation}
	\Delta \text{minADE}_k (i, i-1) =  \text{minADE}_k (i) - \text{minADE}_k (i-1) \ , \label{eq:delta-min-ade}
\end{equation} 
and find correlations to the deterministic model uncertainty. In this sense, we aim to capture the model's changing confidence over time, and answer the \textit{research question (ii)} from Sec.~\ref{sec:intro}.

The first uncertainty estimation metric is the \textbf{reconstruction error}. This metric aims to capture disagreement between model components generating predictions and reconstructions (past 'predictions') in inference. If the trajectory prediction of a segment~$i$ significantly deviates from the reconstruction in the same segment, it indicates the epistemic uncertainty of the overall model. Specifically, we measure the deviation between reconstructed $xy$ positions $\hat{Y}_{\tau_{i}, \xi}$ of a segment $i$, obtained through the action reconstruction of the inverse model~$\xi$ (see Fig.~\ref{fig:ssp_components}), and the prediction  $\hat{Y}_{\tau_{i}, \gamma}$, obtained through the action predictor~$\gamma$ (in case of segment~$0$ the ground-truth history~$Y_{\tau_{0}}$)
\begin{equation}
\Delta_{recon}(i) = \begin{cases}
	\Vert\hat{Y}_{\tau_{i}, \xi} - \hat{Y}_{\tau_{i},\gamma}\Vert \quad &i \geq 1\ , \\
	\Vert\hat{Y}_{\tau_{0}, \xi} - Y_{\tau_{0}}\Vert \ \quad  &i = 0 \ .
\end{cases}  
\end{equation}
In this implementation, the reconstruction error is coupled to the \ac{SS-ASP} architecture due to a prerequisite for an inverse (reconstruction) model. However, adding the auxiliary task of inverse predictions to any prediction model can serve as a relatively straightforward-to-use additional regularization.

The second metric is the \textbf{mean of mode variances}. The metric is based on the application of Monte Carlo dropout in order to obtain an uncertainty estimate, naturally provided by Bayesian inference \cite{DropoutAsBayesianApproximation}. Here, we estimate the variances of predicted trajectories $\hat{Y}$ under the dropout parameter distribution $q(w|\mathcal{D}_{train})$ of weights $w$ given the training data $\mathcal{D}_{train}$. For a single $xy$ position in a mode $j$ of segment $i$, it is
	\vspace{-5pt}
\begin{gather}
	\label{eq:mode-var}
	\begin{aligned}
		\sigma_{\hat{Y}_{t,j,i}}^2 = \frac{1}{2}\left( \mathrm{Var}\left(\hat{x}_{t,j,i} \right) + \mathrm{Var}\left(\hat{y}_{t,j,i} \right)\right) \ . 
	\end{aligned}
	\vspace{-10pt}
\end{gather}
In practice, the variances are Monte-Carlo approximated by drawing multiple samples $w\sim q(w|\mathcal{D}_{train})$ with different weights dropped out \cite{DropoutAsBayesianApproximation}. In Eq.~\eqref{eq:mode-var}, we match same modes between different inference runs (by output order). We assume that variance between modes in a single run does not change significantly by applying dropout due to the inherent model determinism. The overall metric is obtained by averaging over modes and time steps (assuming independence)
\vspace{-7pt}
\begin{equation}
		\vspace{-3pt}
	\Delta_{mode-var}(i) =  \textstyle\sum_j^k \textstyle\sum_t^T \frac{\sigma_{\hat{Y}_{t,j,i}}^2}{kT} \label{eq:mean-of-mode-variances} \ . 
\end{equation}
The metric in Eq.~\eqref{eq:mean-of-mode-variances} serves as an estimate of the covariance within a Bayesian network model whose weights are Gaussian distributed, which is theoretically grounded in~\cite{DropoutAsBayesianApproximation}. Therefore, it serves as an indicator of the epistemic model uncertainty.

%% file: images/ssp_components.tex
\tikzset{every picture/.style={line width=0.75pt}} 

\begin{tikzpicture}[x=0.75pt,y=0.75pt,yscale=-1,xscale=1]
	
	\draw   (44,33.33) .. controls (44,21.53) and (53.57,11.97) .. (65.37,11.97) .. controls (77.17,11.97) and (86.73,21.53) .. (86.73,33.33) .. controls (86.73,45.13) and (77.17,54.7) .. (65.37,54.7) .. controls (53.57,54.7) and (44,45.13) .. (44,33.33) -- cycle ;
	\draw    (5.33,33.5) -- (41,33.35) ;
	\draw [shift={(44,33.33)}, rotate = 179.75] [fill={rgb, 255:red, 0; green, 0; blue, 0 }  ][line width=0.08]  [draw opacity=0] (8.93,-4.29) -- (0,0) -- (8.93,4.29) -- cycle    ;
	\draw    (86.73,33.33) -- (109.72,33.23) -- (132.3,32.94) -- (131.93,73.94) ;
	\draw [shift={(131.9,76.94)}, rotate = 270.52] [fill={rgb, 255:red, 0; green, 0; blue, 0 }  ][line width=0.08]  [draw opacity=0] (8.93,-4.29) -- (0,0) -- (8.93,4.29) -- cycle    ;
	\draw   (44,157.33) .. controls (44,145.53) and (53.57,135.97) .. (65.37,135.97) .. controls (77.17,135.97) and (86.73,145.53) .. (86.73,157.33) .. controls (86.73,169.13) and (77.17,178.7) .. (65.37,178.7) .. controls (53.57,178.7) and (44,169.13) .. (44,157.33) -- cycle ;
	\draw    (5.33,157.5) -- (41,157.35) ;
	\draw [shift={(44,157.33)}, rotate = 179.75] [fill={rgb, 255:red, 0; green, 0; blue, 0 }  ][line width=0.08]  [draw opacity=0] (8.93,-4.29) -- (0,0) -- (8.93,4.29) -- cycle    ;
	\draw    (86.73,157.33) -- (109.72,157.23) -- (133.37,157.47) -- (133.49,116.74) ;
	\draw [shift={(133.5,113.74)}, rotate = 90.17] [fill={rgb, 255:red, 0; green, 0; blue, 0 }  ][line width=0.08]  [draw opacity=0] (8.93,-4.29) -- (0,0) -- (8.93,4.29) -- cycle    ;
	\draw   (113,85.43) .. controls (113,81.44) and (116.24,78.2) .. (120.23,78.2) -- (146.51,78.2) .. controls (150.5,78.2) and (153.73,81.44) .. (153.73,85.43) -- (153.73,107.11) .. controls (153.73,111.1) and (150.5,114.33) .. (146.51,114.33) -- (120.23,114.33) .. controls (116.24,114.33) and (113,111.1) .. (113,107.11) -- cycle ;
	\draw    (154.13,97.1) -- (189.8,96.95) ;
	\draw [shift={(192.8,96.93)}, rotate = 179.75] [fill={rgb, 255:red, 0; green, 0; blue, 0 }  ][line width=0.08]  [draw opacity=0] (8.93,-4.29) -- (0,0) -- (8.93,4.29) -- cycle    ;
	\draw   (264.4,33.53) .. controls (264.4,21.73) and (273.97,12.17) .. (285.77,12.17) .. controls (297.57,12.17) and (307.13,21.73) .. (307.13,33.53) .. controls (307.13,45.33) and (297.57,54.9) .. (285.77,54.9) .. controls (273.97,54.9) and (264.4,45.33) .. (264.4,33.53) -- cycle ;
	\draw    (225.73,33.7) -- (261.4,33.55) ;
	\draw [shift={(264.4,33.53)}, rotate = 179.75] [fill={rgb, 255:red, 0; green, 0; blue, 0 }  ][line width=0.08]  [draw opacity=0] (8.93,-4.29) -- (0,0) -- (8.93,4.29) -- cycle    ;
	\draw [color={rgb, 255:red, 74; green, 144; blue, 226 }  ,draw opacity=1 ]   (352.7,33.14) -- (352.7,33.14) -- (352.33,74.14) ;
	\draw [shift={(352.3,77.14)}, rotate = 270.52] [fill={rgb, 255:red, 74; green, 144; blue, 226 }  ,fill opacity=1 ][line width=0.08]  [draw opacity=0] (8.93,-4.29) -- (0,0) -- (8.93,4.29) -- cycle    ;
	\draw   (264.4,157.53) .. controls (264.4,145.73) and (273.97,136.17) .. (285.77,136.17) .. controls (297.57,136.17) and (307.13,145.73) .. (307.13,157.53) .. controls (307.13,169.33) and (297.57,178.9) .. (285.77,178.9) .. controls (273.97,178.9) and (264.4,169.33) .. (264.4,157.53) -- cycle ;
	\draw    (225.73,157.7) -- (261.4,157.55) ;
	\draw [shift={(264.4,157.53)}, rotate = 179.75] [fill={rgb, 255:red, 0; green, 0; blue, 0 }  ][line width=0.08]  [draw opacity=0] (8.93,-4.29) -- (0,0) -- (8.93,4.29) -- cycle    ;
	\draw [color={rgb, 255:red, 74; green, 144; blue, 226 }  ,draw opacity=1 ]   (353.77,157.67) -- (353.77,157.67) -- (353.89,116.94) ;
	\draw [shift={(353.9,113.94)}, rotate = 90.17] [fill={rgb, 255:red, 74; green, 144; blue, 226 }  ,fill opacity=1 ][line width=0.08]  [draw opacity=0] (8.93,-4.29) -- (0,0) -- (8.93,4.29) -- cycle    ;
	\draw  [color={rgb, 255:red, 74; green, 144; blue, 226 }  ,draw opacity=1 ] (333.4,85.63) .. controls (333.4,81.64) and (336.64,78.4) .. (340.63,78.4) -- (366.91,78.4) .. controls (370.9,78.4) and (374.13,81.64) .. (374.13,85.63) -- (374.13,107.31) .. controls (374.13,111.3) and (370.9,114.53) .. (366.91,114.53) -- (340.63,114.53) .. controls (336.64,114.53) and (333.4,111.3) .. (333.4,107.31) -- cycle ;
	\draw [color={rgb, 255:red, 74; green, 144; blue, 226 }  ,draw opacity=1 ]   (374.53,97.3) -- (427.2,97.14) ;
	\draw [shift={(430.2,97.13)}, rotate = 179.83] [fill={rgb, 255:red, 74; green, 144; blue, 226 }  ,fill opacity=1 ][line width=0.08]  [draw opacity=0] (8.93,-4.29) -- (0,0) -- (8.93,4.29) -- cycle    ;
	\draw    (353.77,157.67) -- (376.76,157.57) -- (449.4,157.8) -- (449.52,117.07) ;
	\draw [shift={(449.53,114.07)}, rotate = 90.17] [fill={rgb, 255:red, 0; green, 0; blue, 0 }  ][line width=0.08]  [draw opacity=0] (8.93,-4.29) -- (0,0) -- (8.93,4.29) -- cycle    ;
	\draw    (352.7,33.14) -- (375.69,33.04) -- (449.27,32.75) -- (448.89,73.75) ;
	\draw [shift={(448.87,76.75)}, rotate = 270.52] [fill={rgb, 255:red, 0; green, 0; blue, 0 }  ][line width=0.08]  [draw opacity=0] (8.93,-4.29) -- (0,0) -- (8.93,4.29) -- cycle    ;
	\draw   (430,85.43) .. controls (430,81.44) and (433.24,78.2) .. (437.23,78.2) -- (463.51,78.2) .. controls (467.5,78.2) and (470.73,81.44) .. (470.73,85.43) -- (470.73,107.11) .. controls (470.73,111.1) and (467.5,114.33) .. (463.51,114.33) -- (437.23,114.33) .. controls (433.24,114.33) and (430,111.1) .. (430,107.11) -- cycle ;
	\draw    (471.13,97.1) -- (526.8,96.94) ;
	\draw [shift={(529.8,96.93)}, rotate = 179.84] [fill={rgb, 255:red, 0; green, 0; blue, 0 }  ][line width=0.08]  [draw opacity=0] (8.93,-4.29) -- (0,0) -- (8.93,4.29) -- cycle    ;
	\draw  [color={rgb, 255:red, 74; green, 144; blue, 226 }  ,draw opacity=1 ] (265.4,219.53) .. controls (265.4,207.73) and (274.97,198.17) .. (286.77,198.17) .. controls (298.57,198.17) and (308.13,207.73) .. (308.13,219.53) .. controls (308.13,231.33) and (298.57,240.9) .. (286.77,240.9) .. controls (274.97,240.9) and (265.4,231.33) .. (265.4,219.53) -- cycle ;
	\draw [color={rgb, 255:red, 74; green, 144; blue, 226 }  ,draw opacity=1 ]   (226.73,219.7) -- (262.4,219.55) ;
	\draw [shift={(265.4,219.53)}, rotate = 179.75] [fill={rgb, 255:red, 74; green, 144; blue, 226 }  ,fill opacity=1 ][line width=0.08]  [draw opacity=0] (8.93,-4.29) -- (0,0) -- (8.93,4.29) -- cycle    ;
	\draw [color={rgb, 255:red, 74; green, 144; blue, 226 }  ,draw opacity=1 ]   (308.13,219.53) -- (331.12,219.43) -- (354.77,219.67) -- (425.61,219.31) ;
	\draw [shift={(428.61,219.3)}, rotate = 179.72] [fill={rgb, 255:red, 74; green, 144; blue, 226 }  ,fill opacity=1 ][line width=0.08]  [draw opacity=0] (8.93,-4.29) -- (0,0) -- (8.93,4.29) -- cycle    ;
	\draw  [color={rgb, 255:red, 74; green, 144; blue, 226 }  ,draw opacity=1 ] (428.33,208.89) .. controls (428.33,204.9) and (431.57,201.67) .. (435.56,201.67) -- (461.84,201.67) .. controls (465.83,201.67) and (469.07,204.9) .. (469.07,208.89) -- (469.07,230.57) .. controls (469.07,234.56) and (465.83,237.8) .. (461.84,237.8) -- (435.56,237.8) .. controls (431.57,237.8) and (428.33,234.56) .. (428.33,230.57) -- cycle ;
	\draw [color={rgb, 255:red, 74; green, 144; blue, 226 }  ,draw opacity=1 ]   (449,238) -- (449,256) ;
	\draw [shift={(449,259)}, rotate = 270] [fill={rgb, 255:red, 74; green, 144; blue, 226 }  ,fill opacity=1 ][line width=0.08]  [draw opacity=0] (8.93,-4.29) -- (0,0) -- (8.93,4.29) -- cycle    ;
	\draw    (307.13,33.53) -- (352.7,33.14) ;
	\draw    (307.13,157.53) -- (353.77,157.67) ;
	\draw  [dash pattern={on 4.5pt off 4.5pt}]  (207.6,2.73) -- (208.6,287.23) ;
	\draw [color={rgb, 255:red, 74; green, 144; blue, 226 }  ,draw opacity=1 ]   (449.4,157.8) -- (449.4,157.8) -- (449.03,198.8) ;
	\draw [shift={(449,201.8)}, rotate = 270.52] [fill={rgb, 255:red, 74; green, 144; blue, 226 }  ,fill opacity=1 ][line width=0.08]  [draw opacity=0] (8.93,-4.29) -- (0,0) -- (8.93,4.29) -- cycle    ;
	
	\draw (44,36) node [anchor=north west][inner sep=0.75pt]   [align=left] { };
	\draw (23.23,18.88) node  [font=\large]  {${\textstyle a_{\tau _{0}}}$};
	\draw (103.23,18.88) node  [font=\large]  {${\displaystyle \alpha_{\tau _{0}}}$};
	\draw (65.37,33.33) node  [font=\large]  {$\alpha $};
	\draw (42,60.67) node [anchor=north west][inner sep=0.75pt]   [align=left] {Action\\encoder};
	\draw (44,160) node [anchor=north west][inner sep=0.75pt]   [align=left] { };
	\draw (23.23,142.88) node  [font=\large]  {${\textstyle o_{\tau _{0}}}$};
	\draw (103.23,142.88) node  [font=\large]  {${\displaystyle z_{\tau _{0}}}$};
	\draw (65.37,157.33) node  [font=\large]  {$\phi $};
	\draw (42,184.67) node [anchor=north west][inner sep=0.75pt]   [align=left] {Context\\encoder};
	\draw (133.37,96.27) node  [font=\large]  {$\gamma $};
	\draw (146.2,120.47) node [anchor=north west][inner sep=0.75pt]   [align=left] {Action\\decoder};
	\draw (174.83,82.88) node  [font=\large]  {${\displaystyle \hat{{\displaystyle a}}_{\tau _{1}}}$};
	\draw (264.4,36.2) node [anchor=north west][inner sep=0.75pt]   [align=left] { };
	\draw (243.63,19.08) node  [font=\large]  {${\textstyle a_{\tau _{0}}}$};
	\draw (323.63,19.08) node  [font=\large]  {${\displaystyle a_{\tau _{0}}}$};
	\draw (285.77,33.53) node  [font=\large]  {$\alpha $};
	\draw (262.4,60.87) node [anchor=north west][inner sep=0.75pt]   [align=left] {Action\\encoder};
	\draw (264.4,160.2) node [anchor=north west][inner sep=0.75pt]   [align=left] { };
	\draw (243.63,143.08) node  [font=\large]  {${\textstyle o_{\tau _{0}}}$};
	\draw (323.63,143.08) node  [font=\large]  {${\displaystyle z_{\tau _{0}}}$};
	\draw (285.77,157.53) node  [font=\large]  {$\phi $};
	\draw (262.4,251.37) node [anchor=north west][inner sep=0.75pt]  [color={rgb, 255:red, 74; green, 144; blue, 226 }  ,opacity=1 ] [align=left] {Context\\encoder};
	\draw (353.77,96.47) node  [font=\large,color={rgb, 255:red, 74; green, 144; blue, 226 }  ,opacity=1 ]  {$\psi $};
	\draw (366.6,120.17) node [anchor=north west][inner sep=0.75pt]  [color={rgb, 255:red, 74; green, 144; blue, 226 }  ,opacity=1 ] [align=left] {Context\\predictor};
	\draw (395.23,82.58) node  [font=\large,color={rgb, 255:red, 74; green, 144; blue, 226 }  ,opacity=1 ]  {${\displaystyle \hat{{\displaystyle z}}_{\tau _{1}}}$};
	\draw (463.2,120.47) node [anchor=north west][inner sep=0.75pt]   [align=left] {Action\\decoder};
	\draw (491.83,82.88) node  [font=\large]  {${\displaystyle \hat{{\displaystyle a}}_{\tau _{1}}}$};
	\draw (450.37,96.27) node  [font=\large]  {$\gamma $};
	\draw (265.4,222.2) node [anchor=north west][inner sep=0.75pt]   [align=left] { };
	\draw (244.63,204.58) node  [font=\large,color={rgb, 255:red, 74; green, 144; blue, 226 }  ,opacity=1 ]  {${\textstyle o_{\tau _{1}}}$};
	\draw (324.63,204.58) node  [font=\large,color={rgb, 255:red, 74; green, 144; blue, 226 }  ,opacity=1 ]  {${\displaystyle z_{\tau _{1}}}$};
	\draw (286.77,219.03) node  [font=\large,color={rgb, 255:red, 74; green, 144; blue, 226 }  ,opacity=1 ]  {$\phi $};
	\draw (448.7,219.23) node  [font=\large,color={rgb, 255:red, 74; green, 144; blue, 226 }  ,opacity=1 ]  {$\xi $};
	\draw (474,245.3) node [anchor=north west][inner sep=0.75pt]  [color={rgb, 255:red, 74; green, 144; blue, 226 }  ,opacity=1 ] [align=left] {Action\\reconstructor};
	\draw (454.57,273.18) node  [font=\large,color={rgb, 255:red, 74; green, 144; blue, 226 }  ,opacity=1 ]  {$\hat{a}_{\tau _{0}}$};

\end{tikzpicture}

%% file: mutli_segment_ss_asp_vs_ss_asp.tex
\tikzset{every picture/.style={line width=0.75pt}} 

\begin{tikzpicture}[x=0.75pt,y=0.75pt,yscale=-1,xscale=1]
	
	\draw   (60,101.33) .. controls (60,89.53) and (69.57,79.97) .. (81.37,79.97) .. controls (93.17,79.97) and (102.73,89.53) .. (102.73,101.33) .. controls (102.73,113.13) and (93.17,122.7) .. (81.37,122.7) .. controls (69.57,122.7) and (60,113.13) .. (60,101.33) -- cycle ;
	\draw   (60,38.33) .. controls (60,26.53) and (69.57,16.97) .. (81.37,16.97) .. controls (93.17,16.97) and (102.73,26.53) .. (102.73,38.33) .. controls (102.73,50.13) and (93.17,59.7) .. (81.37,59.7) .. controls (69.57,59.7) and (60,50.13) .. (60,38.33) -- cycle ;
	\draw   (172,100.33) .. controls (172,88.53) and (181.57,78.97) .. (193.37,78.97) .. controls (205.17,78.97) and (214.73,88.53) .. (214.73,100.33) .. controls (214.73,112.13) and (205.17,121.7) .. (193.37,121.7) .. controls (181.57,121.7) and (172,112.13) .. (172,100.33) -- cycle ;
	\draw   (60,223.33) .. controls (60,211.53) and (69.57,201.97) .. (81.37,201.97) .. controls (93.17,201.97) and (102.73,211.53) .. (102.73,223.33) .. controls (102.73,235.13) and (93.17,244.7) .. (81.37,244.7) .. controls (69.57,244.7) and (60,235.13) .. (60,223.33) -- cycle ;
	\draw   (60,160.33) .. controls (60,148.53) and (69.57,138.97) .. (81.37,138.97) .. controls (93.17,138.97) and (102.73,148.53) .. (102.73,160.33) .. controls (102.73,172.13) and (93.17,181.7) .. (81.37,181.7) .. controls (69.57,181.7) and (60,172.13) .. (60,160.33) -- cycle ;
	\draw   (60,346.33) .. controls (60,334.53) and (69.57,324.97) .. (81.37,324.97) .. controls (93.17,324.97) and (102.73,334.53) .. (102.73,346.33) .. controls (102.73,358.13) and (93.17,367.7) .. (81.37,367.7) .. controls (69.57,367.7) and (60,358.13) .. (60,346.33) -- cycle ;
	\draw   (60,283.33) .. controls (60,271.53) and (69.57,261.97) .. (81.37,261.97) .. controls (93.17,261.97) and (102.73,271.53) .. (102.73,283.33) .. controls (102.73,295.13) and (93.17,304.7) .. (81.37,304.7) .. controls (69.57,304.7) and (60,295.13) .. (60,283.33) -- cycle ;
	\draw   (134,345.33) .. controls (134,333.53) and (143.57,323.97) .. (155.37,323.97) .. controls (167.17,323.97) and (176.73,333.53) .. (176.73,345.33) .. controls (176.73,357.13) and (167.17,366.7) .. (155.37,366.7) .. controls (143.57,366.7) and (134,357.13) .. (134,345.33) -- cycle ;
	\draw  [color={rgb, 255:red, 74; green, 144; blue, 226 }  ,draw opacity=1 ] (134,283.33) .. controls (134,271.53) and (143.57,261.97) .. (155.37,261.97) .. controls (167.17,261.97) and (176.73,271.53) .. (176.73,283.33) .. controls (176.73,295.13) and (167.17,304.7) .. (155.37,304.7) .. controls (143.57,304.7) and (134,295.13) .. (134,283.33) -- cycle ;
	\draw   (208,345.33) .. controls (208,333.53) and (217.57,323.97) .. (229.37,323.97) .. controls (241.17,323.97) and (250.73,333.53) .. (250.73,345.33) .. controls (250.73,357.13) and (241.17,366.7) .. (229.37,366.7) .. controls (217.57,366.7) and (208,357.13) .. (208,345.33) -- cycle ;
	\draw  [color={rgb, 255:red, 74; green, 144; blue, 226 }  ,draw opacity=1 ] (208,281.33) .. controls (208,269.53) and (217.57,259.97) .. (229.37,259.97) .. controls (241.17,259.97) and (250.73,269.53) .. (250.73,281.33) .. controls (250.73,293.13) and (241.17,302.7) .. (229.37,302.7) .. controls (217.57,302.7) and (208,293.13) .. (208,281.33) -- cycle ;
	\draw   (172,223.33) .. controls (172,211.53) and (181.57,201.97) .. (193.37,201.97) .. controls (205.17,201.97) and (214.73,211.53) .. (214.73,223.33) .. controls (214.73,235.13) and (205.17,244.7) .. (193.37,244.7) .. controls (181.57,244.7) and (172,235.13) .. (172,223.33) -- cycle ;
	\draw  [color={rgb, 255:red, 74; green, 144; blue, 226 }  ,draw opacity=1 ] (172,160.33) .. controls (172,148.53) and (181.57,138.97) .. (193.37,138.97) .. controls (205.17,138.97) and (214.73,148.53) .. (214.73,160.33) .. controls (214.73,172.13) and (205.17,181.7) .. (193.37,181.7) .. controls (181.57,181.7) and (172,172.13) .. (172,160.33) -- cycle ;
	\draw  [dash pattern={on 4.5pt off 4.5pt}]  (118.3,-0.6) -- (117.3,393.4) ;
	\draw  [dash pattern={on 4.5pt off 4.5pt}]  (262.3,11.3) -- (261.3,392.3) ;
	\draw  [dash pattern={on 4.5pt off 4.5pt}]  (192.3,275.3) -- (192.3,394.3) ;
	\draw    (102.73,38.33) -- (174.26,84.48) ;
	\draw [shift={(176.78,86.11)}, rotate = 212.83] [fill={rgb, 255:red, 0; green, 0; blue, 0 }  ][line width=0.08]  [draw opacity=0] (8.93,-4.29) -- (0,0) -- (8.93,4.29) -- cycle    ;
	\draw    (102.73,101.33) -- (169,101.33) ;
	\draw [shift={(172,101.33)}, rotate = 180] [fill={rgb, 255:red, 0; green, 0; blue, 0 }  ][line width=0.08]  [draw opacity=0] (8.93,-4.29) -- (0,0) -- (8.93,4.29) -- cycle    ;
	\draw    (102.73,223.33) -- (169,223.33) ;
	\draw [shift={(172,223.33)}, rotate = 180] [fill={rgb, 255:red, 0; green, 0; blue, 0 }  ][line width=0.08]  [draw opacity=0] (8.93,-4.29) -- (0,0) -- (8.93,4.29) -- cycle    ;
	\draw [color={rgb, 255:red, 74; green, 144; blue, 226 }  ,draw opacity=1 ]   (102.73,160.33) -- (169,160.33) ;
	\draw [shift={(172,160.33)}, rotate = 180] [fill={rgb, 255:red, 74; green, 144; blue, 226 }  ,fill opacity=1 ][line width=0.08]  [draw opacity=0] (8.93,-4.29) -- (0,0) -- (8.93,4.29) -- cycle    ;
	\draw [color={rgb, 255:red, 74; green, 144; blue, 226 }  ,draw opacity=1 ]   (193.37,181.7) -- (193.37,198.97) ;
	\draw [shift={(193.37,201.97)}, rotate = 270] [fill={rgb, 255:red, 74; green, 144; blue, 226 }  ,fill opacity=1 ][line width=0.08]  [draw opacity=0] (8.93,-4.29) -- (0,0) -- (8.93,4.29) -- cycle    ;
	\draw    (102.73,346.33) -- (131,346.33) ;
	\draw [shift={(134,346.33)}, rotate = 180] [fill={rgb, 255:red, 0; green, 0; blue, 0 }  ][line width=0.08]  [draw opacity=0] (8.93,-4.29) -- (0,0) -- (8.93,4.29) -- cycle    ;
	\draw [color={rgb, 255:red, 74; green, 144; blue, 226 }  ,draw opacity=1 ]   (102.73,283.33) -- (131,283.33) ;
	\draw [shift={(134,283.33)}, rotate = 180] [fill={rgb, 255:red, 74; green, 144; blue, 226 }  ,fill opacity=1 ][line width=0.08]  [draw opacity=0] (8.93,-4.29) -- (0,0) -- (8.93,4.29) -- cycle    ;
	\draw [color={rgb, 255:red, 74; green, 144; blue, 226 }  ,draw opacity=1 ]   (176.73,283.33) -- (205,283.33) ;
	\draw [shift={(208,283.33)}, rotate = 180] [fill={rgb, 255:red, 74; green, 144; blue, 226 }  ,fill opacity=1 ][line width=0.08]  [draw opacity=0] (8.93,-4.29) -- (0,0) -- (8.93,4.29) -- cycle    ;
	\draw    (176.73,345.33) -- (205,345.33) ;
	\draw [shift={(208,345.33)}, rotate = 180] [fill={rgb, 255:red, 0; green, 0; blue, 0 }  ][line width=0.08]  [draw opacity=0] (8.93,-4.29) -- (0,0) -- (8.93,4.29) -- cycle    ;
	\draw [color={rgb, 255:red, 74; green, 144; blue, 226 }  ,draw opacity=1 ]   (229.37,302.7) -- (229.37,319.97) ;
	\draw [shift={(229.37,322.97)}, rotate = 270] [fill={rgb, 255:red, 74; green, 144; blue, 226 }  ,fill opacity=1 ][line width=0.08]  [draw opacity=0] (8.93,-4.29) -- (0,0) -- (8.93,4.29) -- cycle    ;
	\draw [color={rgb, 255:red, 74; green, 144; blue, 226 }  ,draw opacity=1 ]   (155.37,303.7) -- (155.37,320.97) ;
	\draw [shift={(155.37,323.97)}, rotate = 270] [fill={rgb, 255:red, 74; green, 144; blue, 226 }  ,fill opacity=1 ][line width=0.08]  [draw opacity=0] (8.93,-4.29) -- (0,0) -- (8.93,4.29) -- cycle    ;
	\draw [color={rgb, 255:red, 0; green, 0; blue, 0 }  ,draw opacity=1 ]   (102.73,283.33) -- (139.49,327.89) ;
	\draw [shift={(141.4,330.2)}, rotate = 230.48] [fill={rgb, 255:red, 0; green, 0; blue, 0 }  ,fill opacity=1 ][line width=0.08]  [draw opacity=0] (8.93,-4.29) -- (0,0) -- (8.93,4.29) -- cycle    ;
	\draw [color={rgb, 255:red, 74; green, 144; blue, 226 }  ,draw opacity=1 ]   (176.73,283.33) -- (213.1,327.68) ;
	\draw [shift={(215,330)}, rotate = 230.65] [fill={rgb, 255:red, 74; green, 144; blue, 226 }  ,fill opacity=1 ][line width=0.08]  [draw opacity=0] (8.93,-4.29) -- (0,0) -- (8.93,4.29) -- cycle    ;
	\draw [color={rgb, 255:red, 0; green, 0; blue, 0 }  ,draw opacity=1 ]   (102.73,160.33) -- (174.26,206.48) ;
	\draw [shift={(176.78,208.11)}, rotate = 212.83] [fill={rgb, 255:red, 0; green, 0; blue, 0 }  ,fill opacity=1 ][line width=0.08]  [draw opacity=0] (8.93,-4.29) -- (0,0) -- (8.93,4.29) -- cycle    ;
	
	\draw (60,104) node [anchor=north west][inner sep=0.75pt]   [align=left] { };
	\draw (84.37,103.93) node  [font=\large]  {${\displaystyle a_{\tau _{0}}}$};
	\draw (60,41) node [anchor=north west][inner sep=0.75pt]   [align=left] { };
	\draw (84.37,41.33) node  [font=\large]  {${\displaystyle z_{\tau _{0}}}$};
	\draw (172,103) node [anchor=north west][inner sep=0.75pt]   [align=left] { };
	\draw (196.37,103.33) node  [font=\large]  {${\displaystyle \hat{{\displaystyle a}}_{\tau _{1}}}$};
	\draw (51,248) node [anchor=north west][inner sep=0.75pt]   [align=left] { };
	\draw (84.37,226.33) node  [font=\large]  {${\displaystyle a_{\tau _{0}}}$};
	\draw (60,163) node [anchor=north west][inner sep=0.75pt]   [align=left] { };
	\draw (84.37,163.33) node  [font=\large]  {${\displaystyle z_{\tau _{0}}}$};
	\draw (60,349) node [anchor=north west][inner sep=0.75pt]   [align=left] { };
	\draw (84.37,349.33) node  [font=\large]  {${\displaystyle a_{\tau _{0}}}$};
	\draw (60,286) node [anchor=north west][inner sep=0.75pt]   [align=left] { };
	\draw (84.37,286.33) node  [font=\large]  {${\displaystyle z_{\tau _{0}}}$};
	\draw (134,370) node [anchor=north west][inner sep=0.75pt]   [align=left] { };
	\draw (158.37,348.33) node  [font=\large]  {${\displaystyle \hat{{\displaystyle a}}_{\tau _{1}}}$};
	\draw (134,285) node [anchor=north west][inner sep=0.75pt]   [align=left] { };
	\draw (158.37,285.33) node  [font=\large,color={rgb, 255:red, 74; green, 144; blue, 226 }  ,opacity=1 ]  {${\displaystyle \hat{{\displaystyle z}}_{\tau _{1}}}$};
	\draw (209,348) node [anchor=north west][inner sep=0.75pt]   [align=left] { };
	\draw (232.37,348.33) node  [font=\large]  {${\displaystyle \hat{{\displaystyle a}}_{\tau _{2}}}$};
	\draw (209,306) node [anchor=north west][inner sep=0.75pt]   [align=left] { };
	\draw (232.37,284.33) node  [font=\large,color={rgb, 255:red, 74; green, 144; blue, 226 }  ,opacity=1 ]  {${\displaystyle \hat{{\displaystyle z}}_{\tau _{2}}}$};
	\draw (196.37,226.33) node  [font=\large]  {${\displaystyle \hat{{\displaystyle a}}_{\tau _{1}}}$};
	\draw (172,176.6) node [anchor=north west][inner sep=0.75pt]   [align=left] { };
	\draw (196.37,163.33) node  [font=\large,color={rgb, 255:red, 74; green, 144; blue, 226 }  ,opacity=1 ]  {${\displaystyle \hat{{\displaystyle z}}_{\tau _{1}}}$};
	\draw (4,61) node [anchor=north west][inner sep=0.75pt]   [align=left] {FF-ASP};
	\draw (4,188) node [anchor=north west][inner sep=0.75pt]   [align=left] {SS-ASP};
	\draw (3.6,305) node [anchor=north west][inner sep=0.75pt]   [align=left] {Multi-seg.\\SS-ASP};
	\draw (66,-5) node [anchor=north west][inner sep=0.75pt]   [align=left] {Past};
	\draw (170,-4.5) node [anchor=north west][inner sep=0.75pt]   [align=left] {Future};
	\draw (133.4,376) node [anchor=north west][inner sep=0.75pt]   [align=left] {Seg. 1};
	\draw (199,376.3) node [anchor=north west][inner sep=0.75pt]   [align=left] {Seg. 2};
	\draw (61.8,375.9) node [anchor=north west][inner sep=0.75pt]   [align=left] {Seg. 0};

\end{tikzpicture}

%% file: multi_branch_ss_asp_abstract_latent_context_view.tex
\tikzset{every picture/.style={line width=0.75pt}} 

\begin{tikzpicture}[x=0.75pt,y=0.75pt,yscale=-1,xscale=1]
	
	\draw   (14,50.33) .. controls (14,38.53) and (23.57,28.97) .. (35.37,28.97) .. controls (47.17,28.97) and (56.73,38.53) .. (56.73,50.33) .. controls (56.73,62.13) and (47.17,71.7) .. (35.37,71.7) .. controls (23.57,71.7) and (14,62.13) .. (14,50.33) -- cycle ;
	\draw  [color={rgb, 255:red, 0; green, 0; blue, 0 }  ,draw opacity=1 ] (88,49.33) .. controls (88,37.53) and (97.57,27.97) .. (109.37,27.97) .. controls (121.17,27.97) and (130.73,37.53) .. (130.73,49.33) .. controls (130.73,61.13) and (121.17,70.7) .. (109.37,70.7) .. controls (97.57,70.7) and (88,61.13) .. (88,49.33) -- cycle ;
	\draw  [color={rgb, 255:red, 0; green, 0; blue, 0 }  ,draw opacity=1 ] (162,48.33) .. controls (162,36.53) and (171.57,26.97) .. (183.37,26.97) .. controls (195.17,26.97) and (204.73,36.53) .. (204.73,48.33) .. controls (204.73,60.13) and (195.17,69.7) .. (183.37,69.7) .. controls (171.57,69.7) and (162,60.13) .. (162,48.33) -- cycle ;
	\draw  [dash pattern={on 4.5pt off 4.5pt}]  (75.3,16.4) -- (74.3,264.4) ;
	\draw  [dash pattern={on 4.5pt off 4.5pt}]  (215.3,15.3) -- (214.3,265.3) ;
	\draw  [dash pattern={on 4.5pt off 4.5pt}]  (146.3,15.3) -- (146.3,195) -- (146.3,260.3) ;
	\draw [color={rgb, 255:red, 0; green, 0; blue, 0 }  ,draw opacity=1 ]   (56.73,50.33) -- (85,50.33) ;
	\draw [shift={(88,50.33)}, rotate = 180] [fill={rgb, 255:red, 0; green, 0; blue, 0 }  ,fill opacity=1 ][line width=0.08]  [draw opacity=0] (8.93,-4.29) -- (0,0) -- (8.93,4.29) -- cycle    ;
	\draw [color={rgb, 255:red, 0; green, 0; blue, 0 }  ,draw opacity=1 ]   (130.73,49.33) -- (159,49.33) ;
	\draw [shift={(162,49.33)}, rotate = 180] [fill={rgb, 255:red, 0; green, 0; blue, 0 }  ,fill opacity=1 ][line width=0.08]  [draw opacity=0] (8.93,-4.29) -- (0,0) -- (8.93,4.29) -- cycle    ;
	\draw  [color={rgb, 255:red, 0; green, 0; blue, 0 }  ,draw opacity=1 ] (236.5,48.17) .. controls (236.5,36.37) and (246.07,26.8) .. (257.87,26.8) .. controls (269.67,26.8) and (279.23,36.37) .. (279.23,48.17) .. controls (279.23,59.97) and (269.67,69.53) .. (257.87,69.53) .. controls (246.07,69.53) and (236.5,59.97) .. (236.5,48.17) -- cycle ;
	\draw [color={rgb, 255:red, 0; green, 0; blue, 0 }  ,draw opacity=1 ]   (205.23,49.17) -- (233.5,49.17) ;
	\draw [shift={(236.5,49.17)}, rotate = 180] [fill={rgb, 255:red, 0; green, 0; blue, 0 }  ,fill opacity=1 ][line width=0.08]  [draw opacity=0] (8.93,-4.29) -- (0,0) -- (8.93,4.29) -- cycle    ;
	\draw  [color={rgb, 255:red, 0; green, 0; blue, 0 }  ,draw opacity=1 ] (87.5,132.57) .. controls (87.5,120.77) and (97.07,111.2) .. (108.87,111.2) .. controls (120.67,111.2) and (130.23,120.77) .. (130.23,132.57) .. controls (130.23,144.37) and (120.67,153.93) .. (108.87,153.93) .. controls (97.07,153.93) and (87.5,144.37) .. (87.5,132.57) -- cycle ;
	\draw  [color={rgb, 255:red, 0; green, 0; blue, 0 }  ,draw opacity=1 ] (161.5,131.57) .. controls (161.5,119.77) and (171.07,110.2) .. (182.87,110.2) .. controls (194.67,110.2) and (204.23,119.77) .. (204.23,131.57) .. controls (204.23,143.37) and (194.67,152.93) .. (182.87,152.93) .. controls (171.07,152.93) and (161.5,143.37) .. (161.5,131.57) -- cycle ;
	\draw [color={rgb, 255:red, 0; green, 0; blue, 0 }  ,draw opacity=1 ]   (130.23,132.57) -- (158.5,132.57) ;
	\draw [shift={(161.5,132.57)}, rotate = 180] [fill={rgb, 255:red, 0; green, 0; blue, 0 }  ,fill opacity=1 ][line width=0.08]  [draw opacity=0] (8.93,-4.29) -- (0,0) -- (8.93,4.29) -- cycle    ;
	\draw  [color={rgb, 255:red, 0; green, 0; blue, 0 }  ,draw opacity=1 ] (236,131.4) .. controls (236,119.6) and (245.57,110.03) .. (257.37,110.03) .. controls (269.17,110.03) and (278.73,119.6) .. (278.73,131.4) .. controls (278.73,143.2) and (269.17,152.77) .. (257.37,152.77) .. controls (245.57,152.77) and (236,143.2) .. (236,131.4) -- cycle ;
	\draw [color={rgb, 255:red, 0; green, 0; blue, 0 }  ,draw opacity=1 ]   (204.73,132.4) -- (233,132.4) ;
	\draw [shift={(236,132.4)}, rotate = 180] [fill={rgb, 255:red, 0; green, 0; blue, 0 }  ,fill opacity=1 ][line width=0.08]  [draw opacity=0] (8.93,-4.29) -- (0,0) -- (8.93,4.29) -- cycle    ;
	\draw  [color={rgb, 255:red, 0; green, 0; blue, 0 }  ,draw opacity=1 ] (160.5,207.57) .. controls (160.5,195.77) and (170.07,186.2) .. (181.87,186.2) .. controls (193.67,186.2) and (203.23,195.77) .. (203.23,207.57) .. controls (203.23,219.37) and (193.67,228.93) .. (181.87,228.93) .. controls (170.07,228.93) and (160.5,219.37) .. (160.5,207.57) -- cycle ;
	\draw  [color={rgb, 255:red, 0; green, 0; blue, 0 }  ,draw opacity=1 ] (235,207.4) .. controls (235,195.6) and (244.57,186.03) .. (256.37,186.03) .. controls (268.17,186.03) and (277.73,195.6) .. (277.73,207.4) .. controls (277.73,219.2) and (268.17,228.77) .. (256.37,228.77) .. controls (244.57,228.77) and (235,219.2) .. (235,207.4) -- cycle ;
	\draw [color={rgb, 255:red, 0; green, 0; blue, 0 }  ,draw opacity=1 ]   (203.73,208.4) -- (232,208.4) ;
	\draw [shift={(235,208.4)}, rotate = 180] [fill={rgb, 255:red, 0; green, 0; blue, 0 }  ,fill opacity=1 ][line width=0.08]  [draw opacity=0] (8.93,-4.29) -- (0,0) -- (8.93,4.29) -- cycle    ;
	
	\draw (38.37,53.33) node  [font=\large]  {${\displaystyle a_{\tau _{0}}}$};
	\draw (112.37,52.33) node  [font=\large,color={rgb, 255:red, 0; green, 0; blue, 0 }  ,opacity=1 ]  {${\displaystyle \hat{a}_{\tau _{1}}^{0}}$};
	\draw (186.37,51.33) node  [font=\large,color={rgb, 255:red, 0; green, 0; blue, 0 }  ,opacity=1 ]  {${\displaystyle \hat{a}_{\tau _{2}}^{0}}$};
	\draw (260.87,51.17) node  [font=\large,color={rgb, 255:red, 0; green, 0; blue, 0 }  ,opacity=1 ]  {${\displaystyle \hat{a}_{\tau _{3}}^{0}}$};
	\draw (111.87,135.57) node  [font=\large,color={rgb, 255:red, 0; green, 0; blue, 0 }  ,opacity=1 ]  {${\displaystyle a_{\tau _{1}}}$};
	\draw (185.87,134.57) node  [font=\large,color={rgb, 255:red, 0; green, 0; blue, 0 }  ,opacity=1 ]  {${\displaystyle \hat{a}_{\tau _{2}}^{1}}$};
	\draw (260.37,134.4) node  [font=\large,color={rgb, 255:red, 0; green, 0; blue, 0 }  ,opacity=1 ]  {${\displaystyle \hat{a}_{\tau _{3}}^{1}}$};
	\draw (184.87,210.57) node  [font=\large,color={rgb, 255:red, 0; green, 0; blue, 0 }  ,opacity=1 ]  {${\displaystyle a_{\tau _{2}}}$};
	\draw (88.9,246.23) node [anchor=north west][inner sep=0.75pt]   [align=left] {Seg. 1};
	\draw (158.5,246.53) node [anchor=north west][inner sep=0.75pt]   [align=left] {Seg. 2};
	\draw (19.3,246.13) node [anchor=north west][inner sep=0.75pt]   [align=left] {Seg. 0};
	\draw (230,246.53) node [anchor=north west][inner sep=0.75pt]   [align=left] {Seg. 3};
	\draw (259.37,210.4) node  [font=\large,color={rgb, 255:red, 0; green, 0; blue, 0 }  ,opacity=1 ]  {${\displaystyle \hat{a}_{\tau _{3}}^{2}}$};
	\draw (8.3,3.3) node [anchor=north west][inner sep=0.75pt]   [align=left] {Branch 0};
	\draw (82,85) node [anchor=north west][inner sep=0.75pt]   [align=left] {Branch 1};
	\draw (154,162) node [anchor=north west][inner sep=0.75pt]   [align=left] {Branch 2};

\end{tikzpicture}

%% file: images/multimodality_strategies/all_modes.tex
\tikzset{every picture/.style={line width=0.75pt}} 

\begin{tikzpicture}[x=0.75pt,y=0.75pt,yscale=-1,xscale=1]
	
	\draw [color={rgb, 255:red, 74; green, 144; blue, 226 }  ,draw opacity=1 ][line width=1.5]    (60,119.58) .. controls (91.78,114.22) and (109.82,121.17) .. (120.4,121.53) ;
	\draw [shift={(124.33,121.26)}, rotate = 182.71] [fill={rgb, 255:red, 74; green, 144; blue, 226 }  ,fill opacity=1 ][line width=0.08]  [draw opacity=0] (9.29,-4.46) -- (0,0) -- (9.29,4.46) -- cycle    ;
	\draw [color={rgb, 255:red, 74; green, 144; blue, 226 }  ,draw opacity=1 ][line width=1.5]    (60,119.58) .. controls (79.95,124.42) and (93.33,150.76) .. (118.67,151.36) ;
	\draw [shift={(122.33,151.27)}, rotate = 199.35] [fill={rgb, 255:red, 74; green, 144; blue, 226 }  ,fill opacity=1 ][line width=0.08]  [draw opacity=0] (9.29,-4.46) -- (0,0) -- (9.29,4.46) -- cycle    ;
	\draw [color={rgb, 255:red, 74; green, 144; blue, 226 }  ,draw opacity=1 ][line width=1.5]    (60,119.58) .. controls (94.2,105.05) and (100.1,97.45) .. (124.74,95.03) ;
	\draw [shift={(128.35,94.73)}, rotate = 171.4] [fill={rgb, 255:red, 74; green, 144; blue, 226 }  ,fill opacity=1 ][line width=0.08]  [draw opacity=0] (9.29,-4.46) -- (0,0) -- (9.29,4.46) -- cycle    ;
	\draw [color={rgb, 255:red, 74; green, 144; blue, 226 }  ,draw opacity=1 ][line width=1.5]    (0.33,111.63) .. controls (28.17,96.06) and (47.35,117.95) .. (57.53,119.7) ;
	\draw [shift={(61.33,119.04)}, rotate = 191.95] [fill={rgb, 255:red, 74; green, 144; blue, 226 }  ,fill opacity=1 ][line width=0.08]  [draw opacity=0] (9.29,-4.46) -- (0,0) -- (9.29,4.46) -- cycle    ;
	\draw [color={rgb, 255:red, 74; green, 144; blue, 226 }  ,draw opacity=1 ][line width=1.5]    (0.33,111.63) .. controls (6.78,116.82) and (13.19,130.01) .. (19.85,144.7) .. controls (31.23,169.79) and (43.33,199.26) .. (57.57,200.82) ;
	\draw [shift={(61.33,200.57)}, rotate = 200.95] [fill={rgb, 255:red, 74; green, 144; blue, 226 }  ,fill opacity=1 ][line width=0.08]  [draw opacity=0] (9.29,-4.46) -- (0,0) -- (9.29,4.46) -- cycle    ;
	\draw [color={rgb, 255:red, 74; green, 144; blue, 226 }  ,draw opacity=1 ][line width=1.5]    (0.33,111.63) .. controls (30.31,69.38) and (35.6,47.2) .. (57.02,40.01) ;
	\draw [shift={(60.54,38.99)}, rotate = 157.73] [fill={rgb, 255:red, 74; green, 144; blue, 226 }  ,fill opacity=1 ][line width=0.08]  [draw opacity=0] (9.29,-4.46) -- (0,0) -- (9.29,4.46) -- cycle    ;
	\draw [color={rgb, 255:red, 74; green, 144; blue, 226 }  ,draw opacity=1 ][line width=1.5]    (64.24,200.15) .. controls (81.97,205.55) and (99.67,239.85) .. (122.66,241.32) ;
	\draw [shift={(126.33,241.27)}, rotate = 205.64] [fill={rgb, 255:red, 74; green, 144; blue, 226 }  ,fill opacity=1 ][line width=0.08]  [draw opacity=0] (9.29,-4.46) -- (0,0) -- (9.29,4.46) -- cycle    ;
	\draw [color={rgb, 255:red, 74; green, 144; blue, 226 }  ,draw opacity=1 ][line width=1.5]    (64.24,200.15) .. controls (94.48,184.05) and (93.72,184.58) .. (114.48,182.76) ;
	\draw [shift={(118.33,182.42)}, rotate = 174.9] [fill={rgb, 255:red, 74; green, 144; blue, 226 }  ,fill opacity=1 ][line width=0.08]  [draw opacity=0] (9.29,-4.46) -- (0,0) -- (9.29,4.46) -- cycle    ;
	\draw [color={rgb, 255:red, 74; green, 144; blue, 226 }  ,draw opacity=1 ][line width=1.5]    (60.09,36.5) .. controls (90.49,31.14) and (111.39,38.75) .. (122.39,39.27) ;
	\draw [shift={(126.33,39.03)}, rotate = 183.81] [fill={rgb, 255:red, 74; green, 144; blue, 226 }  ,fill opacity=1 ][line width=0.08]  [draw opacity=0] (9.29,-4.46) -- (0,0) -- (9.29,4.46) -- cycle    ;
	\draw [color={rgb, 255:red, 74; green, 144; blue, 226 }  ,draw opacity=1 ][line width=1.5]    (60.09,36.5) .. controls (79.08,41.31) and (97.81,66.27) .. (122.4,67.02) ;
	\draw [shift={(126.33,66.93)}, rotate = 197.65] [fill={rgb, 255:red, 74; green, 144; blue, 226 }  ,fill opacity=1 ][line width=0.08]  [draw opacity=0] (9.29,-4.46) -- (0,0) -- (9.29,4.46) -- cycle    ;
	\draw [color={rgb, 255:red, 74; green, 144; blue, 226 }  ,draw opacity=1 ][line width=1.5]    (60.09,36.5) .. controls (92.64,22.04) and (98.39,14.45) .. (121.65,11.99) ;
	\draw [shift={(125.48,11.65)}, rotate = 170.81] [fill={rgb, 255:red, 74; green, 144; blue, 226 }  ,fill opacity=1 ][line width=0.08]  [draw opacity=0] (9.29,-4.46) -- (0,0) -- (9.29,4.46) -- cycle    ;
	\draw [color={rgb, 255:red, 74; green, 144; blue, 226 }  ,draw opacity=1 ][line width=1.5]    (123.29,240.8) .. controls (140.36,241.37) and (154.48,243.05) .. (165.67,244.67) .. controls (167,244.87) and (168.29,245.06) .. (169.53,245.25) .. controls (177.44,246.46) and (183.67,247.56) .. (188.26,248.07) ;
	\draw [shift={(192.2,248.36)}, rotate = 188.39] [fill={rgb, 255:red, 74; green, 144; blue, 226 }  ,fill opacity=1 ][line width=0.08]  [draw opacity=0] (9.29,-4.46) -- (0,0) -- (9.29,4.46) -- cycle    ;
	\draw [color={rgb, 255:red, 74; green, 144; blue, 226 }  ,draw opacity=1 ][line width=1.5]    (124.31,240.56) .. controls (144.06,244.26) and (162.99,255.42) .. (188.79,258.08) ;
	\draw [shift={(192.48,258.4)}, rotate = 192.05] [fill={rgb, 255:red, 74; green, 144; blue, 226 }  ,fill opacity=1 ][line width=0.08]  [draw opacity=0] (9.29,-4.46) -- (0,0) -- (9.29,4.46) -- cycle    ;
	\draw [color={rgb, 255:red, 74; green, 144; blue, 226 }  ,draw opacity=1 ][line width=1.5]    (124.31,240.56) .. controls (158.77,238.48) and (164.84,236.25) .. (189.45,237.72) ;
	\draw [shift={(193.06,237.95)}, rotate = 182.14] [fill={rgb, 255:red, 74; green, 144; blue, 226 }  ,fill opacity=1 ][line width=0.08]  [draw opacity=0] (9.29,-4.46) -- (0,0) -- (9.29,4.46) -- cycle    ;
	\draw [color={rgb, 255:red, 74; green, 144; blue, 226 }  ,draw opacity=1 ][line width=1.5]    (120.11,211.78) .. controls (137.17,210.75) and (151.36,211.09) .. (162.63,211.67) .. controls (163.97,211.74) and (165.27,211.81) .. (166.52,211.88) .. controls (174.48,212.34) and (180.78,212.86) .. (185.38,212.93) ;
	\draw [shift={(189.33,212.85)}, rotate = 183.02] [fill={rgb, 255:red, 74; green, 144; blue, 226 }  ,fill opacity=1 ][line width=0.08]  [draw opacity=0] (9.29,-4.46) -- (0,0) -- (9.29,4.46) -- cycle    ;
	\draw [color={rgb, 255:red, 74; green, 144; blue, 226 }  ,draw opacity=1 ][line width=1.5]    (121.12,211.45) .. controls (141.06,213.28) and (160.72,222.63) .. (186.61,222.87) ;
	\draw [shift={(190.32,222.84)}, rotate = 186.71] [fill={rgb, 255:red, 74; green, 144; blue, 226 }  ,fill opacity=1 ][line width=0.08]  [draw opacity=0] (9.29,-4.46) -- (0,0) -- (9.29,4.46) -- cycle    ;
	\draw [color={rgb, 255:red, 74; green, 144; blue, 226 }  ,draw opacity=1 ][line width=1.5]    (121.12,211.45) .. controls (155.14,206.17) and (161.15,203.38) .. (185.46,202.52) ;
	\draw [shift={(189.45,202.4)}, rotate = 176.76] [fill={rgb, 255:red, 74; green, 144; blue, 226 }  ,fill opacity=1 ][line width=0.08]  [draw opacity=0] (9.29,-4.46) -- (0,0) -- (9.29,4.46) -- cycle    ;
	\draw [color={rgb, 255:red, 74; green, 144; blue, 226 }  ,draw opacity=1 ][line width=1.5]    (119.11,183) .. controls (136.17,181.94) and (150.36,182.25) .. (161.63,182.81) .. controls (162.97,182.87) and (164.27,182.94) .. (165.52,183.01) .. controls (173.48,183.46) and (179.78,183.96) .. (184.38,184.03) ;
	\draw [shift={(188.33,183.94)}, rotate = 182.91] [fill={rgb, 255:red, 74; green, 144; blue, 226 }  ,fill opacity=1 ][line width=0.08]  [draw opacity=0] (9.29,-4.46) -- (0,0) -- (9.29,4.46) -- cycle    ;
	\draw [color={rgb, 255:red, 74; green, 144; blue, 226 }  ,draw opacity=1 ][line width=1.5]    (120.12,182.67) .. controls (140.06,184.46) and (159.73,193.78) .. (185.63,193.96) ;
	\draw [shift={(189.33,193.92)}, rotate = 186.6] [fill={rgb, 255:red, 74; green, 144; blue, 226 }  ,fill opacity=1 ][line width=0.08]  [draw opacity=0] (9.29,-4.46) -- (0,0) -- (9.29,4.46) -- cycle    ;
	\draw [color={rgb, 255:red, 74; green, 144; blue, 226 }  ,draw opacity=1 ][line width=1.5]    (120.12,182.67) .. controls (154.13,177.33) and (160.14,174.52) .. (184.44,173.61) ;
	\draw [shift={(188.44,173.48)}, rotate = 176.65] [fill={rgb, 255:red, 74; green, 144; blue, 226 }  ,fill opacity=1 ][line width=0.08]  [draw opacity=0] (9.29,-4.46) -- (0,0) -- (9.29,4.46) -- cycle    ;
	\draw [color={rgb, 255:red, 74; green, 144; blue, 226 }  ,draw opacity=1 ][line width=1.5]    (119.11,150.69) .. controls (136.17,149.63) and (150.36,149.95) .. (161.63,150.5) .. controls (162.97,150.56) and (164.27,150.63) .. (165.52,150.7) .. controls (173.48,151.15) and (179.78,151.65) .. (184.38,151.72) ;
	\draw [shift={(188.33,151.63)}, rotate = 182.91] [fill={rgb, 255:red, 74; green, 144; blue, 226 }  ,fill opacity=1 ][line width=0.08]  [draw opacity=0] (9.29,-4.46) -- (0,0) -- (9.29,4.46) -- cycle    ;
	\draw [color={rgb, 255:red, 74; green, 144; blue, 226 }  ,draw opacity=1 ][line width=1.5]    (120.12,150.36) .. controls (140.06,152.15) and (159.73,161.47) .. (185.63,161.65) ;
	\draw [shift={(189.33,161.61)}, rotate = 186.6] [fill={rgb, 255:red, 74; green, 144; blue, 226 }  ,fill opacity=1 ][line width=0.08]  [draw opacity=0] (9.29,-4.46) -- (0,0) -- (9.29,4.46) -- cycle    ;
	\draw [color={rgb, 255:red, 74; green, 144; blue, 226 }  ,draw opacity=1 ][line width=1.5]    (120.12,150.36) .. controls (154.13,145.02) and (160.14,142.21) .. (184.44,141.3) ;
	\draw [shift={(188.44,141.17)}, rotate = 176.65] [fill={rgb, 255:red, 74; green, 144; blue, 226 }  ,fill opacity=1 ][line width=0.08]  [draw opacity=0] (9.29,-4.46) -- (0,0) -- (9.29,4.46) -- cycle    ;
	\draw [color={rgb, 255:red, 74; green, 144; blue, 226 }  ,draw opacity=1 ][line width=1.5]    (64.24,200.15) .. controls (76.5,203.88) and (87.46,208.49) .. (100.56,210.57) .. controls (106.34,211.48) and (112.54,211.91) .. (119.45,211.55) ;
	\draw [shift={(123.33,211.27)}, rotate = 183.52] [fill={rgb, 255:red, 74; green, 144; blue, 226 }  ,fill opacity=1 ][line width=0.08]  [draw opacity=0] (9.29,-4.46) -- (0,0) -- (9.29,4.46) -- cycle    ;
	\draw [color={rgb, 255:red, 74; green, 144; blue, 226 }  ,draw opacity=1 ][line width=1.5]    (122.11,123) .. controls (138.44,121.93) and (152.03,122.25) .. (162.81,122.8) .. controls (164.09,122.87) and (165.34,122.94) .. (166.54,123.01) .. controls (174.06,123.45) and (180.03,123.94) .. (184.42,124.02) ;
	\draw [shift={(188.37,123.94)}, rotate = 183.17] [fill={rgb, 255:red, 74; green, 144; blue, 226 }  ,fill opacity=1 ][line width=0.08]  [draw opacity=0] (9.29,-4.46) -- (0,0) -- (9.29,4.46) -- cycle    ;
	\draw [color={rgb, 255:red, 74; green, 144; blue, 226 }  ,draw opacity=1 ][line width=1.5]    (123.07,122.67) .. controls (142.07,124.45) and (160.8,133.68) .. (185.4,133.95) ;
	\draw [shift={(189.33,133.92)}, rotate = 187.24] [fill={rgb, 255:red, 74; green, 144; blue, 226 }  ,fill opacity=1 ][line width=0.08]  [draw opacity=0] (9.29,-4.46) -- (0,0) -- (9.29,4.46) -- cycle    ;
	\draw [color={rgb, 255:red, 74; green, 144; blue, 226 }  ,draw opacity=1 ][line width=1.5]    (123.07,122.67) .. controls (155.63,117.32) and (161.38,114.52) .. (184.65,113.61) ;
	\draw [shift={(188.48,113.48)}, rotate = 176.33] [fill={rgb, 255:red, 74; green, 144; blue, 226 }  ,fill opacity=1 ][line width=0.08]  [draw opacity=0] (9.29,-4.46) -- (0,0) -- (9.29,4.46) -- cycle    ;
	\draw [color={rgb, 255:red, 74; green, 144; blue, 226 }  ,draw opacity=1 ][line width=1.5]    (123.11,95.3) .. controls (139.44,94.24) and (153.03,94.56) .. (163.81,95.11) .. controls (165.09,95.18) and (166.34,95.25) .. (167.54,95.32) .. controls (175.06,95.76) and (181.03,96.25) .. (185.42,96.33) ;
	\draw [shift={(189.37,96.24)}, rotate = 183.17] [fill={rgb, 255:red, 74; green, 144; blue, 226 }  ,fill opacity=1 ][line width=0.08]  [draw opacity=0] (9.29,-4.46) -- (0,0) -- (9.29,4.46) -- cycle    ;
	\draw [color={rgb, 255:red, 74; green, 144; blue, 226 }  ,draw opacity=1 ][line width=1.5]    (124.07,94.97) .. controls (143.07,96.76) and (161.8,105.98) .. (186.4,106.26) ;
	\draw [shift={(190.33,106.22)}, rotate = 187.24] [fill={rgb, 255:red, 74; green, 144; blue, 226 }  ,fill opacity=1 ][line width=0.08]  [draw opacity=0] (9.29,-4.46) -- (0,0) -- (9.29,4.46) -- cycle    ;
	\draw [color={rgb, 255:red, 74; green, 144; blue, 226 }  ,draw opacity=1 ][line width=1.5]    (124.07,94.97) .. controls (156.63,89.63) and (162.38,86.83) .. (185.65,85.92) ;
	\draw [shift={(189.48,85.79)}, rotate = 176.33] [fill={rgb, 255:red, 74; green, 144; blue, 226 }  ,fill opacity=1 ][line width=0.08]  [draw opacity=0] (9.29,-4.46) -- (0,0) -- (9.29,4.46) -- cycle    ;
	\draw [color={rgb, 255:red, 74; green, 144; blue, 226 }  ,draw opacity=1 ][line width=1.5]    (121.11,11.07) .. controls (138.17,10.01) and (152.36,10.33) .. (163.63,10.88) .. controls (164.97,10.94) and (166.27,11.01) .. (167.52,11.08) .. controls (175.48,11.53) and (181.78,12.03) .. (186.38,12.1) ;
	\draw [shift={(190.33,12.01)}, rotate = 182.91] [fill={rgb, 255:red, 74; green, 144; blue, 226 }  ,fill opacity=1 ][line width=0.08]  [draw opacity=0] (9.29,-4.46) -- (0,0) -- (9.29,4.46) -- cycle    ;
	\draw [color={rgb, 255:red, 74; green, 144; blue, 226 }  ,draw opacity=1 ][line width=1.5]    (122.12,10.74) .. controls (142.06,12.53) and (161.73,21.85) .. (187.63,22.03) ;
	\draw [shift={(191.33,21.99)}, rotate = 186.6] [fill={rgb, 255:red, 74; green, 144; blue, 226 }  ,fill opacity=1 ][line width=0.08]  [draw opacity=0] (9.29,-4.46) -- (0,0) -- (9.29,4.46) -- cycle    ;
	\draw [color={rgb, 255:red, 74; green, 144; blue, 226 }  ,draw opacity=1 ][line width=1.5]    (122.12,10.74) .. controls (156.13,5.4) and (162.14,2.59) .. (186.44,1.68) ;
	\draw [shift={(190.44,1.55)}, rotate = 176.65] [fill={rgb, 255:red, 74; green, 144; blue, 226 }  ,fill opacity=1 ][line width=0.08]  [draw opacity=0] (9.29,-4.46) -- (0,0) -- (9.29,4.46) -- cycle    ;
	\draw [color={rgb, 255:red, 74; green, 144; blue, 226 }  ,draw opacity=1 ][line width=1.5]    (124.11,39.92) .. controls (141.17,38.86) and (155.36,39.17) .. (166.63,39.73) .. controls (167.97,39.79) and (169.27,39.86) .. (170.52,39.93) .. controls (178.48,40.38) and (184.78,40.88) .. (189.38,40.94) ;
	\draw [shift={(193.33,40.86)}, rotate = 182.91] [fill={rgb, 255:red, 74; green, 144; blue, 226 }  ,fill opacity=1 ][line width=0.08]  [draw opacity=0] (9.29,-4.46) -- (0,0) -- (9.29,4.46) -- cycle    ;
	\draw [color={rgb, 255:red, 74; green, 144; blue, 226 }  ,draw opacity=1 ][line width=1.5]    (125.12,39.59) .. controls (145.06,41.38) and (164.73,50.69) .. (190.63,50.88) ;
	\draw [shift={(194.33,50.84)}, rotate = 186.6] [fill={rgb, 255:red, 74; green, 144; blue, 226 }  ,fill opacity=1 ][line width=0.08]  [draw opacity=0] (9.29,-4.46) -- (0,0) -- (9.29,4.46) -- cycle    ;
	\draw [color={rgb, 255:red, 74; green, 144; blue, 226 }  ,draw opacity=1 ][line width=1.5]    (125.12,39.59) .. controls (159.13,34.24) and (165.14,31.44) .. (189.44,30.53) ;
	\draw [shift={(193.44,30.4)}, rotate = 176.65] [fill={rgb, 255:red, 74; green, 144; blue, 226 }  ,fill opacity=1 ][line width=0.08]  [draw opacity=0] (9.29,-4.46) -- (0,0) -- (9.29,4.46) -- cycle    ;
	\draw [color={rgb, 255:red, 74; green, 144; blue, 226 }  ,draw opacity=1 ][line width=1.5]    (123.11,67.61) .. controls (140.17,66.55) and (154.36,66.87) .. (165.63,67.42) .. controls (166.97,67.48) and (168.27,67.55) .. (169.52,67.62) .. controls (177.48,68.07) and (183.78,68.57) .. (188.38,68.64) ;
	\draw [shift={(192.33,68.55)}, rotate = 182.91] [fill={rgb, 255:red, 74; green, 144; blue, 226 }  ,fill opacity=1 ][line width=0.08]  [draw opacity=0] (9.29,-4.46) -- (0,0) -- (9.29,4.46) -- cycle    ;
	\draw [color={rgb, 255:red, 74; green, 144; blue, 226 }  ,draw opacity=1 ][line width=1.5]    (124.12,67.28) .. controls (144.06,69.07) and (163.73,78.39) .. (189.63,78.57) ;
	\draw [shift={(193.33,78.53)}, rotate = 186.6] [fill={rgb, 255:red, 74; green, 144; blue, 226 }  ,fill opacity=1 ][line width=0.08]  [draw opacity=0] (9.29,-4.46) -- (0,0) -- (9.29,4.46) -- cycle    ;
	\draw [color={rgb, 255:red, 74; green, 144; blue, 226 }  ,draw opacity=1 ][line width=1.5]    (124.12,67.28) .. controls (158.13,61.94) and (164.14,59.13) .. (188.44,58.22) ;
	\draw [shift={(192.44,58.09)}, rotate = 176.65] [fill={rgb, 255:red, 74; green, 144; blue, 226 }  ,fill opacity=1 ][line width=0.08]  [draw opacity=0] (9.29,-4.46) -- (0,0) -- (9.29,4.46) -- cycle    ;

\end{tikzpicture}

%% file: images/multimodality_strategies/best_mode.tex
\tikzset{every picture/.style={line width=0.75pt}} 

\begin{tikzpicture}[x=0.75pt,y=0.75pt,yscale=-1,xscale=1]
	
	\draw [color={rgb, 255:red, 74; green, 144; blue, 226 }  ,draw opacity=1 ][line width=1.5]    (59,123.58) .. controls (90.78,118.22) and (108.82,125.17) .. (119.4,125.53) ;
	\draw [shift={(123.33,125.26)}, rotate = 182.71] [fill={rgb, 255:red, 74; green, 144; blue, 226 }  ,fill opacity=1 ][line width=0.08]  [draw opacity=0] (9.29,-4.46) -- (0,0) -- (9.29,4.46) -- cycle    ;
	\draw [color={rgb, 255:red, 155; green, 155; blue, 155 }  ,draw opacity=1 ][line width=1.5]    (59,123.58) .. controls (78.95,128.42) and (92.33,154.76) .. (117.67,155.36) ;
	\draw [shift={(121.33,155.27)}, rotate = 199.35] [fill={rgb, 255:red, 155; green, 155; blue, 155 }  ,fill opacity=1 ][line width=0.08]  [draw opacity=0] (9.29,-4.46) -- (0,0) -- (9.29,4.46) -- cycle    ;
	\draw [color={rgb, 255:red, 155; green, 155; blue, 155 }  ,draw opacity=1 ][line width=1.5]    (59,123.58) .. controls (93.2,109.05) and (99.1,101.45) .. (123.74,99.03) ;
	\draw [shift={(127.35,98.73)}, rotate = 171.4] [fill={rgb, 255:red, 155; green, 155; blue, 155 }  ,fill opacity=1 ][line width=0.08]  [draw opacity=0] (9.29,-4.46) -- (0,0) -- (9.29,4.46) -- cycle    ;
	\draw [color={rgb, 255:red, 74; green, 144; blue, 226 }  ,draw opacity=1 ][line width=1.5]    (-0.67,115.63) .. controls (27.17,100.06) and (46.35,121.95) .. (56.53,123.7) ;
	\draw [shift={(60.33,123.04)}, rotate = 191.95] [fill={rgb, 255:red, 74; green, 144; blue, 226 }  ,fill opacity=1 ][line width=0.08]  [draw opacity=0] (9.29,-4.46) -- (0,0) -- (9.29,4.46) -- cycle    ;
	\draw [color={rgb, 255:red, 155; green, 155; blue, 155 }  ,draw opacity=1 ][line width=1.5]    (-0.67,115.63) .. controls (5.78,120.82) and (12.19,134.01) .. (18.85,148.7) .. controls (30.23,173.79) and (42.33,203.26) .. (56.57,204.82) ;
	\draw [shift={(60.33,204.57)}, rotate = 200.95] [fill={rgb, 255:red, 155; green, 155; blue, 155 }  ,fill opacity=1 ][line width=0.08]  [draw opacity=0] (9.29,-4.46) -- (0,0) -- (9.29,4.46) -- cycle    ;
	\draw [color={rgb, 255:red, 155; green, 155; blue, 155 }  ,draw opacity=1 ][line width=1.5]    (-0.67,115.63) .. controls (29.31,73.38) and (34.6,51.2) .. (56.02,44.01) ;
	\draw [shift={(59.54,42.99)}, rotate = 157.73] [fill={rgb, 255:red, 155; green, 155; blue, 155 }  ,fill opacity=1 ][line width=0.08]  [draw opacity=0] (9.29,-4.46) -- (0,0) -- (9.29,4.46) -- cycle    ;
	\draw [color={rgb, 255:red, 74; green, 144; blue, 226 }  ,draw opacity=1 ][line width=1.5]    (121.11,127) .. controls (137.44,125.93) and (151.03,126.25) .. (161.81,126.8) .. controls (163.09,126.87) and (164.34,126.94) .. (165.54,127.01) .. controls (173.06,127.45) and (179.03,127.94) .. (183.42,128.02) ;
	\draw [shift={(187.37,127.94)}, rotate = 183.17] [fill={rgb, 255:red, 74; green, 144; blue, 226 }  ,fill opacity=1 ][line width=0.08]  [draw opacity=0] (9.29,-4.46) -- (0,0) -- (9.29,4.46) -- cycle    ;
	\draw [color={rgb, 255:red, 155; green, 155; blue, 155 }  ,draw opacity=1 ][line width=1.5]    (122.07,126.67) .. controls (141.07,128.45) and (159.8,137.68) .. (184.4,137.95) ;
	\draw [shift={(188.33,137.92)}, rotate = 187.24] [fill={rgb, 255:red, 155; green, 155; blue, 155 }  ,fill opacity=1 ][line width=0.08]  [draw opacity=0] (9.29,-4.46) -- (0,0) -- (9.29,4.46) -- cycle    ;
	\draw [color={rgb, 255:red, 155; green, 155; blue, 155 }  ,draw opacity=1 ][line width=1.5]    (122.07,126.67) .. controls (154.63,121.32) and (160.38,118.52) .. (183.65,117.61) ;
	\draw [shift={(187.48,117.48)}, rotate = 176.33] [fill={rgb, 255:red, 155; green, 155; blue, 155 }  ,fill opacity=1 ][line width=0.08]  [draw opacity=0] (9.29,-4.46) -- (0,0) -- (9.29,4.46) -- cycle    ;

\end{tikzpicture}

%% file: images/multimodality_strategies/start_multimodal.tex
\tikzset{every picture/.style={line width=0.75pt}} 

\begin{tikzpicture}[x=0.75pt,y=0.75pt,yscale=-1,xscale=1]
	
	\draw [color={rgb, 255:red, 74; green, 144; blue, 226 }  ,draw opacity=1 ][line width=1.5]    (60,121.58) .. controls (91.78,116.22) and (109.82,123.17) .. (120.4,123.53) ;
	\draw [shift={(124.33,123.26)}, rotate = 182.71] [fill={rgb, 255:red, 74; green, 144; blue, 226 }  ,fill opacity=1 ][line width=0.08]  [draw opacity=0] (9.29,-4.46) -- (0,0) -- (9.29,4.46) -- cycle    ;
	\draw [color={rgb, 255:red, 155; green, 155; blue, 155 }  ,draw opacity=1 ][line width=1.5]    (60,121.58) .. controls (79.95,126.42) and (93.33,152.76) .. (118.67,153.36) ;
	\draw [shift={(122.33,153.27)}, rotate = 199.35] [fill={rgb, 255:red, 155; green, 155; blue, 155 }  ,fill opacity=1 ][line width=0.08]  [draw opacity=0] (9.29,-4.46) -- (0,0) -- (9.29,4.46) -- cycle    ;
	\draw [color={rgb, 255:red, 155; green, 155; blue, 155 }  ,draw opacity=1 ][line width=1.5]    (60,121.58) .. controls (94.2,107.05) and (100.1,99.45) .. (124.74,97.03) ;
	\draw [shift={(128.35,96.73)}, rotate = 171.4] [fill={rgb, 255:red, 155; green, 155; blue, 155 }  ,fill opacity=1 ][line width=0.08]  [draw opacity=0] (9.29,-4.46) -- (0,0) -- (9.29,4.46) -- cycle    ;
	\draw [color={rgb, 255:red, 74; green, 144; blue, 226 }  ,draw opacity=1 ][line width=1.5]    (0.33,113.63) .. controls (28.17,98.06) and (47.35,119.95) .. (57.53,121.7) ;
	\draw [shift={(61.33,121.04)}, rotate = 191.95] [fill={rgb, 255:red, 74; green, 144; blue, 226 }  ,fill opacity=1 ][line width=0.08]  [draw opacity=0] (9.29,-4.46) -- (0,0) -- (9.29,4.46) -- cycle    ;
	\draw [color={rgb, 255:red, 74; green, 144; blue, 226 }  ,draw opacity=1 ][line width=1.5]    (0.33,113.63) .. controls (6.78,118.82) and (13.19,132.01) .. (19.85,146.7) .. controls (31.23,171.79) and (43.33,201.26) .. (57.57,202.82) ;
	\draw [shift={(61.33,202.57)}, rotate = 200.95] [fill={rgb, 255:red, 74; green, 144; blue, 226 }  ,fill opacity=1 ][line width=0.08]  [draw opacity=0] (9.29,-4.46) -- (0,0) -- (9.29,4.46) -- cycle    ;
	\draw [color={rgb, 255:red, 74; green, 144; blue, 226 }  ,draw opacity=1 ][line width=1.5]    (0.33,113.63) .. controls (30.31,71.38) and (35.6,49.2) .. (57.02,42.01) ;
	\draw [shift={(60.54,40.99)}, rotate = 157.73] [fill={rgb, 255:red, 74; green, 144; blue, 226 }  ,fill opacity=1 ][line width=0.08]  [draw opacity=0] (9.29,-4.46) -- (0,0) -- (9.29,4.46) -- cycle    ;
	\draw [color={rgb, 255:red, 74; green, 144; blue, 226 }  ,draw opacity=1 ][line width=1.5]    (64.24,202.15) .. controls (81.97,207.55) and (99.67,241.85) .. (122.66,243.32) ;
	\draw [shift={(126.33,243.27)}, rotate = 205.64] [fill={rgb, 255:red, 74; green, 144; blue, 226 }  ,fill opacity=1 ][line width=0.08]  [draw opacity=0] (9.29,-4.46) -- (0,0) -- (9.29,4.46) -- cycle    ;
	\draw [color={rgb, 255:red, 155; green, 155; blue, 155 }  ,draw opacity=1 ][line width=1.5]    (64.24,202.15) .. controls (94.48,186.05) and (93.72,186.58) .. (114.48,184.76) ;
	\draw [shift={(118.33,184.42)}, rotate = 174.9] [fill={rgb, 255:red, 155; green, 155; blue, 155 }  ,fill opacity=1 ][line width=0.08]  [draw opacity=0] (9.29,-4.46) -- (0,0) -- (9.29,4.46) -- cycle    ;
	\draw [color={rgb, 255:red, 155; green, 155; blue, 155 }  ,draw opacity=1 ][line width=1.5]    (60.09,38.5) .. controls (90.49,33.14) and (111.39,40.75) .. (122.39,41.27) ;
	\draw [shift={(126.33,41.03)}, rotate = 183.81] [fill={rgb, 255:red, 155; green, 155; blue, 155 }  ,fill opacity=1 ][line width=0.08]  [draw opacity=0] (9.29,-4.46) -- (0,0) -- (9.29,4.46) -- cycle    ;
	\draw [color={rgb, 255:red, 155; green, 155; blue, 155 }  ,draw opacity=1 ][line width=1.5]    (60.09,38.5) .. controls (79.08,43.31) and (97.81,68.27) .. (122.4,69.02) ;
	\draw [shift={(126.33,68.93)}, rotate = 197.65] [fill={rgb, 255:red, 155; green, 155; blue, 155 }  ,fill opacity=1 ][line width=0.08]  [draw opacity=0] (9.29,-4.46) -- (0,0) -- (9.29,4.46) -- cycle    ;
	\draw [color={rgb, 255:red, 74; green, 144; blue, 226 }  ,draw opacity=1 ][line width=1.5]    (60.09,38.5) .. controls (92.64,24.04) and (98.39,16.45) .. (121.65,13.99) ;
	\draw [shift={(125.48,13.65)}, rotate = 170.81] [fill={rgb, 255:red, 74; green, 144; blue, 226 }  ,fill opacity=1 ][line width=0.08]  [draw opacity=0] (9.29,-4.46) -- (0,0) -- (9.29,4.46) -- cycle    ;
	\draw [color={rgb, 255:red, 74; green, 144; blue, 226 }  ,draw opacity=1 ][line width=1.5]    (123.29,242.8) .. controls (140.36,243.37) and (154.48,245.05) .. (165.67,246.67) .. controls (167,246.87) and (168.29,247.06) .. (169.53,247.25) .. controls (177.44,248.46) and (183.67,249.56) .. (188.26,250.07) ;
	\draw [shift={(192.2,250.36)}, rotate = 188.39] [fill={rgb, 255:red, 74; green, 144; blue, 226 }  ,fill opacity=1 ][line width=0.08]  [draw opacity=0] (9.29,-4.46) -- (0,0) -- (9.29,4.46) -- cycle    ;
	\draw [color={rgb, 255:red, 155; green, 155; blue, 155 }  ,draw opacity=1 ][line width=1.5]    (124.31,242.56) .. controls (144.06,246.26) and (162.99,257.42) .. (188.79,260.08) ;
	\draw [shift={(192.48,260.4)}, rotate = 192.05] [fill={rgb, 255:red, 155; green, 155; blue, 155 }  ,fill opacity=1 ][line width=0.08]  [draw opacity=0] (9.29,-4.46) -- (0,0) -- (9.29,4.46) -- cycle    ;
	\draw [color={rgb, 255:red, 155; green, 155; blue, 155 }  ,draw opacity=1 ][line width=1.5]    (124.31,242.56) .. controls (158.77,240.48) and (164.84,238.25) .. (189.45,239.72) ;
	\draw [shift={(193.06,239.95)}, rotate = 182.14] [fill={rgb, 255:red, 155; green, 155; blue, 155 }  ,fill opacity=1 ][line width=0.08]  [draw opacity=0] (9.29,-4.46) -- (0,0) -- (9.29,4.46) -- cycle    ;
	\draw [color={rgb, 255:red, 155; green, 155; blue, 155 }  ,draw opacity=1 ][line width=1.5]    (64.24,202.15) .. controls (76.5,205.88) and (87.46,210.49) .. (100.56,212.57) .. controls (106.34,213.48) and (112.54,213.91) .. (119.45,213.55) ;
	\draw [shift={(123.33,213.27)}, rotate = 183.52] [fill={rgb, 255:red, 155; green, 155; blue, 155 }  ,fill opacity=1 ][line width=0.08]  [draw opacity=0] (9.29,-4.46) -- (0,0) -- (9.29,4.46) -- cycle    ;
	\draw [color={rgb, 255:red, 155; green, 155; blue, 155 }  ,draw opacity=1 ][line width=1.5]    (122.11,125) .. controls (138.44,123.93) and (152.03,124.25) .. (162.81,124.8) .. controls (164.09,124.87) and (165.34,124.94) .. (166.54,125.01) .. controls (174.06,125.45) and (180.03,125.94) .. (184.42,126.02) ;
	\draw [shift={(188.37,125.94)}, rotate = 183.17] [fill={rgb, 255:red, 155; green, 155; blue, 155 }  ,fill opacity=1 ][line width=0.08]  [draw opacity=0] (9.29,-4.46) -- (0,0) -- (9.29,4.46) -- cycle    ;
	\draw [color={rgb, 255:red, 155; green, 155; blue, 155 }  ,draw opacity=1 ][line width=1.5]    (123.07,124.67) .. controls (142.07,126.45) and (160.8,135.68) .. (185.4,135.95) ;
	\draw [shift={(189.33,135.92)}, rotate = 187.24] [fill={rgb, 255:red, 155; green, 155; blue, 155 }  ,fill opacity=1 ][line width=0.08]  [draw opacity=0] (9.29,-4.46) -- (0,0) -- (9.29,4.46) -- cycle    ;
	\draw [color={rgb, 255:red, 74; green, 144; blue, 226 }  ,draw opacity=1 ][line width=1.5]    (123.07,124.67) .. controls (155.63,119.32) and (161.38,116.52) .. (184.65,115.61) ;
	\draw [shift={(188.48,115.48)}, rotate = 176.33] [fill={rgb, 255:red, 74; green, 144; blue, 226 }  ,fill opacity=1 ][line width=0.08]  [draw opacity=0] (9.29,-4.46) -- (0,0) -- (9.29,4.46) -- cycle    ;
	\draw [color={rgb, 255:red, 155; green, 155; blue, 155 }  ,draw opacity=1 ][line width=1.5]    (121.11,13.07) .. controls (138.17,12.01) and (152.36,12.33) .. (163.63,12.88) .. controls (164.97,12.94) and (166.27,13.01) .. (167.52,13.08) .. controls (175.48,13.53) and (181.78,14.03) .. (186.38,14.1) ;
	\draw [shift={(190.33,14.01)}, rotate = 182.91] [fill={rgb, 255:red, 155; green, 155; blue, 155 }  ,fill opacity=1 ][line width=0.08]  [draw opacity=0] (9.29,-4.46) -- (0,0) -- (9.29,4.46) -- cycle    ;
	\draw [color={rgb, 255:red, 155; green, 155; blue, 155 }  ,draw opacity=1 ][line width=1.5]    (122.12,12.74) .. controls (142.06,14.53) and (161.73,23.85) .. (187.63,24.03) ;
	\draw [shift={(191.33,23.99)}, rotate = 186.6] [fill={rgb, 255:red, 155; green, 155; blue, 155 }  ,fill opacity=1 ][line width=0.08]  [draw opacity=0] (9.29,-4.46) -- (0,0) -- (9.29,4.46) -- cycle    ;
	\draw [color={rgb, 255:red, 74; green, 144; blue, 226 }  ,draw opacity=1 ][line width=1.5]    (122.12,12.74) .. controls (156.13,7.4) and (162.14,4.59) .. (186.44,3.68) ;
	\draw [shift={(190.44,3.55)}, rotate = 176.65] [fill={rgb, 255:red, 74; green, 144; blue, 226 }  ,fill opacity=1 ][line width=0.08]  [draw opacity=0] (9.29,-4.46) -- (0,0) -- (9.29,4.46) -- cycle    ;

\end{tikzpicture}

%% file: images/multimodality_strategies/end_multimodal.tex
\tikzset{every picture/.style={line width=0.75pt}} 

\begin{tikzpicture}[x=0.75pt,y=0.75pt,yscale=-1,xscale=1]
	
	\draw [color={rgb, 255:red, 74; green, 144; blue, 226 }  ,draw opacity=1 ][line width=1.5]    (60,123.58) .. controls (91.78,118.22) and (109.82,125.17) .. (120.4,125.53) ;
	\draw [shift={(124.33,125.26)}, rotate = 182.71] [fill={rgb, 255:red, 74; green, 144; blue, 226 }  ,fill opacity=1 ][line width=0.08]  [draw opacity=0] (9.29,-4.46) -- (0,0) -- (9.29,4.46) -- cycle    ;
	\draw [color={rgb, 255:red, 155; green, 155; blue, 155 }  ,draw opacity=1 ][line width=1.5]    (60,123.58) .. controls (79.95,128.42) and (93.33,154.76) .. (118.67,155.36) ;
	\draw [shift={(122.33,155.27)}, rotate = 199.35] [fill={rgb, 255:red, 155; green, 155; blue, 155 }  ,fill opacity=1 ][line width=0.08]  [draw opacity=0] (9.29,-4.46) -- (0,0) -- (9.29,4.46) -- cycle    ;
	\draw [color={rgb, 255:red, 155; green, 155; blue, 155 }  ,draw opacity=1 ][line width=1.5]    (60,123.58) .. controls (94.2,109.05) and (100.1,101.45) .. (124.74,99.03) ;
	\draw [shift={(128.35,98.73)}, rotate = 171.4] [fill={rgb, 255:red, 155; green, 155; blue, 155 }  ,fill opacity=1 ][line width=0.08]  [draw opacity=0] (9.29,-4.46) -- (0,0) -- (9.29,4.46) -- cycle    ;
	\draw [color={rgb, 255:red, 74; green, 144; blue, 226 }  ,draw opacity=1 ][line width=1.5]    (0.33,115.63) .. controls (28.17,100.06) and (47.35,121.95) .. (57.53,123.7) ;
	\draw [shift={(61.33,123.04)}, rotate = 191.95] [fill={rgb, 255:red, 74; green, 144; blue, 226 }  ,fill opacity=1 ][line width=0.08]  [draw opacity=0] (9.29,-4.46) -- (0,0) -- (9.29,4.46) -- cycle    ;
	\draw [color={rgb, 255:red, 155; green, 155; blue, 155 }  ,draw opacity=1 ][line width=1.5]    (0.33,115.63) .. controls (6.78,120.82) and (13.19,134.01) .. (19.85,148.7) .. controls (31.23,173.79) and (43.33,203.26) .. (57.57,204.82) ;
	\draw [shift={(61.33,204.57)}, rotate = 200.95] [fill={rgb, 255:red, 155; green, 155; blue, 155 }  ,fill opacity=1 ][line width=0.08]  [draw opacity=0] (9.29,-4.46) -- (0,0) -- (9.29,4.46) -- cycle    ;
	\draw [color={rgb, 255:red, 155; green, 155; blue, 155 }  ,draw opacity=1 ][line width=1.5]    (0.33,115.63) .. controls (30.31,73.38) and (35.6,51.2) .. (57.02,44.01) ;
	\draw [shift={(60.54,42.99)}, rotate = 157.73] [fill={rgb, 255:red, 155; green, 155; blue, 155 }  ,fill opacity=1 ][line width=0.08]  [draw opacity=0] (9.29,-4.46) -- (0,0) -- (9.29,4.46) -- cycle    ;
	\draw [color={rgb, 255:red, 74; green, 144; blue, 226 }  ,draw opacity=1 ][line width=1.5]    (122.11,127) .. controls (138.44,125.93) and (152.03,126.25) .. (162.81,126.8) .. controls (164.09,126.87) and (165.34,126.94) .. (166.54,127.01) .. controls (174.06,127.45) and (180.03,127.94) .. (184.42,128.02) ;
	\draw [shift={(188.37,127.94)}, rotate = 183.17] [fill={rgb, 255:red, 74; green, 144; blue, 226 }  ,fill opacity=1 ][line width=0.08]  [draw opacity=0] (9.29,-4.46) -- (0,0) -- (9.29,4.46) -- cycle    ;
	\draw [color={rgb, 255:red, 74; green, 144; blue, 226 }  ,draw opacity=1 ][line width=1.5]    (123.07,126.67) .. controls (142.07,128.45) and (160.8,137.68) .. (185.4,137.95) ;
	\draw [shift={(189.33,137.92)}, rotate = 187.24] [fill={rgb, 255:red, 74; green, 144; blue, 226 }  ,fill opacity=1 ][line width=0.08]  [draw opacity=0] (9.29,-4.46) -- (0,0) -- (9.29,4.46) -- cycle    ;
	\draw [color={rgb, 255:red, 74; green, 144; blue, 226 }  ,draw opacity=1 ][line width=1.5]    (123.07,126.67) .. controls (155.63,121.32) and (161.38,118.52) .. (184.65,117.61) ;
	\draw [shift={(188.48,117.48)}, rotate = 176.33] [fill={rgb, 255:red, 74; green, 144; blue, 226 }  ,fill opacity=1 ][line width=0.08]  [draw opacity=0] (9.29,-4.46) -- (0,0) -- (9.29,4.46) -- cycle    ;

\end{tikzpicture}

%% file: images/multimodality_strategies/maximum_modes.tex
\tikzset{every picture/.style={line width=0.75pt}} 

\begin{tikzpicture}[x=0.75pt,y=0.75pt,yscale=-1,xscale=1]
	
	\draw [color={rgb, 255:red, 74; green, 144; blue, 226 }  ,draw opacity=1 ][line width=1.5]    (60,119.58) .. controls (91.78,114.22) and (109.82,121.17) .. (120.4,121.53) ;
	\draw [shift={(124.33,121.26)}, rotate = 182.71] [fill={rgb, 255:red, 74; green, 144; blue, 226 }  ,fill opacity=1 ][line width=0.08]  [draw opacity=0] (9.29,-4.46) -- (0,0) -- (9.29,4.46) -- cycle    ;
	\draw [color={rgb, 255:red, 74; green, 144; blue, 226 }  ,draw opacity=1 ][line width=1.5]    (60,119.58) .. controls (79.95,124.42) and (93.33,150.76) .. (118.67,151.36) ;
	\draw [shift={(122.33,151.27)}, rotate = 199.35] [fill={rgb, 255:red, 74; green, 144; blue, 226 }  ,fill opacity=1 ][line width=0.08]  [draw opacity=0] (9.29,-4.46) -- (0,0) -- (9.29,4.46) -- cycle    ;
	\draw [color={rgb, 255:red, 155; green, 155; blue, 155 }  ,draw opacity=1 ][line width=1.5]    (60,119.58) .. controls (94.2,105.05) and (100.1,97.45) .. (124.74,95.03) ;
	\draw [shift={(128.35,94.73)}, rotate = 171.4] [fill={rgb, 255:red, 155; green, 155; blue, 155 }  ,fill opacity=1 ][line width=0.08]  [draw opacity=0] (9.29,-4.46) -- (0,0) -- (9.29,4.46) -- cycle    ;
	\draw [color={rgb, 255:red, 74; green, 144; blue, 226 }  ,draw opacity=1 ][line width=1.5]    (0.33,111.63) .. controls (28.17,96.06) and (47.35,117.95) .. (57.53,119.7) ;
	\draw [shift={(61.33,119.04)}, rotate = 191.95] [fill={rgb, 255:red, 74; green, 144; blue, 226 }  ,fill opacity=1 ][line width=0.08]  [draw opacity=0] (9.29,-4.46) -- (0,0) -- (9.29,4.46) -- cycle    ;
	\draw [color={rgb, 255:red, 74; green, 144; blue, 226 }  ,draw opacity=1 ][line width=1.5]    (0.33,111.63) .. controls (6.78,116.82) and (13.19,130.01) .. (19.85,144.7) .. controls (31.23,169.79) and (43.33,199.26) .. (57.57,200.82) ;
	\draw [shift={(61.33,200.57)}, rotate = 200.95] [fill={rgb, 255:red, 74; green, 144; blue, 226 }  ,fill opacity=1 ][line width=0.08]  [draw opacity=0] (9.29,-4.46) -- (0,0) -- (9.29,4.46) -- cycle    ;
	\draw [color={rgb, 255:red, 74; green, 144; blue, 226 }  ,draw opacity=1 ][line width=1.5]    (0.33,111.63) .. controls (30.31,69.38) and (35.6,47.2) .. (57.02,40.01) ;
	\draw [shift={(60.54,38.99)}, rotate = 157.73] [fill={rgb, 255:red, 74; green, 144; blue, 226 }  ,fill opacity=1 ][line width=0.08]  [draw opacity=0] (9.29,-4.46) -- (0,0) -- (9.29,4.46) -- cycle    ;
	\draw [color={rgb, 255:red, 155; green, 155; blue, 155 }  ,draw opacity=1 ][line width=1.5]    (64.24,200.15) .. controls (81.97,205.55) and (99.67,239.85) .. (122.66,241.32) ;
	\draw [shift={(126.33,241.27)}, rotate = 205.64] [fill={rgb, 255:red, 155; green, 155; blue, 155 }  ,fill opacity=1 ][line width=0.08]  [draw opacity=0] (9.29,-4.46) -- (0,0) -- (9.29,4.46) -- cycle    ;
	\draw [color={rgb, 255:red, 155; green, 155; blue, 155 }  ,draw opacity=1 ][line width=1.5]    (64.24,200.15) .. controls (94.48,184.05) and (93.72,184.58) .. (114.48,182.76) ;
	\draw [shift={(118.33,182.42)}, rotate = 174.9] [fill={rgb, 255:red, 155; green, 155; blue, 155 }  ,fill opacity=1 ][line width=0.08]  [draw opacity=0] (9.29,-4.46) -- (0,0) -- (9.29,4.46) -- cycle    ;
	\draw [color={rgb, 255:red, 155; green, 155; blue, 155 }  ,draw opacity=1 ][line width=1.5]    (60.09,36.5) .. controls (90.49,31.14) and (111.39,38.75) .. (122.39,39.27) ;
	\draw [shift={(126.33,39.03)}, rotate = 183.81] [fill={rgb, 255:red, 155; green, 155; blue, 155 }  ,fill opacity=1 ][line width=0.08]  [draw opacity=0] (9.29,-4.46) -- (0,0) -- (9.29,4.46) -- cycle    ;
	\draw [color={rgb, 255:red, 155; green, 155; blue, 155 }  ,draw opacity=1 ][line width=1.5]    (60.09,36.5) .. controls (79.08,41.31) and (97.81,66.27) .. (122.4,67.02) ;
	\draw [shift={(126.33,66.93)}, rotate = 197.65] [fill={rgb, 255:red, 155; green, 155; blue, 155 }  ,fill opacity=1 ][line width=0.08]  [draw opacity=0] (9.29,-4.46) -- (0,0) -- (9.29,4.46) -- cycle    ;
	\draw [color={rgb, 255:red, 74; green, 144; blue, 226 }  ,draw opacity=1 ][line width=1.5]    (60.09,36.5) .. controls (92.64,22.04) and (98.39,14.45) .. (121.65,11.99) ;
	\draw [shift={(125.48,11.65)}, rotate = 170.81] [fill={rgb, 255:red, 74; green, 144; blue, 226 }  ,fill opacity=1 ][line width=0.08]  [draw opacity=0] (9.29,-4.46) -- (0,0) -- (9.29,4.46) -- cycle    ;
	\draw [color={rgb, 255:red, 155; green, 155; blue, 155 }  ,draw opacity=1 ][line width=1.5]    (119.11,150.69) .. controls (136.17,149.63) and (150.36,149.95) .. (161.63,150.5) .. controls (162.97,150.56) and (164.27,150.63) .. (165.52,150.7) .. controls (173.48,151.15) and (179.78,151.65) .. (184.38,151.72) ;
	\draw [shift={(188.33,151.63)}, rotate = 182.91] [fill={rgb, 255:red, 155; green, 155; blue, 155 }  ,fill opacity=1 ][line width=0.08]  [draw opacity=0] (9.29,-4.46) -- (0,0) -- (9.29,4.46) -- cycle    ;
	\draw [color={rgb, 255:red, 155; green, 155; blue, 155 }  ,draw opacity=1 ][line width=1.5]    (120.12,150.36) .. controls (140.06,152.15) and (159.73,161.47) .. (185.63,161.65) ;
	\draw [shift={(189.33,161.61)}, rotate = 186.6] [fill={rgb, 255:red, 155; green, 155; blue, 155 }  ,fill opacity=1 ][line width=0.08]  [draw opacity=0] (9.29,-4.46) -- (0,0) -- (9.29,4.46) -- cycle    ;
	\draw [color={rgb, 255:red, 155; green, 155; blue, 155 }  ,draw opacity=1 ][line width=1.5]    (120.12,150.36) .. controls (154.13,145.02) and (160.14,142.21) .. (184.44,141.3) ;
	\draw [shift={(188.44,141.17)}, rotate = 176.65] [fill={rgb, 255:red, 155; green, 155; blue, 155 }  ,fill opacity=1 ][line width=0.08]  [draw opacity=0] (9.29,-4.46) -- (0,0) -- (9.29,4.46) -- cycle    ;
	\draw [color={rgb, 255:red, 155; green, 155; blue, 155 }  ,draw opacity=1 ][line width=1.5]    (64.24,200.15) .. controls (76.5,203.88) and (87.46,208.49) .. (100.56,210.57) .. controls (106.34,211.48) and (112.54,211.91) .. (119.45,211.55) ;
	\draw [shift={(123.33,211.27)}, rotate = 183.52] [fill={rgb, 255:red, 155; green, 155; blue, 155 }  ,fill opacity=1 ][line width=0.08]  [draw opacity=0] (9.29,-4.46) -- (0,0) -- (9.29,4.46) -- cycle    ;
	\draw [color={rgb, 255:red, 155; green, 155; blue, 155 }  ,draw opacity=1 ][line width=1.5]    (122.11,123) .. controls (138.44,121.93) and (152.03,122.25) .. (162.81,122.8) .. controls (164.09,122.87) and (165.34,122.94) .. (166.54,123.01) .. controls (174.06,123.45) and (180.03,123.94) .. (184.42,124.02) ;
	\draw [shift={(188.37,123.94)}, rotate = 183.17] [fill={rgb, 255:red, 155; green, 155; blue, 155 }  ,fill opacity=1 ][line width=0.08]  [draw opacity=0] (9.29,-4.46) -- (0,0) -- (9.29,4.46) -- cycle    ;
	\draw [color={rgb, 255:red, 74; green, 144; blue, 226 }  ,draw opacity=1 ][line width=1.5]    (123.07,122.67) .. controls (142.07,124.45) and (160.8,133.68) .. (185.4,133.95) ;
	\draw [shift={(189.33,133.92)}, rotate = 187.24] [fill={rgb, 255:red, 74; green, 144; blue, 226 }  ,fill opacity=1 ][line width=0.08]  [draw opacity=0] (9.29,-4.46) -- (0,0) -- (9.29,4.46) -- cycle    ;
	\draw [color={rgb, 255:red, 74; green, 144; blue, 226 }  ,draw opacity=1 ][line width=1.5]    (123.07,122.67) .. controls (155.63,117.32) and (161.38,114.52) .. (184.65,113.61) ;
	\draw [shift={(188.48,113.48)}, rotate = 176.33] [fill={rgb, 255:red, 74; green, 144; blue, 226 }  ,fill opacity=1 ][line width=0.08]  [draw opacity=0] (9.29,-4.46) -- (0,0) -- (9.29,4.46) -- cycle    ;
	\draw [color={rgb, 255:red, 74; green, 144; blue, 226 }  ,draw opacity=1 ][line width=1.5]    (121.11,11.07) .. controls (138.17,10.01) and (152.36,10.33) .. (163.63,10.88) .. controls (164.97,10.94) and (166.27,11.01) .. (167.52,11.08) .. controls (175.48,11.53) and (181.78,12.03) .. (186.38,12.1) ;
	\draw [shift={(190.33,12.01)}, rotate = 182.91] [fill={rgb, 255:red, 74; green, 144; blue, 226 }  ,fill opacity=1 ][line width=0.08]  [draw opacity=0] (9.29,-4.46) -- (0,0) -- (9.29,4.46) -- cycle    ;
	\draw [color={rgb, 255:red, 155; green, 155; blue, 155 }  ,draw opacity=1 ][line width=1.5]    (122.12,10.74) .. controls (142.06,12.53) and (161.73,21.85) .. (187.63,22.03) ;
	\draw [shift={(191.33,21.99)}, rotate = 186.6] [fill={rgb, 255:red, 155; green, 155; blue, 155 }  ,fill opacity=1 ][line width=0.08]  [draw opacity=0] (9.29,-4.46) -- (0,0) -- (9.29,4.46) -- cycle    ;
	\draw [color={rgb, 255:red, 155; green, 155; blue, 155 }  ,draw opacity=1 ][line width=1.5]    (122.12,10.74) .. controls (156.13,5.4) and (162.14,2.59) .. (186.44,1.68) ;
	\draw [shift={(190.44,1.55)}, rotate = 176.65] [fill={rgb, 255:red, 155; green, 155; blue, 155 }  ,fill opacity=1 ][line width=0.08]  [draw opacity=0] (9.29,-4.46) -- (0,0) -- (9.29,4.46) -- cycle    ;

\end{tikzpicture}

%% file: images/multimodality_strategies/best_x_modes.tex
\tikzset{every picture/.style={line width=0.75pt}} 

\begin{tikzpicture}[x=0.75pt,y=0.75pt,yscale=-1,xscale=1]
	
	\draw [color={rgb, 255:red, 155; green, 155; blue, 155 }  ,draw opacity=1 ][line width=1.5]    (3.33,115.63) .. controls (31.17,100.06) and (50.35,121.95) .. (60.53,123.7) ;
	\draw [shift={(64.33,123.04)}, rotate = 191.95] [fill={rgb, 255:red, 155; green, 155; blue, 155 }  ,fill opacity=1 ][line width=0.08]  [draw opacity=0] (9.29,-4.46) -- (0,0) -- (9.29,4.46) -- cycle    ;
	\draw [color={rgb, 255:red, 74; green, 144; blue, 226 }  ,draw opacity=1 ][line width=1.5]    (3.33,115.63) .. controls (9.78,120.82) and (16.19,134.01) .. (22.85,148.7) .. controls (34.23,173.79) and (46.33,203.26) .. (60.57,204.82) ;
	\draw [shift={(64.33,204.57)}, rotate = 200.95] [fill={rgb, 255:red, 74; green, 144; blue, 226 }  ,fill opacity=1 ][line width=0.08]  [draw opacity=0] (9.29,-4.46) -- (0,0) -- (9.29,4.46) -- cycle    ;
	\draw [color={rgb, 255:red, 74; green, 144; blue, 226 }  ,draw opacity=1 ][line width=1.5]    (3.33,115.63) .. controls (33.31,73.38) and (38.6,51.2) .. (60.02,44.01) ;
	\draw [shift={(63.54,42.99)}, rotate = 157.73] [fill={rgb, 255:red, 74; green, 144; blue, 226 }  ,fill opacity=1 ][line width=0.08]  [draw opacity=0] (9.29,-4.46) -- (0,0) -- (9.29,4.46) -- cycle    ;
	\draw [color={rgb, 255:red, 155; green, 155; blue, 155 }  ,draw opacity=1 ][line width=1.5]    (67.24,204.15) .. controls (84.97,209.55) and (102.67,243.85) .. (125.66,245.32) ;
	\draw [shift={(129.33,245.27)}, rotate = 205.64] [fill={rgb, 255:red, 155; green, 155; blue, 155 }  ,fill opacity=1 ][line width=0.08]  [draw opacity=0] (9.29,-4.46) -- (0,0) -- (9.29,4.46) -- cycle    ;
	\draw [color={rgb, 255:red, 74; green, 144; blue, 226 }  ,draw opacity=1 ][line width=1.5]    (67.24,204.15) .. controls (97.48,188.05) and (96.72,188.58) .. (117.48,186.76) ;
	\draw [shift={(121.33,186.42)}, rotate = 174.9] [fill={rgb, 255:red, 74; green, 144; blue, 226 }  ,fill opacity=1 ][line width=0.08]  [draw opacity=0] (9.29,-4.46) -- (0,0) -- (9.29,4.46) -- cycle    ;
	\draw [color={rgb, 255:red, 155; green, 155; blue, 155 }  ,draw opacity=1 ][line width=1.5]    (63.09,40.5) .. controls (93.49,35.14) and (114.39,42.75) .. (125.39,43.27) ;
	\draw [shift={(129.33,43.03)}, rotate = 183.81] [fill={rgb, 255:red, 155; green, 155; blue, 155 }  ,fill opacity=1 ][line width=0.08]  [draw opacity=0] (9.29,-4.46) -- (0,0) -- (9.29,4.46) -- cycle    ;
	\draw [color={rgb, 255:red, 74; green, 144; blue, 226 }  ,draw opacity=1 ][line width=1.5]    (63.09,40.5) .. controls (82.08,45.31) and (100.81,70.27) .. (125.4,71.02) ;
	\draw [shift={(129.33,70.93)}, rotate = 197.65] [fill={rgb, 255:red, 74; green, 144; blue, 226 }  ,fill opacity=1 ][line width=0.08]  [draw opacity=0] (9.29,-4.46) -- (0,0) -- (9.29,4.46) -- cycle    ;
	\draw [color={rgb, 255:red, 74; green, 144; blue, 226 }  ,draw opacity=1 ][line width=1.5]    (63.09,40.5) .. controls (95.64,26.04) and (101.39,18.45) .. (124.65,15.99) ;
	\draw [shift={(128.48,15.65)}, rotate = 170.81] [fill={rgb, 255:red, 74; green, 144; blue, 226 }  ,fill opacity=1 ][line width=0.08]  [draw opacity=0] (9.29,-4.46) -- (0,0) -- (9.29,4.46) -- cycle    ;
	\draw [color={rgb, 255:red, 155; green, 155; blue, 155 }  ,draw opacity=1 ][line width=1.5]    (123.11,215.78) .. controls (140.17,214.75) and (154.36,215.09) .. (165.63,215.67) .. controls (166.97,215.74) and (168.27,215.81) .. (169.52,215.88) .. controls (177.48,216.34) and (183.78,216.86) .. (188.38,216.93) ;
	\draw [shift={(192.33,216.85)}, rotate = 183.02] [fill={rgb, 255:red, 155; green, 155; blue, 155 }  ,fill opacity=1 ][line width=0.08]  [draw opacity=0] (9.29,-4.46) -- (0,0) -- (9.29,4.46) -- cycle    ;
	\draw [color={rgb, 255:red, 74; green, 144; blue, 226 }  ,draw opacity=1 ][line width=1.5]    (124.12,215.45) .. controls (144.06,217.28) and (163.72,226.63) .. (189.61,226.87) ;
	\draw [shift={(193.32,226.84)}, rotate = 186.71] [fill={rgb, 255:red, 74; green, 144; blue, 226 }  ,fill opacity=1 ][line width=0.08]  [draw opacity=0] (9.29,-4.46) -- (0,0) -- (9.29,4.46) -- cycle    ;
	\draw [color={rgb, 255:red, 74; green, 144; blue, 226 }  ,draw opacity=1 ][line width=1.5]    (124.12,215.45) .. controls (158.14,210.17) and (164.15,207.38) .. (188.46,206.52) ;
	\draw [shift={(192.45,206.4)}, rotate = 176.76] [fill={rgb, 255:red, 74; green, 144; blue, 226 }  ,fill opacity=1 ][line width=0.08]  [draw opacity=0] (9.29,-4.46) -- (0,0) -- (9.29,4.46) -- cycle    ;
	\draw [color={rgb, 255:red, 155; green, 155; blue, 155 }  ,draw opacity=1 ][line width=1.5]    (122.11,187) .. controls (139.17,185.94) and (153.36,186.25) .. (164.63,186.81) .. controls (165.97,186.87) and (167.27,186.94) .. (168.52,187.01) .. controls (176.48,187.46) and (182.78,187.96) .. (187.38,188.03) ;
	\draw [shift={(191.33,187.94)}, rotate = 182.91] [fill={rgb, 255:red, 155; green, 155; blue, 155 }  ,fill opacity=1 ][line width=0.08]  [draw opacity=0] (9.29,-4.46) -- (0,0) -- (9.29,4.46) -- cycle    ;
	\draw [color={rgb, 255:red, 74; green, 144; blue, 226 }  ,draw opacity=1 ][line width=1.5]    (123.12,186.67) .. controls (143.06,188.46) and (162.73,197.78) .. (188.63,197.96) ;
	\draw [shift={(192.33,197.92)}, rotate = 186.6] [fill={rgb, 255:red, 74; green, 144; blue, 226 }  ,fill opacity=1 ][line width=0.08]  [draw opacity=0] (9.29,-4.46) -- (0,0) -- (9.29,4.46) -- cycle    ;
	\draw [color={rgb, 255:red, 74; green, 144; blue, 226 }  ,draw opacity=1 ][line width=1.5]    (123.12,186.67) .. controls (157.13,181.33) and (163.14,178.52) .. (187.44,177.61) ;
	\draw [shift={(191.44,177.48)}, rotate = 176.65] [fill={rgb, 255:red, 74; green, 144; blue, 226 }  ,fill opacity=1 ][line width=0.08]  [draw opacity=0] (9.29,-4.46) -- (0,0) -- (9.29,4.46) -- cycle    ;
	\draw [color={rgb, 255:red, 74; green, 144; blue, 226 }  ,draw opacity=1 ][line width=1.5]    (67.24,204.15) .. controls (79.5,207.88) and (90.46,212.49) .. (103.56,214.57) .. controls (109.34,215.48) and (115.54,215.91) .. (122.45,215.55) ;
	\draw [shift={(126.33,215.27)}, rotate = 183.52] [fill={rgb, 255:red, 74; green, 144; blue, 226 }  ,fill opacity=1 ][line width=0.08]  [draw opacity=0] (9.29,-4.46) -- (0,0) -- (9.29,4.46) -- cycle    ;
	\draw [color={rgb, 255:red, 74; green, 144; blue, 226 }  ,draw opacity=1 ][line width=1.5]    (124.11,15.07) .. controls (141.17,14.01) and (155.36,14.33) .. (166.63,14.88) .. controls (167.97,14.94) and (169.27,15.01) .. (170.52,15.08) .. controls (178.48,15.53) and (184.78,16.03) .. (189.38,16.1) ;
	\draw [shift={(193.33,16.01)}, rotate = 182.91] [fill={rgb, 255:red, 74; green, 144; blue, 226 }  ,fill opacity=1 ][line width=0.08]  [draw opacity=0] (9.29,-4.46) -- (0,0) -- (9.29,4.46) -- cycle    ;
	\draw [color={rgb, 255:red, 74; green, 144; blue, 226 }  ,draw opacity=1 ][line width=1.5]    (125.12,14.74) .. controls (145.06,16.53) and (164.73,25.85) .. (190.63,26.03) ;
	\draw [shift={(194.33,25.99)}, rotate = 186.6] [fill={rgb, 255:red, 74; green, 144; blue, 226 }  ,fill opacity=1 ][line width=0.08]  [draw opacity=0] (9.29,-4.46) -- (0,0) -- (9.29,4.46) -- cycle    ;
	\draw [color={rgb, 255:red, 155; green, 155; blue, 155 }  ,draw opacity=1 ][line width=1.5]    (125.12,14.74) .. controls (159.13,9.4) and (165.14,6.59) .. (189.44,5.68) ;
	\draw [shift={(193.44,5.55)}, rotate = 176.65] [fill={rgb, 255:red, 155; green, 155; blue, 155 }  ,fill opacity=1 ][line width=0.08]  [draw opacity=0] (9.29,-4.46) -- (0,0) -- (9.29,4.46) -- cycle    ;
	\draw [color={rgb, 255:red, 155; green, 155; blue, 155 }  ,draw opacity=1 ][line width=1.5]    (126.11,71.61) .. controls (143.17,70.55) and (157.36,70.87) .. (168.63,71.42) .. controls (169.97,71.48) and (171.27,71.55) .. (172.52,71.62) .. controls (180.48,72.07) and (186.78,72.57) .. (191.38,72.64) ;
	\draw [shift={(195.33,72.55)}, rotate = 182.91] [fill={rgb, 255:red, 155; green, 155; blue, 155 }  ,fill opacity=1 ][line width=0.08]  [draw opacity=0] (9.29,-4.46) -- (0,0) -- (9.29,4.46) -- cycle    ;
	\draw [color={rgb, 255:red, 74; green, 144; blue, 226 }  ,draw opacity=1 ][line width=1.5]    (127.12,71.28) .. controls (147.06,73.07) and (166.73,82.39) .. (192.63,82.57) ;
	\draw [shift={(196.33,82.53)}, rotate = 186.6] [fill={rgb, 255:red, 74; green, 144; blue, 226 }  ,fill opacity=1 ][line width=0.08]  [draw opacity=0] (9.29,-4.46) -- (0,0) -- (9.29,4.46) -- cycle    ;
	\draw [color={rgb, 255:red, 74; green, 144; blue, 226 }  ,draw opacity=1 ][line width=1.5]    (127.12,71.28) .. controls (161.13,65.94) and (167.14,63.13) .. (191.44,62.22) ;
	\draw [shift={(195.44,62.09)}, rotate = 176.65] [fill={rgb, 255:red, 74; green, 144; blue, 226 }  ,fill opacity=1 ][line width=0.08]  [draw opacity=0] (9.29,-4.46) -- (0,0) -- (9.29,4.46) -- cycle    ;

\end{tikzpicture}

%% file: chapters/results.tex
\vspace{-5pt}
\section{RESULTS}
\vspace{-5pt}
\subsection{Implementation}\label{sec:implementation}
\vspace{-5pt}
In implementing the Multi-Branch \ac{SS-ASP} model, we use various network types to implement the components in Fig.~\ref{fig:ssp_components}. The context encoder $\phi$ embeds semantic images containing minimal driving context information (see Fig.~\ref{fig:example_predictions}) via a ResNet18 CNN with output feature dimension 256. The action encoder $\alpha$ is a 1D-CNN ActorNet model adapted from~\cite{liang2020learning} generating 128-dim. output features. The action and context predictors $\gamma$ and $\psi$, as well as the reconstructor $\xi$, are realized by three linear layers of dimensions \{512, 256, 256\} (with $\tanh$ activation). The predictors $\gamma$ and $\psi$ have an additional two-layer \ac{GRU} with hidden state dimension 256, called iteratively three times. At the output of the action predictor $\gamma$, we use an additional linear layer and a \texttt{softmax} operation to map the feature vector to pseudo-probabilities of predicted modes. We use the same kinematic bicycle model setup as in~\cite{janjos2021action} to obtain positions.

\newcommand{\hcw}{0.35}
\begin{figure}
\centering
\begin{subfigure}[]{\hcw\columnwidth}
	\includegraphics[width=1\linewidth,trim={1cm 0.6cm 1cm 0.7cm},clip]{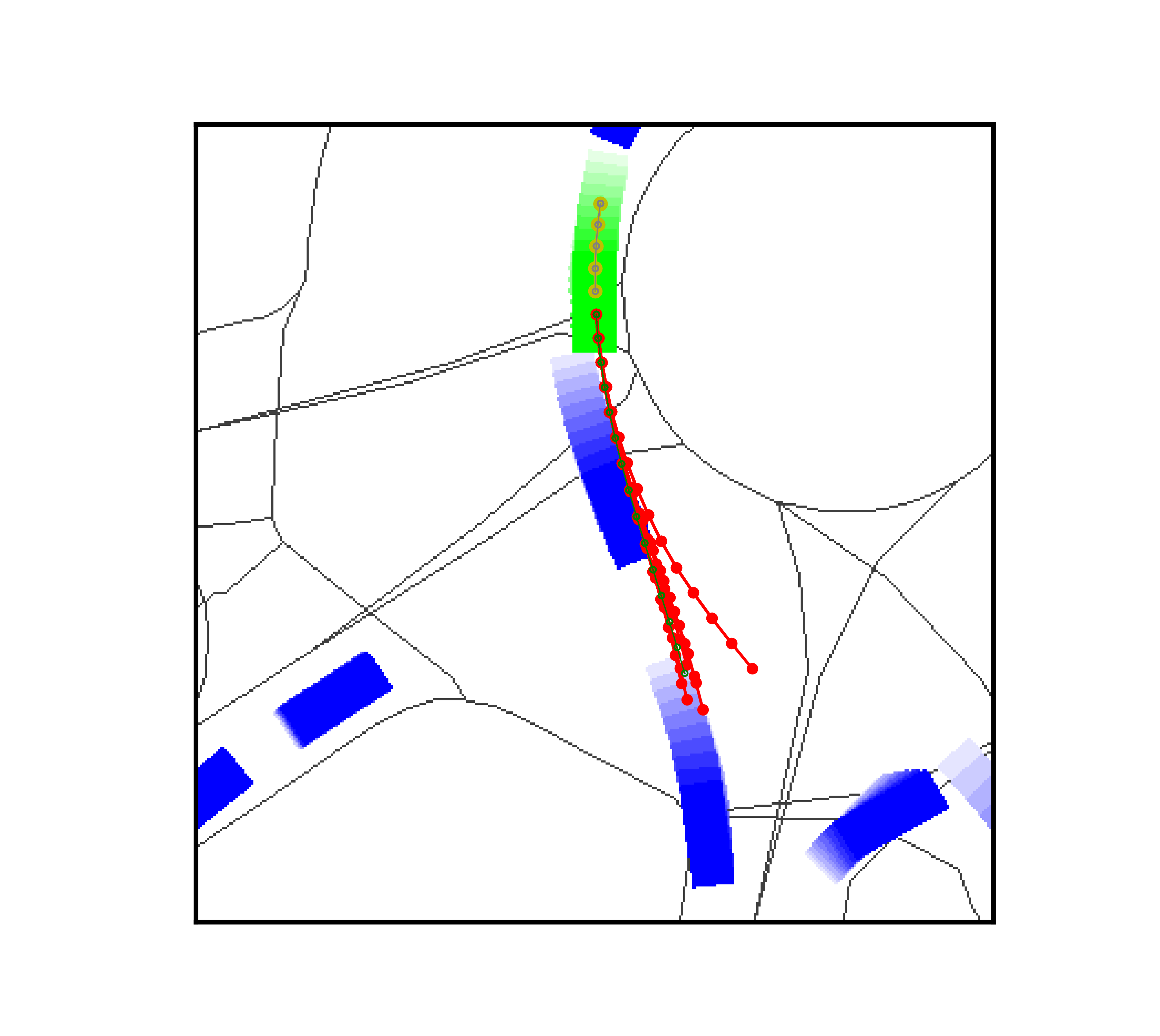}	
\end{subfigure}\hspace{-4pt}
\begin{subfigure}[]{\hcw\columnwidth}
	\includegraphics[width=1\linewidth,trim={1cm 0.6cm 1cm 0.7cm},clip]{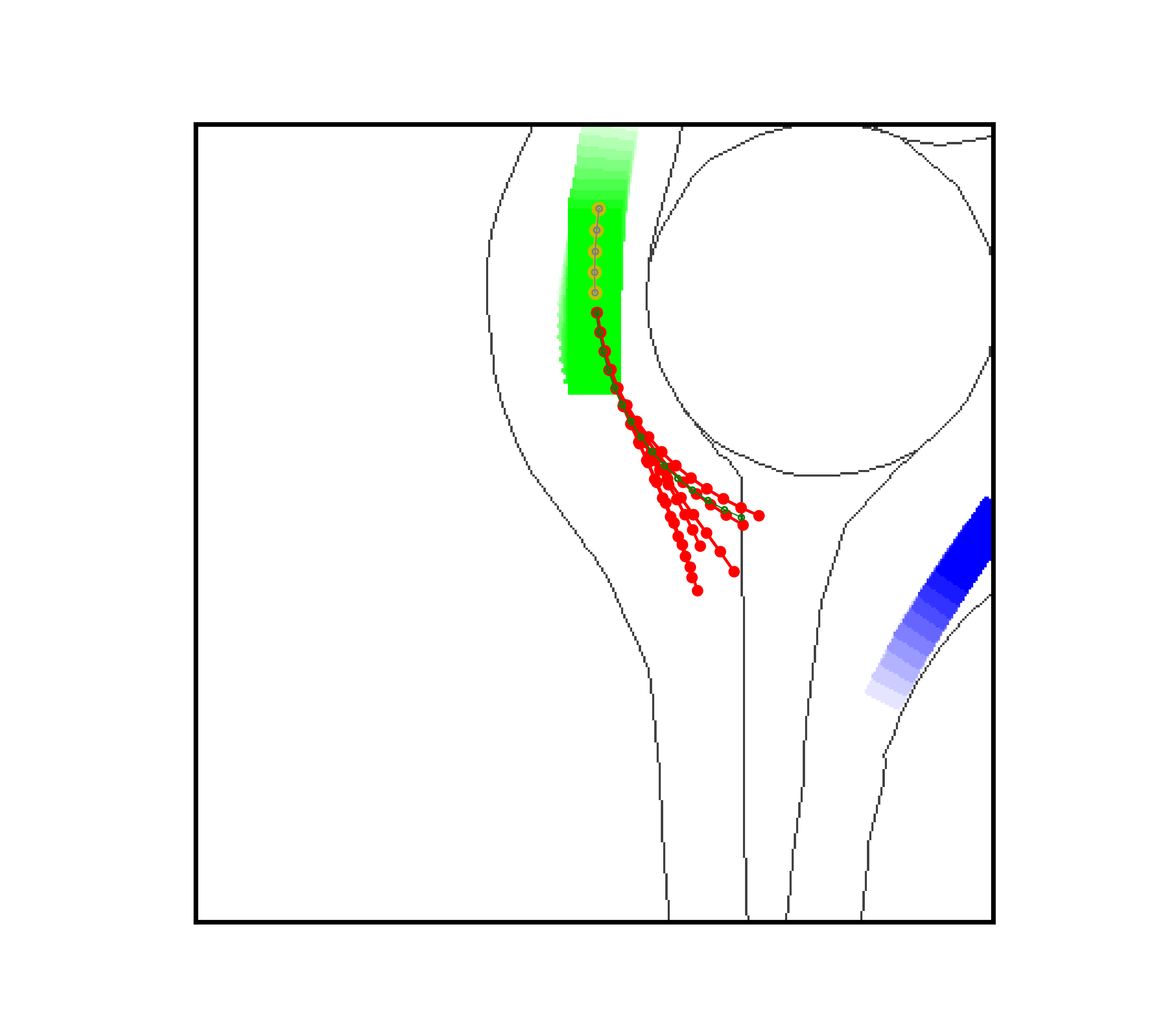}
\end{subfigure}
\begin{subfigure}[]{\hcw\columnwidth}
	\includegraphics[width=1\linewidth,trim={1cm 0.6cm 1cm 0.7cm},clip]{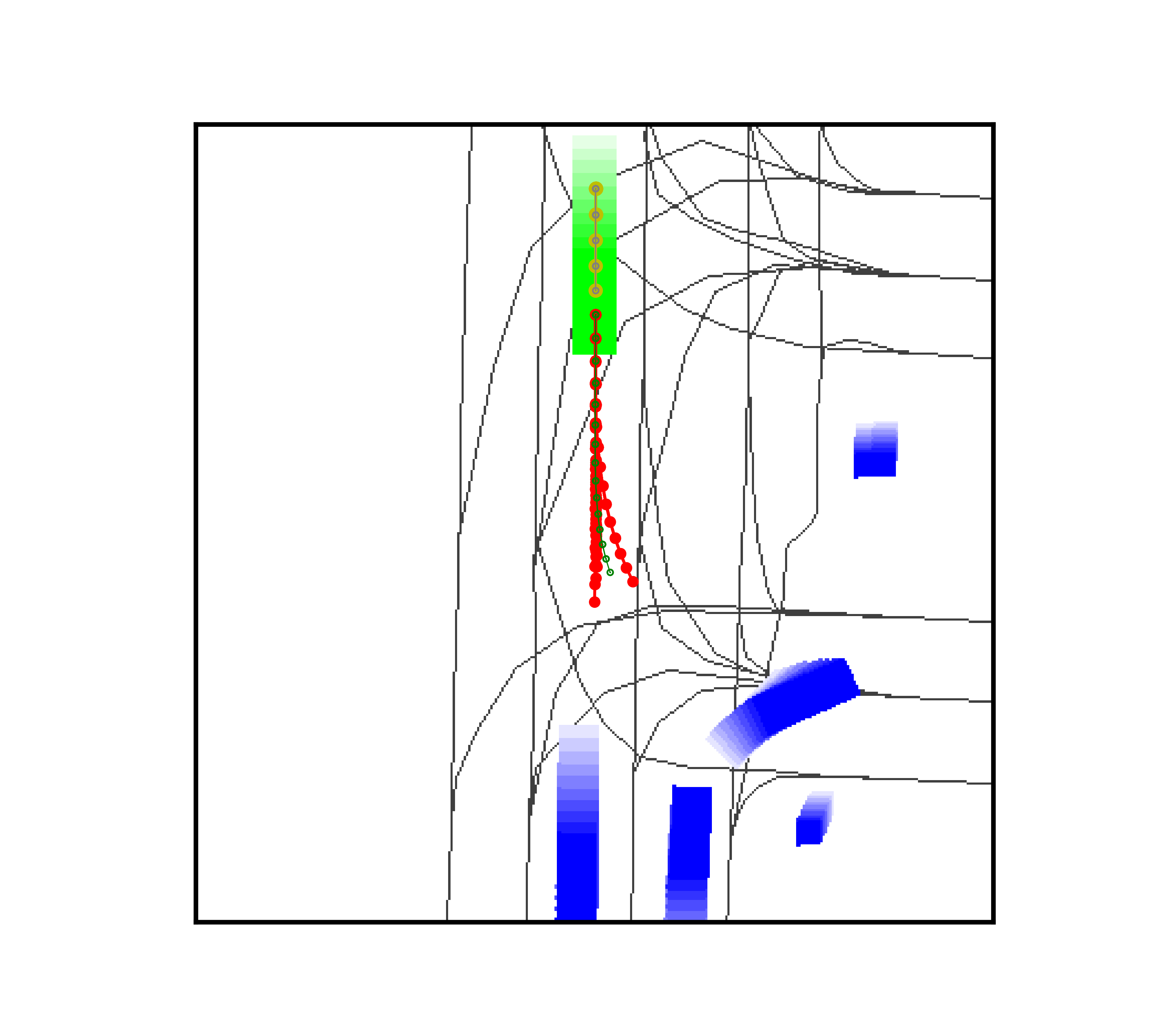}
\end{subfigure}\hspace{-4pt}
\begin{subfigure}[]{\hcw\columnwidth}
	\includegraphics[width=1\linewidth,trim={1cm 0.6cm 1cm 0.7cm},clip]{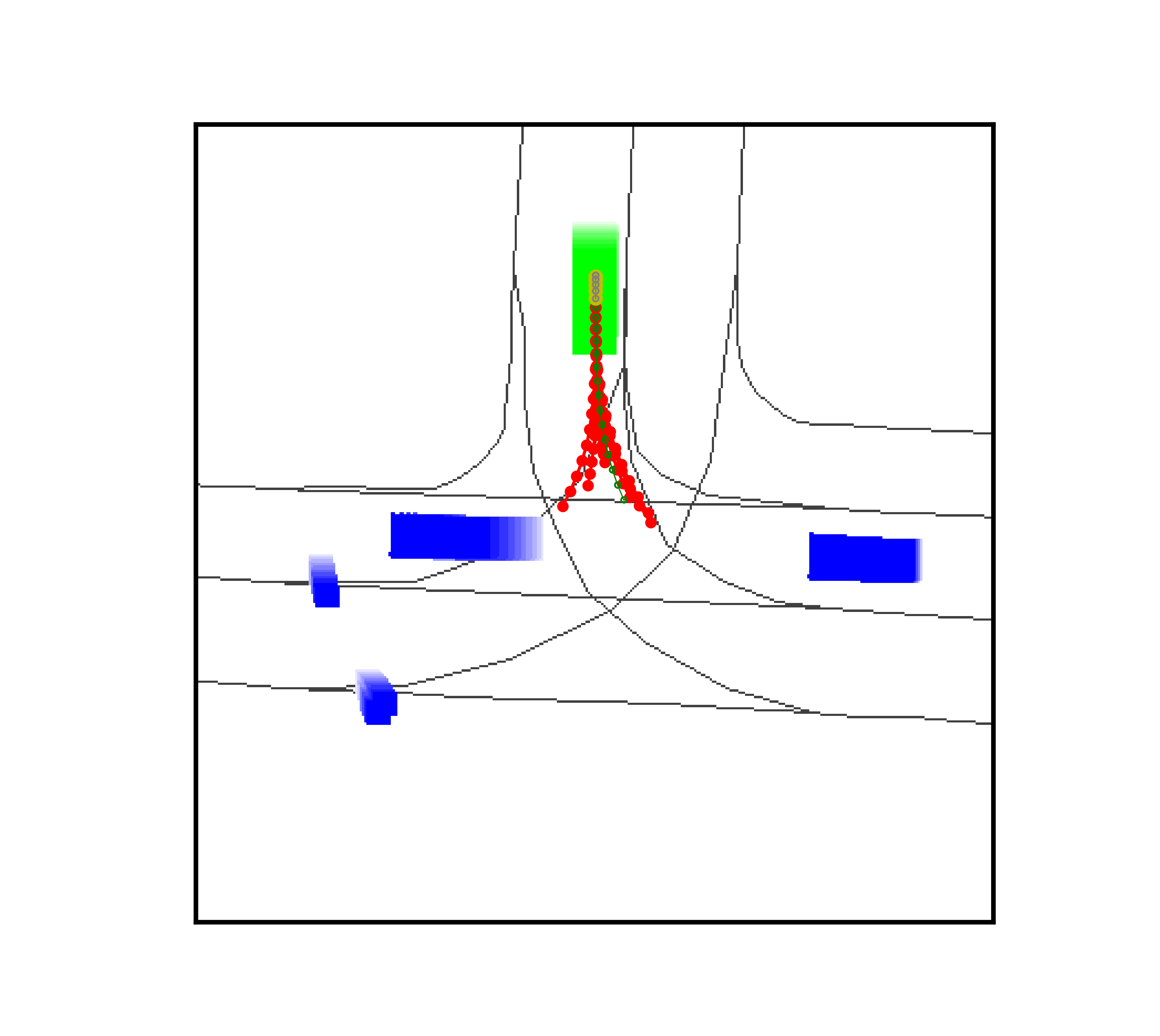}
\end{subfigure}
\caption{Multi-Branch SS-ASP predictions on the INTERACTION dataset. The model uses a simplistic image-based representation of the driving context as input, representing past agent tracks by faded bounding boxes (prediction-ego in green, other agents in blue). The $6$ predicted modes are shown in red and the ground-truth in green. The past reconstruction is shown within the prediction-ego depiction.}\vspace{-15pt}
\label{fig:example_predictions}
\end{figure}
In training the model on the loss function in Eq.~\eqref{eq:multi-branch-loss}, we used the Huber loss function for all loss components in Eq.~\eqref{eq:ssp-asp-loss} (equal weights). For the multi-segment formulation, we used the segment length of $1\si{s}$ since we found it strikes a balance in capturing rich information on a short time interval. The loss values of different tasks (trajectory prediction and reconstruction, and context prediction) within a segment are averaged. This ensures that a segment is not over-represented in the overall loss since more terms can be present in later segments depending on the combination strategy. For efficiency, we batch different component calls over modes and among different prediction branches; we observed an approximately 1.75-times increase in training time over \ac{SS-ASP}.

\subsection{Datasets and training setup}\label{sec:dataset-and-training-setup}
The models are trained on the INTERACTION~\cite{zhan2019interaction} and inD~\cite{inD} datasets (using the same partition as in~\cite{janjovs2021starnet}), with $3\si{s}$ predictions based on $1\si{s}$ history ($N=3$ in the multi-segment model). Implementation is done in PyTorch~\cite{paszke2019pytorch} with Adam optimizer~\cite{kingma2014adam} in training over 20 epochs and batch size 32, lasting two days for INTERACTION on a single Nvidia V100 GPU. The learning rate is set to $10^{-4}$ and multiplied with 0.5 if no improvement is observed in two consecutive epochs.

\subsection{Prediction performance}\label{sec:prediction-performance}

We ablate the different multi-modal trajectory combination strategies in Tab.~\ref{tab:strategy_ablation}. We see that Start-$k$ outperforms others; this is expected due to its largest diversity in the first segment. Additionally, we offer an ablation study of the proposed approaches in Tab.~\ref{tab:component_ablation}. It can be seen that (i) multi-branching and (ii) context aggregation bring boosts in metrics while multi-segmenting regresses the performance unless augmented with (i) and (ii). This is consistent with the results in~\cite{janjos2021action}, where a naive formulation with an End-$k$-like strategy is proposed. Overall, the Multi-Branch SS-ASP model brings a significant improvement of almost $25\, \%$ over the basic model. It shows that introducing additional training tasks without modifying more problem-relevant aspects (e.g. interaction modeling) can greatly improve prediction performance. 

We compare the Multi-Branch-SS-ASP prediction results to reported results of other state-of-the-art models on the INTERACTION validation dataset in Tab.~\ref{tab:results_literature}. Furthermore, we evaluate the model on the INTERACTION test set online leaderboard\footnote{\url{http://challenge.interaction-dataset.com/leader-board} as of 01-Feb-2023}, where it achieves a competitive \textit{3rd} place in minADE$_6$ and minFDE$_6$. However, it scores \textit{9th} in \ac{MR}; this is understandable since the model components are inherently ill-equipped to handle interaction modeling due to the very low-information-density environment representation and simplistic CNN encoding. For such purposes, many state-of-the-art approaches use graph- or Transformer-based~\cite{vaswani2017attention} architectures in their encoders~\cite{janjovs2021starnet, nayakanti2022wayformer} as well as target selection heuristics in their decoders~\cite{gu2021densetnt, varadarajan2021multipath++}. Such approaches could be easily integrated into the overall architecture. The generality of the self-supervision, segment-wise prediction, and branched training does not preclude component-level improvements.

\begin{table}[t!]
	\centering
	\smallskip
	\caption{\small Comparison of combination strategies for multi-segment multi-modal prediction. The parameters $k$ and $m$ are chosen such that the same number of resulting modes is obtained, $K=8$.}
	\scalebox{0.88}{\input{tables/strategy_ablation.tex}}
	\label{tab:strategy_ablation}\vspace{-10pt}
\end{table}
\begin{table}[t!]
	\centering
	\smallskip
	\caption{\small Ablation study of proposed approaches: self-supervision, multi-segment chaining (with Start-$k$ strategy, $k=9$), branched overshooting, and context aggregation. Multi-branch* denotes branched overshooting (Sec.~\ref{subsubsec:branched_overshoot}) without context aggregation (Sec.~\ref{subsubsec:context_agg}).}
	\scalebox{0.88}{\input{tables/component_ablation.tex}}
	\label{tab:component_ablation}\vspace{-10pt}
\end{table}
\begin{table}[t!]
	\centering
	\smallskip
	\caption{\small Minimal displacement metrics on the INTERACTION validation dataset. All methods predict $K=6$ modes. We do not include~\cite{janjovs2021starnet,SAN} due to a different number of predicted modes.}
	\scalebox{0.88}{\input{tables/results_literature.tex}}
	\label{tab:results_literature}\vspace{-10pt}
\end{table}

\subsection{Prediction uncertainty estimation} \label{sec:prediction-uncertainty-estimation-results}
We evaluate the uncertainty quantification strategies from Sec.~\ref{sec:uncertainty-estimation-methods} by observing whether they correlate with the change in prediction error over successive segments. We quantify the prediction error by the $\Delta \text{minADE}$ in Eq.~\eqref{eq:delta-min-ade}. In this way, a high value of the uncertainty metric could indicate that the model's predictions will deteriorate over time. 

We calculate the metrics from Sec.~\ref{sec:uncertainty-estimation-methods} for each predicted segment on a randomly chosen $10\, \%$ inD subset. The results are visualized in Fig.~\ref{fig:uncertainty_metrics_results}. To ensure comparability between the two methods, we group the (sorted) obtained values into four quarters, where each quarter contains $25 \, \%$ of the overall values (the first quarter is equivalent to the first quantile). Then, within each quarter we approximate the $\Delta$minADE$_k$ error distribution by a four bin histogram (lightest to darkest blue in  Fig.~\ref{fig:uncertainty_metrics_results}). The bin intervals are determined by the quarters of the $\Delta$minADE$_k$ error distribution on the validation set. 

Interpreting Fig.~\ref{fig:uncertainty_metrics_results}, we see that the change in the error distribution between quarters is evident. For example, quartile 1 of segment 1 (Fig.~\ref{fig:recon_error_seg1}) contains the lowest-reconstruction-error and more than $60 \, \%$ of its values lie in the low $\Delta$minADE$_k$ range (lightest blue). Similarly, in quarter 4 of Fig.~\ref{fig:recon_error_seg1} (containing the highest metric values) the histogram distribution is biased towards high $\Delta$minADE$_k$ samples (darkest blue). Therefore, a correlation between the metric and the actual change in prediction error over segments can be confirmed. Similar relationships can be found for the dropout-based mean-of-mode-variances, where $20$ Monte Carlo runs are performed (we dropped-out the two linear layers before the action predictor with $p=0.5$). Furthermore, we observe that in both metrics the histogram distribution favors higher $\Delta$minADE$_k$ at later segments, which is reasonable. 

\newcommand\cw{0.32}
\begin{figure*}
\centering
\begin{subfigure}[]{\cw\columnwidth}
	\includegraphics[width=1\linewidth,trim={0.65cm 0.65cm 0.65cm 0.65cm},clip]{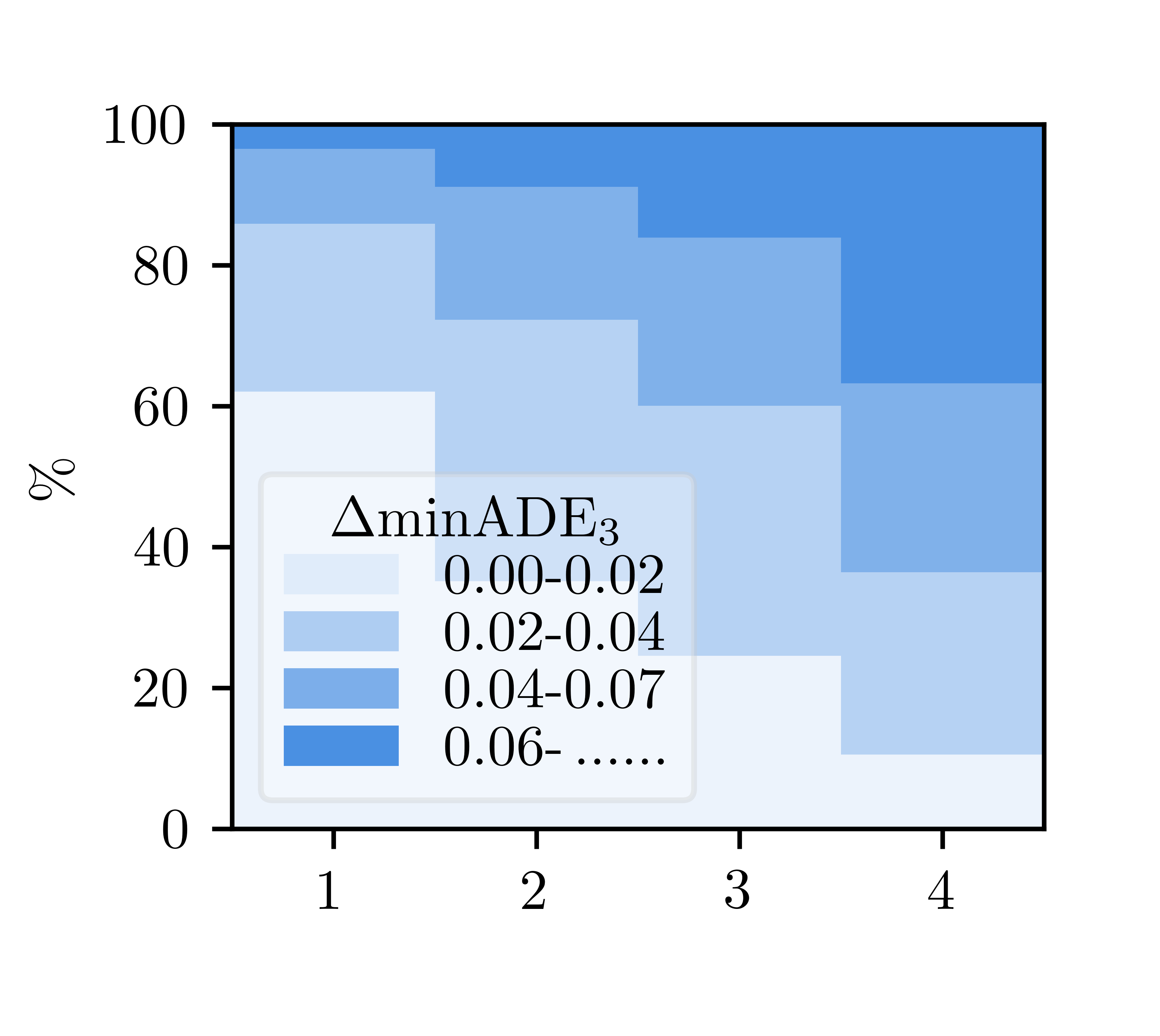}
	\centering
	\caption{Rec. error seg. 1}
	\label{fig:recon_error_seg1}
\end{subfigure}
\hfill
\begin{subfigure}[]{\cw\columnwidth}
	\includegraphics[width=1\linewidth,trim={0.65cm 0.65cm 0.65cm 0.65cm},clip]{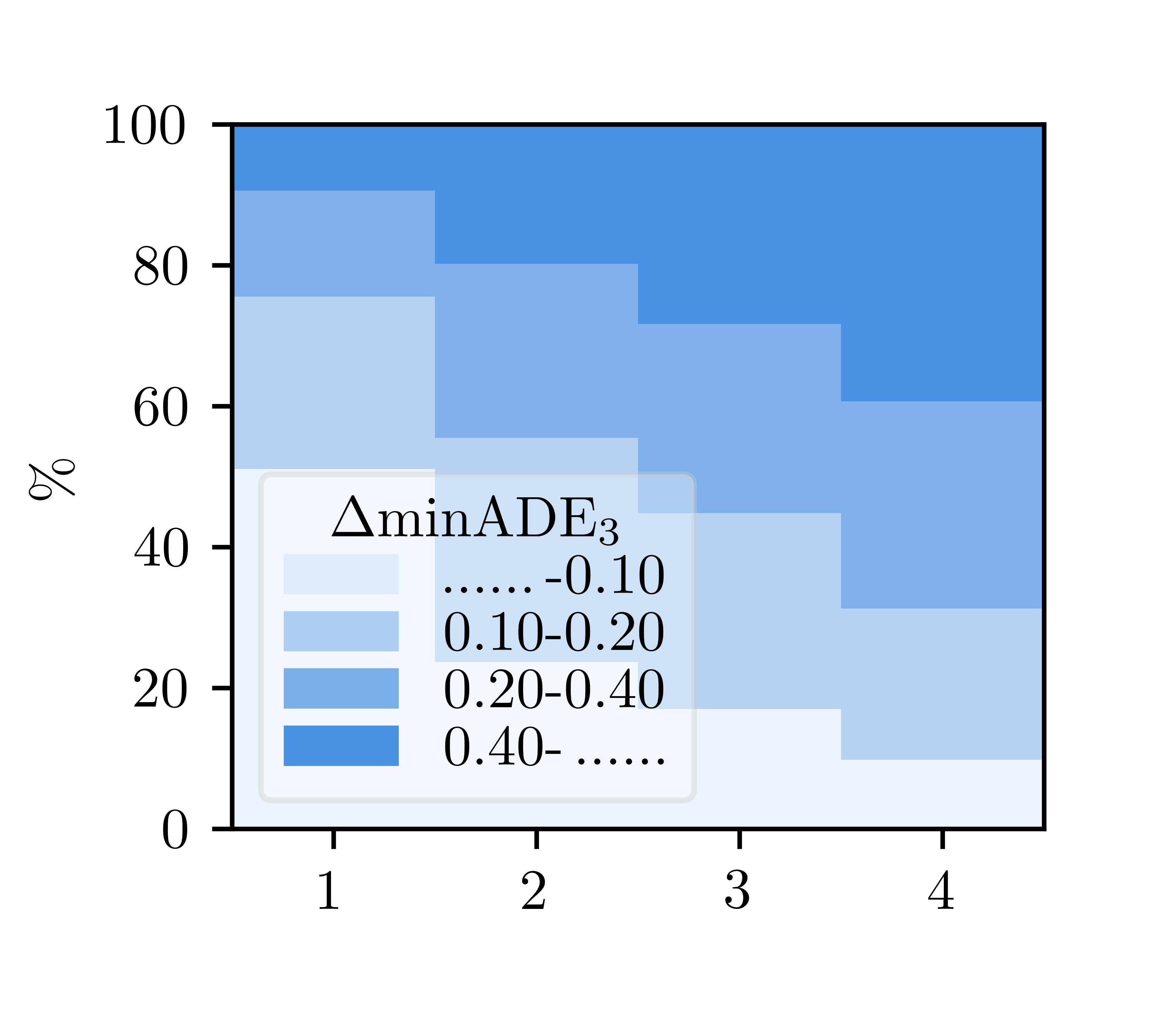}
	\caption{Rec. error seg. 2}
\end{subfigure}
\hfill
\begin{subfigure}[]{\cw\columnwidth}
	\includegraphics[width=1\linewidth,trim={0.65cm 0.65cm 0.65cm 0.65cm},clip]{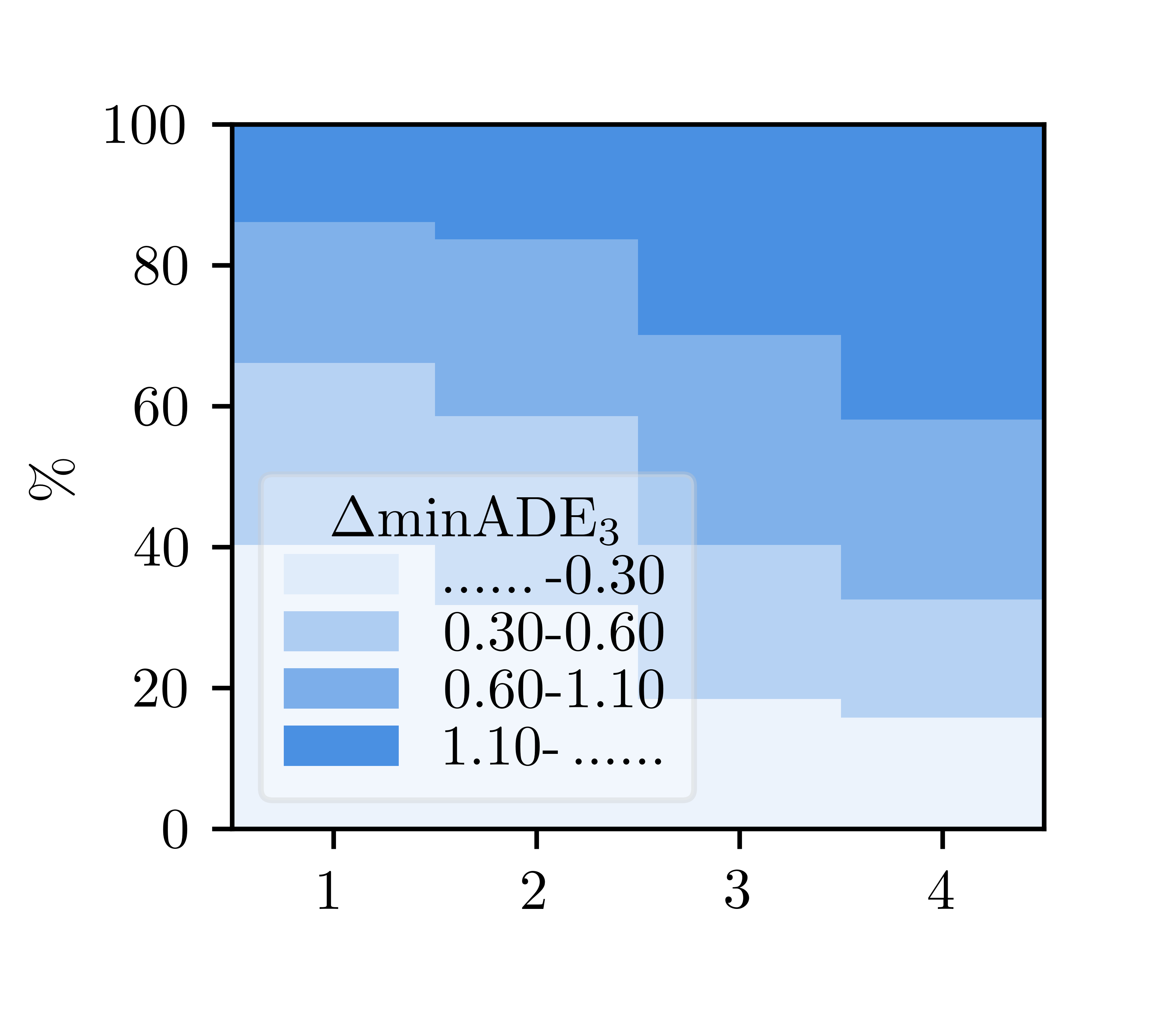}
	\caption{Rec. error seg. 3}
\end{subfigure}
\hfill
\begin{subfigure}[]{\cw\columnwidth}
	\centering
	\includegraphics[width=1\linewidth,trim={0.65cm 0.65cm 0.65cm 0.65cm},clip]{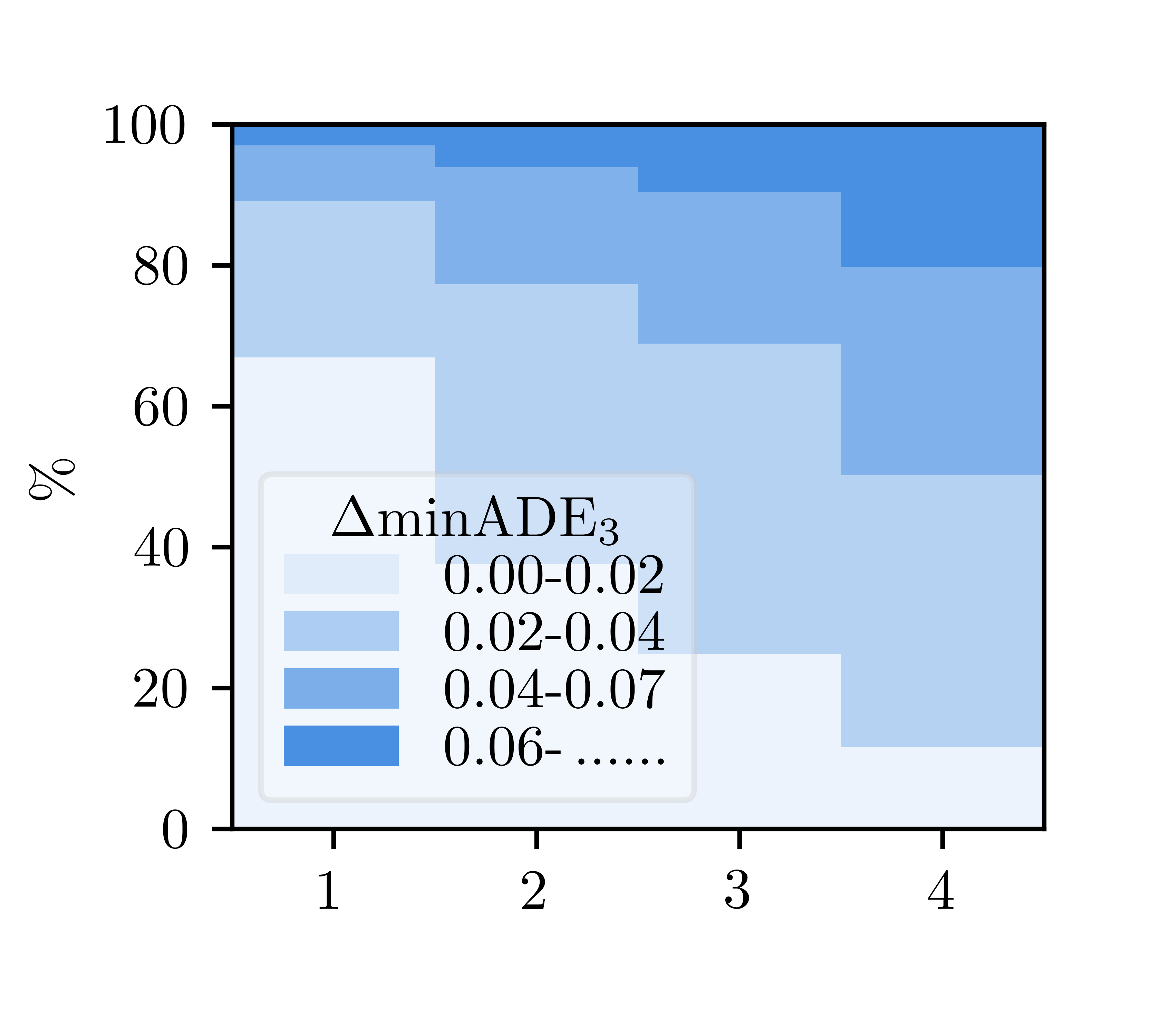}
	\caption{Mode var. seg. 1}
\end{subfigure}
\hfill
\begin{subfigure}[]{\cw\columnwidth}
	\includegraphics[width=1\linewidth,trim={0.65cm 0.65cm 0.5cm 0.65cm},clip]{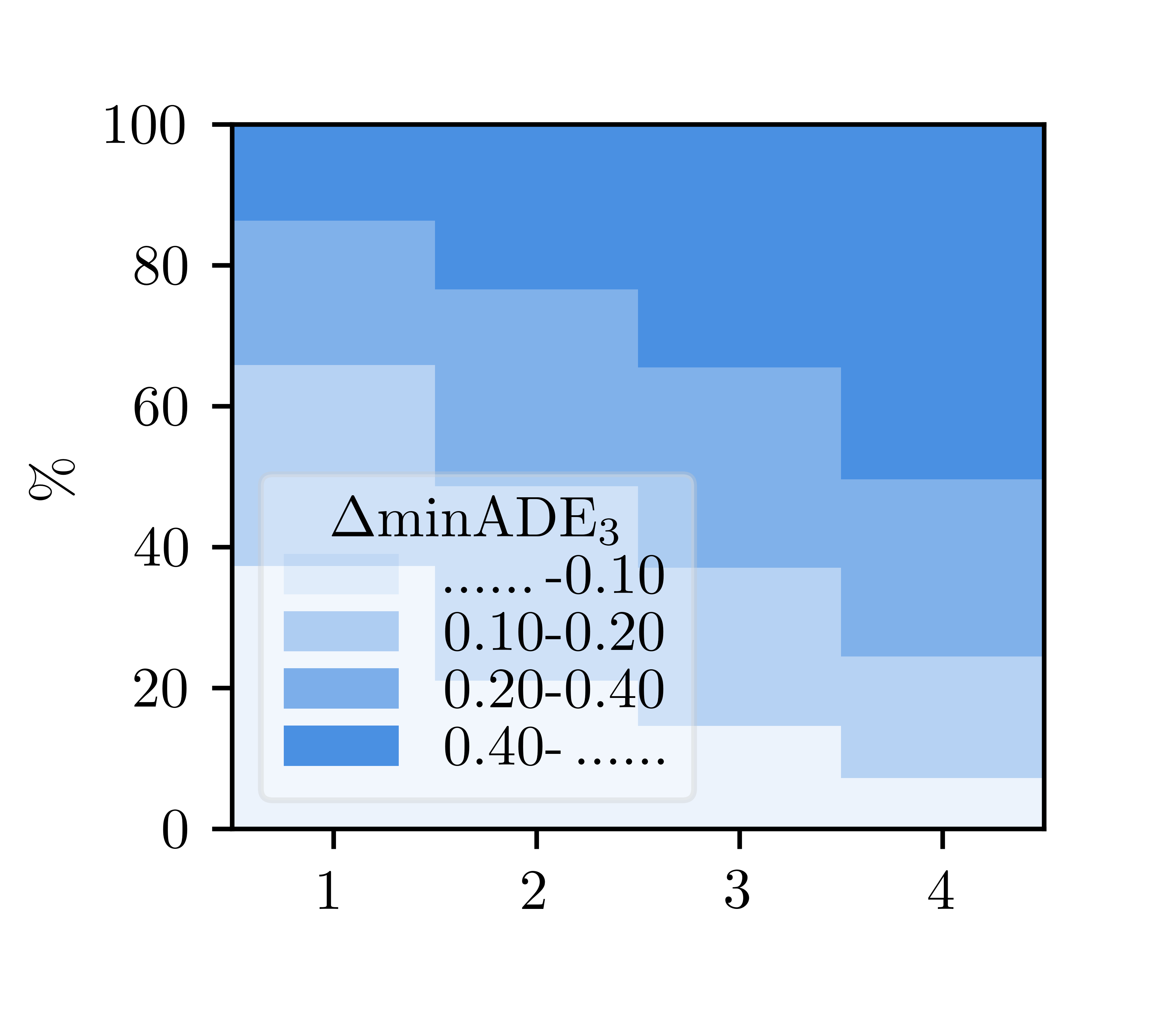}
	\caption{Mode var. seg. 2}
\end{subfigure}
\hfill
\begin{subfigure}[]{\cw\columnwidth}
	\includegraphics[width=1\linewidth,trim={0.65cm 0.65cm 0.65cm 0.65cm},clip]{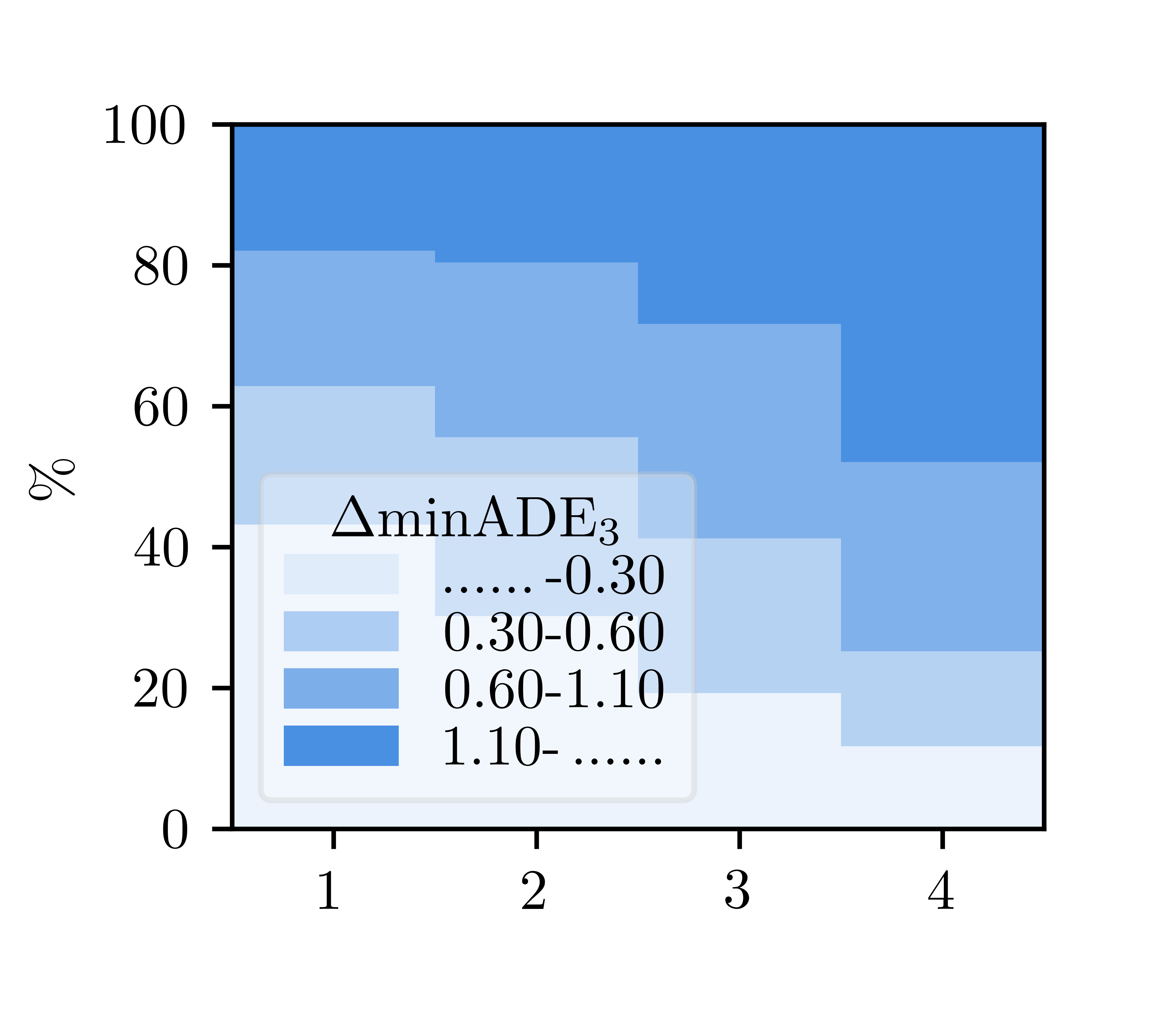}
	\caption{Mode var. seg. 3}
\end{subfigure}
\caption{Evaluation of reconstruction error and mean-of-mode-variances uncertainty estimation metrics over three predicted segments. The metric values are split into four quarters (horizontal axes) and the distribution of each quarter w.r.t. $\Delta$minADE$_k$ (Eq.~\eqref{eq:delta-min-ade}) is shown in different shades of blue (vertical axes indicate $\, \%$). The metrics correlate with prediction error change -- it can be seen that lower metric quarters contain mostly lower $\Delta$minADE$_k$ values (lighter shades of blue) and higher metrics contain higher  $\Delta$minADE$_k$ (darker shades).}\vspace{-10pt}
\label{fig:uncertainty_metrics_results}
\end{figure*}

%% file: tables/strategy_ablation.tex
\begin{tabularx}{0.5\textwidth}{c *{3}{Y}}
	\toprule
	& & \multicolumn{2}{c}{inD \cite{inD}}  \\
	\cmidrule(l){3-4}
	Method & Configuration & minADE$_6$ & minFDE$_6$\\
	\midrule
	All-Modes & $k=2$ & 0.25 & 0.61 \\
	Start-$k$ & $k=8$ & \textbf{0.20} & \textbf{0.51} \\
	End-$k$ & $k=8$ & \textbf{0.20} & 0.53 \\
	Best-$m$-of-All & $m = 8$ & 0.24 & 0.59 \\
	Best-$m$-of-Pred. & $m = 2$ & 0.22 & 0.54\\
	\bottomrule
\end{tabularx}

%% file: tables/component_ablation.tex
\begin{tabularx}{0.5\textwidth}{c *{6}{Y}}
	\toprule
	& & & & & \multicolumn{2}{c}{inD \cite{inD}}  \\
	\cmidrule(l){6-7}
	Model & Self-Sup. & Multi-seg. & Multi-branch & Context agg. & min-ADE$_9$ & min-FDE$_9$\\
	\midrule
	FF-ASP~\cite{janjos2021action} & \xmark & \xmark & \xmark & \xmark & 0.22 & 0.56 \\
	SS-ASP~\cite{janjos2021action} & \cmark & \xmark & \xmark & \xmark & 0.19 & 0.50 \\
	Multi-seg. SS-ASP & \cmark & \cmark & \xmark & \xmark & 0.20 & 0.52 \\
	Multi-branch* SS-ASP& \cmark & \cmark & \cmark & \xmark & 0.17 & 0.45 \\
	Multi-branch SS-ASP& \cmark & \cmark & \cmark & \cmark & \textbf{0.17} & \textbf{0.43} \\
	\bottomrule
\end{tabularx}

%% file: tables/results_literature.tex
\begin{tabularx}{0.5\textwidth}{c *{2}{Y}}
	\toprule
	& \multicolumn{2}{c}{INTERACTION \cite{zhan2019interaction}}  \\
	\cmidrule(l){2-3}
	& minADE$_6$ & minFDE$_6$\\
	\midrule
	TNT \cite{zhao2020tnt}  &  0.21 & 0.67 \\
	STG-DAT \cite{li2021spatio} & 0.29 & 0.54 \\
	ITRA \cite{scibior2021imagining}  &  0.17 & 0.49 \\
	GOHOME \cite{gilles2021gohome} & - & 0.45  \\
	FF-ASP \cite{janjos2021action} & 0.12 & 0.35 \\
	DIPA \cite{knittel2022dipa} & 0.11 & 0.34  \\
	SS-ASP \cite{janjos2021action} & 0.11 & 0.33  \\
	\cmidrule(l){2-3} 
	Multi-Branch SS-ASP & \textbf{0.10} & \textbf{0.30} \\
	\bottomrule
\end{tabularx}

%% file: chapters/conclusion.tex
\section{CONCLUSION}
In this paper, we investigated connections between one-shot and autoregressive trajectory prediction models. We deliberately focused on the structure of output representations and the training approach, as opposed to more problem-relevant aspects such as driving context and interaction modeling, in order to better see the effects of the proposed approach. We found significant gains by converting an existing one-shot predictor into a novel, segment-wise prediction trained with self-supervision and overshooting. Furthermore, we proposed two epistemic uncertainty measures for deterministic predictors. In combination with the segment-wise output structure, they pave way for prediction of a variable time horizon with the goal of providing only confident predictions to a downstream planner.